\documentclass[pdflatex,sn-apa,iicol]{sn-jnl}% APA Reference Style
\usepackage{graphicx}%
\usepackage{multirow}%
\usepackage{amsmath,amssymb,amsfonts}%
\usepackage{amsthm}%
\usepackage{mathrsfs}%
\usepackage[title]{appendix}%
\usepackage[table]{xcolor}%
\usepackage{textcomp}%
\usepackage{manyfoot}%
\usepackage{booktabs}%
\usepackage{algorithm}%
\usepackage{algorithmicx}%
\usepackage{algpseudocode}%
\usepackage{listings}%

\usepackage[printonlyused, nohyperlinks]{acronym}
\usepackage{mdframed}
\usepackage{array}
\usepackage{csquotes}
\usepackage{xltabular}
\usepackage{longtable}
\usepackage{enumitem}
\usepackage{ragged2e}
\theoremstyle{thmstyleone}%
\theoremstyle{thmstyletwo}%

\theoremstyle{thmstylethree}%

\begin{document}

\title[Physics-Informed Machine Learning in Prognostics and Health Management: A Systematic Literature Review]{Physics-Informed Machine Learning in Prognostics and Health Management: A Systematic Literature Review}


\author*[1,2]{\fnm{Christopher} \sur{Braun}\,\orcid{https://orcid.org/0009-0006-4153-4772}}\email{christopher.braun@iff.uni-stuttgart.de}
\equalcont{These authors contributed equally to this work.}

\author[1,2]{\fnm{Julian} \sur{Raible}\,\orcid{https://orcid.org/0000-0003-1696-3533}}\email{julian.raible@iff.uni-stuttgart.de}
\equalcont{These authors contributed equally to this work.}

\author[1,2]{\fnm{Marco F.} \sur{Huber}\,\orcid{https://orcid.org/0000-0002-8250-2092}}\email{marco.huber@ieee.org}

\affil[1]{\orgdiv{Institute of Industrial Manufacturing and Management IFF}, \orgname{University of Stuttgart}, \orgaddress{\street{Allmandring 35}, \city{Stuttgart}, \postcode{70569}, \state{Baden-W\"urttemberg}, \country{Germany}}}

\affil[2]{\orgname{Fraunhofer Institute for Manufacturing Engineering and Automation IPA}, \orgaddress{\street{Nobelstraße 12}, \city{Stuttgart}, \postcode{70569}, \state{Baden-W\"urttemberg}, \country{Germany}}}

% - - - - - - - - - - abstract - - - - - - - - - - %

\abstract{In modern industry, keeping complex systems reliable, safe, and efficient hinges on Prognostics and Health Management (PHM). Machine Learning (ML) has largely driven advancements in diagnostics and prognostics, yet purely data-driven models face inherent limitations, such as poor generalization, an inability to infer causal relationships, and a lack of interpretability. Physics-Informed Machine Learning (PIML) helps mitigate these limitations by incorporating prior physical knowledge directly into the ML pipeline, thereby fostering growing interest in its application to PHM. This work investigates how PIML is being leveraged in the context of PHM through a systematic literature review of 212 studies. The review introduces a four-class classification scheme, consisting of observational bias, inductive bias, learning bias, and hybrid approaches, and further categorizes studies by PHM task. Across all four classes, the reviewed studies consistently demonstrate improved predictive performance over conventional baselines across a broad range of assets, although the literature is heavily skewed toward lithium-ion batteries and bearings, and dominated by problem-specific solutions. Overall, the review indicates that physics-informed approaches already provide tangible benefits, whereas claims of improvements concerning some of the aforementioned limitations lack sufficient supporting evidence. Future research should prioritize transferable design patterns, benchmarks comparing integration strategies, and uncertainty-aware models that are lightweight and robust enough for online deployment in real-world settings.}

\keywords{Systematic literature review, Physics-informed machine learning, Prognostics and health management, Prior physical knowledge, Hybrid approaches
\vspace{-1.5cm}
}



\maketitle
\begin{strip}
\small\textit{The version of record of this article, first published in Journal of Intelligent Manufacturing, is available online at Publisher’s website: \href{https://dx.doi.org/10.1007/s10845-026-02930-3}{https://dx.doi.org/10.1007/s10845-026-02930-3}. This arXiv version is content-equivalent to the version of record but differs in four respects: (i) the list of studies excluded after full-text analysis, available as supplementary material (referred to as Online Resource 1 in the version of record), is included here as an appendix; (ii) citations are consistently disambiguated, resolving cases in which several distinct references share an identical in-text citation string; (iii) section headings are numbered, so that the numbered cross-references used throughout the text can be resolved; and (iv) all figures are embedded such that the text they contain remains selectable and searchable. No claims, results, or conclusions have been altered. Please cite the version of record.}
\normalsize
\end{strip}

% - - - - - - - - - - acronyms - - - - - - - - - - %
\begin{mdframed}
\textbf{Abbreviations (Technical Terms)}

\begin{acronym}[MOSFET] %longest acronym for spacing
  \setlength{\itemsep}{0pt}   
  \setlength{\parskip}{0pt}   
  \RaggedRight
% A
\acro{AE}[AE]{Autoencoder}

% B
\acro{BiLSTM}[BiLSTM]{Bidirectional Long Short-Term Memory}
\acro{BPFI}[BPFI]{Ball Pass Frequency Inner Race}
\acro{BPFO}[BPFO]{Ball Pass Frequency Outer Race}
\acro{BSF}[BSF]{Ball Spin Frequency}

% C
\acro{CAE}[CAE]{Convolutional Autoencoder}
\acro{CM}[CM]{Condition Monitoring}
\acro{CNN}[CNN]{Convolutional Neural Network}
\acroplural{CNN}[CNNs]{Convolutional Neural Networks}
\acro{CWT}[CWT]{Continuous Wavelet Transform}

% D
\acro{DL}[DL]{Deep Learning}
\acro{DOF}[DOF]{Degrees of Freedom}

% E
\acro{ECA}[ECA]{Efficient Channel Attention}
\acro{ECM}[ECM]{Equivalent Circuit Model}
\acroplural{ECM}[ECMs]{Equivalent Circuit Models}
\acro{EKF}[EKF]{Extended Kalman Filter}

% F
\acro{FAC}[FAC]{Frequency-Aware Convolution}
\acro{FE}[FE]{Finite Element}
\acro{FLOP}[FLOP]{Floating Point Operation}
\acroplural{FLOP}[FLOPs]{Floating Point Operations}

% G
\acro{GAN}[GAN]{Generative Adversarial Network}
\acroplural{GAN}[GANs]{Generative Adversarial Networks}
\acro{GNN}[GNN]{Graph Neural Network}
\acroplural{GNN}[GNNs]{Graph Neural Networks}
\acro{GP}[GP]{Gaussian Process}
\acro{GPR}[GPR]{Gaussian Process Regression}
\acro{GRU}[GRU]{Gated Recurrent Unit}

% H 
\acro{HI}[HI]{Health Index}
\acroplural{HI}[HIs]{Health Indices}

% I 
\acro{IML}[IML]{Informed Machine Learning}

% K 
\acro{kNN}[kNN]{K-Nearest Neighbor}

% L
\acro{LSTM}[LSTM]{Long Short-Term Memory}

% M
\acro{ML}[ML]{Machine Learning}
\acro{MLP}[MLP]{Multilayer Perceptron}

% N
\acro{NN}[NN]{Neural Network}
\acroplural{NN}[NNs]{Neural Networks}

% O
\acro{ODE}[ODE]{Ordinary Differential Equation}
\acroplural{ODE}[ODEs]{Ordinary Differential Equations}

% P
\acro{PDE}[PDE]{Partial Differential Equation}
\acroplural{PDE}[PDEs]{Partial Differential Equations}
\acro{PdM}[PdM]{Predictive Maintenance}
\acro{PF}[PF]{Particle Filter}
\acro{PHM}[PHM]{Prognostics and Health Management}
\acro{PIML}[PIML]{Physics-Informed Machine Learning}
\acro{PINN}[PINN]{Physics-Informed Neural Network}
\acroplural{PINN}[PINNs]{Physics-Informed Neural Networks}

% R
\acro{ReLU}[ReLU]{Rectified Linear Unit}
\acro{ResNet}[ResNet]{Residual Network}
\acro{RF}[RF]{Random Forest}
\acroplural{RF}[RFs]{Random Forests}
\acro{RFR}[RFR]{Random Forest Regression}
\acro{RUL}[RUL]{Remaining Useful Life}
\acro{RL}[RL]{Reinforcement Learning}
\acro{RNN}[RNN]{Recurrent Neural Network}

% S
\acro{SEI}[SEI]{Solid Electrolyte Interphase}
\acro{SOC}[SOC]{State of Charge}
\acro{SOH}[SOH]{State of Health}
\acro{SPM}[SPM]{Single-Particle Model}
\acro{SVR}[SVR]{Support Vector Regression}
\acro{SVM}[SVM]{Support Vector Machine}

% T 
\acro{TGDS}[TGDS]{Theory-Guided Data Science}
\acro{TL}[TL]{Transfer Learning}
\acro{TRL}[TRL]{Technology Readiness Level}
\acroplural{TRL}[TRLs]{Technology Readiness Levels}
\end{acronym}
\end{mdframed}

% - - - - - - - - - - subsection - - - - - - - - - - %
\section{Introduction}
\label{sec:introduction}

% Bigger picture: PHM, PdM and the industrial environment
\ac{PHM} has emerged as a cornerstone of modern industrial operations, driven by the increasing need for reliability, safety, and efficiency in complex engineering systems~\citep{vogl2019AReviewOfDiagnostic}. Specifically, \ac{PHM} provides the methodological foundation for \ac{PdM}, enabling organizations to transition from reactive or schedule-based strategies through condition-based monitoring to predictive measures, supporting informed decision-making \citep{huang2024prognostics}. This transition has been accelerated by advances in sensing technologies, connectivity, and industrial digital\-ization---central pillars of the Industry 4.0 paradigm. As industrial assets become more interconnected and operational demands intensify, \ac{PHM} plays a critical role in reducing unplanned downtime, optimizing maintenance costs, and ensuring continuous production.

% ML
\ac{ML} has significantly influenced \ac{PHM} research, with its adoption growing substantially in recent years \citep{sajjadi2025machine}. \ac{ML} models have demonstrated strong performance in tasks such as anomaly detection~\citep{cannizzaro2025MachineLearningEnabled}, fault diagnosis~\citep{zhang2025JointDistributionDomain}, and \ac{RUL} prediction~\citep{yin2025RemainingUsefulLife}. Despite these achievements, purely data-driven approaches face inherent limitations, particularly when deployed in real-world industrial settings. In such settings, sensor measurements are frequently noisy, incomplete, or inconsistent, operational conditions vary widely across assets and environments, and failure data remain sparse due to the rarity of catastrophic events. Moreover, \ac{ML} models often lack interpretability and struggle to extrapolate beyond the state space covered by the training data~\citep{hagmeyer2022integration}. These limitations are particularly critical in industrial sectors where reliability, safety, and trustworthiness are paramount, driving the need for more advanced modeling strategies.

% PIML
\ac{PIML}~\citep{karniadakis2021physics} constitutes a promising paradigm for addressing these challenges by incorporating prior physical knowledge into the \ac{ML} pipeline. By combining the scalability and predictive capabilities of ML with the structure and interpretability of physics-based modeling, \ac{PIML} offers a pathway toward robust, generalizable, and physically consistent \ac{PHM} solutions. These characteristics align strongly with the stringent reliability and transparency demands of industrial settings, where decisions based on model outputs often carry significant operational or safety implications \citep{zio2022prognostics}.

% Explaining the focus on the industrial environment (and the relevance of PIML in this context)
Beyond methodological motivations, the industrial context itself amplifies the relevance of \ac{PIML}. Modern industrial systems operate under harsh, dynamic, and heterogeneous conditions. Assets may experience variable loads, nonlinear degradation, or rapid transitions between operating regimes. Sensor availability and quality can differ across machines and sites, and data sharing is frequently impeded by privacy and security concerns. Fleet-level variability introduces further complexity. Models must generalize across equipment units that share design principles but exhibit different usage patterns or environmental exposures, often driving costly model reengineering~\citep{zeng2025ApplicationofFrequency}. These practical challenges underscore the need for models that do not rely solely on data, but instead leverage physical insights to ensure robustness and adaptability---precisely the strengths that physics-informed approaches promise to provide. Notably, the growing adoption of \ac{PIML} extends beyond \ac{PHM} into adjacent fields such as structural health monitoring \citep{rizvi2023data}, underscoring its cross-domain relevance for systems subject to degradation.

Despite the growing body of research at the intersection of \ac{PIML} and \ac{PHM}, the field remains fragmented across ways of integrating physics, across \ac{PHM} tasks, and across application domains. Existing surveys tend to focus on specific assets, such as lithium-ion batteries~\citep{meng2019Areviewon}, gas turbines~\citep{farhat2025physics}, or bridges~\citep{mammeri2025traditional}. Moreover, they tend to adopt a more exploratory approach, providing detailed insights in specific areas while leaving some aspects less systematically addressed. Given that \ac{PIML} began attracting significant attention following the seminal work by \cite{karniadakis2021physics}, the field has been developing rapidly, making it challenging for recent surveys to fully capture the current state of research. As a result, researchers lack a unified classification that enables systematic comparison of how physics is integrated across different \ac{PHM} tasks and application domains. Practitioners, in turn, have no consolidated evidence base to guide the selection of an appropriate integration strategy for a given industrial setting. This limits both the cumulative advancement of methods and their translation into operational practice.

At its core, this work addresses the question of how \ac{PIML} is being leveraged in the context of \ac{PHM}, and what the key challenges and opportunities are. To answer this systematically and ensure both reproducibility and completeness, the most comprehensive systematic literature review of \ac{PIML} in \ac{PHM} to date is conducted, covering 212 studies. The outcome provides researchers, developers, and practitioners with the necessary insights to advance data-driven \ac{PHM} applications by effectively incorporating prior physical knowledge. Specifically, the research questions this work seeks to answer are:

\begin{enumerate}
    \item \textbf{Knowledge} (a) What types of prior physical knowledge are being leveraged, and (b) what forms of representation are employed?
    \item \textbf{Incorporation} (a) How can prior physical knowledge be incorporated, and (b) how does the form of representation influence which approaches to incorporation are feasible?
    \item \textbf{Practice} (a) How does incorporating prior physical knowledge help overcome limitations of purely data-driven methods, and (b) what are the primary challenges in developing and applying physics-informed approaches?
\end{enumerate}

% Structure of the paper
The remainder of this work is organized as follows. Section~\ref{sec:background} provides background on \ac{PHM}, physics-informed learning, and a brief description of related work. Section~\ref{sec:methodology} describes the methodology underlying the systematic literature review, including the search strategy, as well as the screening and quality assessment procedures. Section~\ref{sec:literature_review} presents the four-class classification scheme, including observational bias, inductive bias, learning bias, and hybrid approaches. Subsequently, the identified studies are summarized. Section~\ref{sec:discussion} analyzes methodological trends and discusses current challenges and opportunities for future advancements in industrial \ac{PIML}-based \ac{PHM}. Finally, Section~\ref{sec:conclusion} concludes with a synthesis of key insights and implications, concisely answering the outlined research questions.

% - - - - - - - - - - section - - - - - - - - - - %
\section{Background}
\label{sec:background}

% Outlook of the section to follow
The following section outlines the theoretical foundation necessary to understand the key concepts in this review. \ac{PHM} is introduced first, covering its core tasks, as well as various approaches to its implemen\-tation. Among these, hybrid approaches are particularly promising, with \ac{PIML} at the forefront. This paradigm is defined and explored, along with related research fields. Finally, an overview of related work is provided to position this review within the broader literature.

% - - - - - - - - - - subsection - - - - - - - - - - %
\subsection{Prognostics and Health Management}
\label{subsec:prognostics_and_health_management}

% Introduction and brief definition of PHM
\ac{PHM} is an engineering discipline that focuses on detecting, isolating, and diagnosing potential faults in a system, assessing its current \ac{SOH}, and predicting its \ac{RUL} with the aim of preventing unplanned downtimes, enhancing reliability, and supporting additional system-level objectives. These diagnostic and prognostic measures play a crucial role in ensuring continuous operation by monitoring system health and predicting incipient faults before they progress into catastrophic failures. Hence, \ac{PHM} not only improves operational efficiency but also enhances the safety of the monitored system. However, implementing \ac{PHM} presents several challenges that vary depending on the chosen approach for modeling the system. While the adoption of \ac{PHM} in industrial settings holds significant potential, it also requires addressing various technical and organizational obstacles to fully realize its benefits. With \ac{PHM} encompassing both diagnostics and prognostics, four essential tasks are typically delineated for implementation \citep{hagmeyer2022integration, jia2018review}:

% PHM tasks
\begin{enumerate}
    \item \textbf{Fault detection} is aimed at determining the presence or absence of a fault.
    \item \textbf{Diagnosis} is aimed at attributing observed faults to their root causes. 
    \item \textbf{Health assessment} is aimed at estimating the system's current \ac{SOH} or risk of failure.
    \item \textbf{Prognosis} is aimed at predicting the future development of the \ac{SOH} or \ac{RUL}.
\end{enumerate}

\begin{table}[t]
    \centering
    \caption{Key strengths and limitations of physical model-based and purely data-driven approaches \citep{karniadakis2021physics, baur2020review}.}
    \label{tab:physics_and_data_approaches}
    \begin{tabular}{
        >{\centering\arraybackslash}m{0.03\linewidth}
        >{\raggedright\arraybackslash}p{0.4\linewidth} 
        >{\raggedright\arraybackslash}p{0.4\linewidth} 
    }
    \hline
    & \textbf{Physical Model-Based Approaches} & \textbf{Purely Data-Driven Approaches} \\
    \hline

    \rotatebox{90}{\parbox{1.5cm}{\centering Strengths}} & 
    % strengths model-based methods
    \begin{itemize}[leftmargin=*, nosep]
        \vspace{-21pt}
        \raggedright    
        \item Accurate and reliable 
        \item Interpretable
        \item Robust
        \vspace{-48pt}
    \end{itemize} &
    % strengths data-driven methods
    % \vspace{-12pt}
    \begin{itemize}[leftmargin=*, nosep]
        \vspace{-21pt}
        \raggedright
        \item Generalizable
        \item Scalable
        \item Low imple\-men\-tation effort
        \vspace{-48pt}
    \end{itemize} \\
    
    \hline
    
    \rotatebox{90}{\parbox{2.5cm}{\centering Limitations}} & 
    % limitations model-based methods
    \begin{itemize}[leftmargin=*, nosep]
        \vspace{-36pt}
        \raggedright
        \item Requires complete system knowledge
        \item High implemen\-tation effort
        \item Highly system-specific
        \vspace{-0pt}
    \end{itemize} &
    % limitations data-driven methods
    
    \begin{itemize}[leftmargin=*, nosep]
        \vspace{-36pt}
        \raggedright
        \item Relies on rep\-re\-sen\-tative and sufficient data
        \item Less interpretable
        \item Poor extrapolation capability
        \item Risk of implausible predictions
        \vspace{-0pt}
    \end{itemize} \\
    
    \hline
    \end{tabular}
\end{table}

% Approaches regarding PHM 
To carry out these tasks, a range of methodological approaches is used, varying in the extent to which they rely on physical knowledge, data-driven insights, or a combination of both. These approaches are generally subdivided into three distinct categories \citep{kim2017prognostics, atamuradov2017prognostics}: 

\begin{itemize}
    \item Physical model-based approaches
    \item Purely data-driven approaches
    \item Hybrid approaches
\end{itemize}

% Model-based, data-driven and hybrid approach
According to~\cite{gouriveau2016prognostics}, physical model-based approaches \textquote{require the construction of a dynamic model representing the behavior of the system and integrating the degradation mechanism (mainly by models of fatigue, wear, or corrosion), whose evolution is modeled by a deterministic law or by a stochastic process.} In contrast, purely data-driven approaches (including statistical and \ac{ML} methods) leverage monitoring data---either directly or via extracted features---to model a system’s behavior and health state \citep{goodman2019prognostics}. Hybrid approaches, which integrate aspects of the aforementioned methods, seek to leverage their strengths while mitigating their individual limitations (see Tab.~\ref{tab:physics_and_data_approaches}). Combining the accuracy and robustness of physical models with the flexibility and adaptability of data-driven models enables enhancing the overall performance of the respective solution. The aim is to capitalize on the synergies between these approaches, ultimately providing a more comprehensive and effective solution that can accommodate diverse operational scenarios. Nonetheless, this does not preclude scenarios in which a physics-based or data-driven approach is preferable.

% Lacking consensus regarding the definition of hybrid approaches and the trend towards integrating prior knowledge into data-driven methods
Definitions of hybrid approaches may vary, however, particularly in terms of the extent to which physics is included. Certain hybrid approaches employ complete physical models, while others rely on prior physical knowledge that may be insufficient for holistic modeling. The choice of approach is generally dictated by the underlying physics of the problem, the availability of relevant data, and specific requirements imposed by the intended solution. The continuous automation of modern industrial machinery leads to increasingly complex degradation processes, which are often poorly understood, dynamic, and highly nonlinear \citep{zio2022prognostics}. As a consequence, high-fidelity physics-based modeling is becoming increasingly difficult, if not impossible. Accordingly, the integration of (partial) physical knowledge into data-driven methods is gaining momentum.

% - - - - - - - - - - subsection - - - - - - - - - - %
\subsection{Physics-Informed Machine Learning}
\label{subsec:physics_informed_machine_learning}

\begin{figure*}
    \centering
    \includegraphics[width=\textwidth]{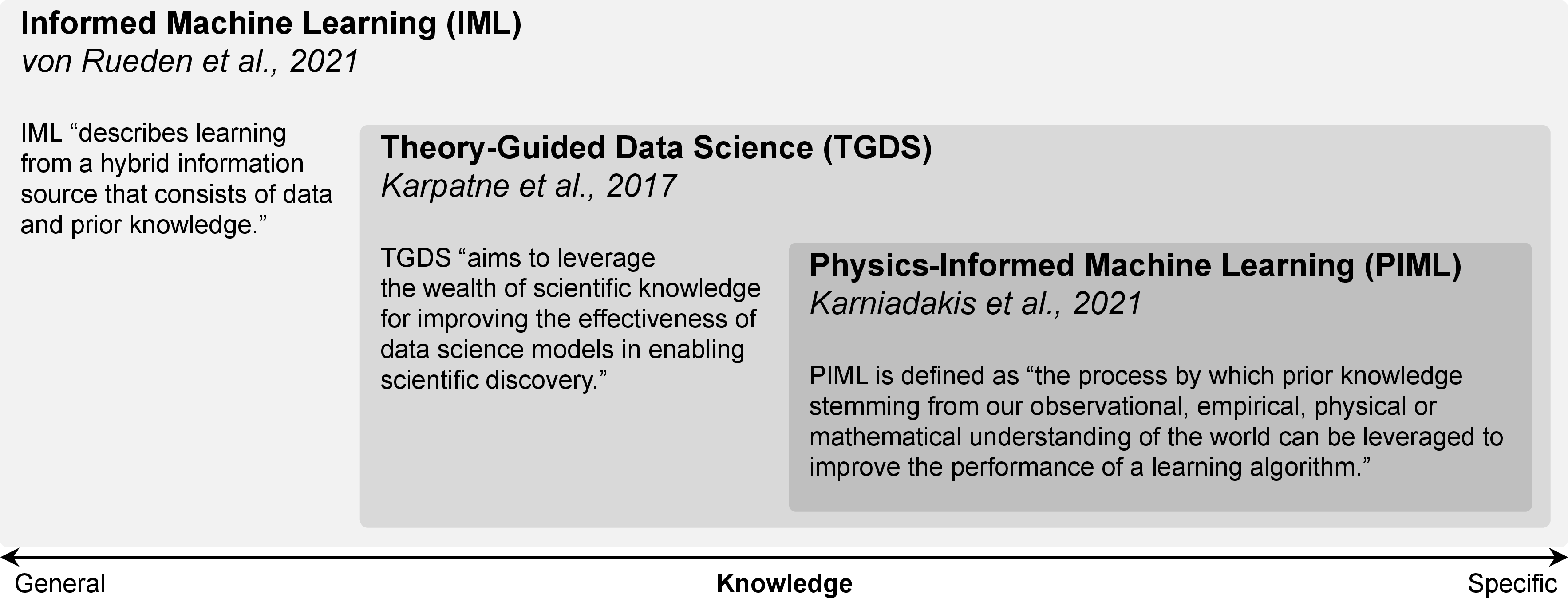}
    \caption{\ac{IML}~\citep{vonrueden2021informed} provides a foundational framework for integrating various forms of prior knowledge into \ac{ML}. Within this context, \ac{TGDS}~\citep{karpatne2017theory} focuses specifically on incorporating scientific knowledge, while \ac{PIML}~\citep{karniadakis2021physics} narrows the scope even further to leverage prior physical knowledge.}
    \label{fig:research_fields}
\end{figure*}

% Introduction of PIML and related research fields
The incorporation of prior knowledge alongside empirical data constitutes a central paradigm in \ac{ML} research (see Fig.~\ref{fig:research_fields}). \cite{vonrueden2021informed} provide a general conceptualization of learning from such hybrid information sources, thereby establishing the foundational framework of \ac{IML}. According to this framework, prior knowledge is expected to originate from independent sources, be formally represented, and be explicitly integrated into the learning process. \cite{karpatne2017theory} propose a related framework with a specific focus on scientific knowledge, referred to as \ac{TGDS}. Narrowing the focus further, \cite{karniadakis2021physics} introduce \ac{PIML} as a means to improve the modeling of physical systems by embedding prior physical knowledge directly into \ac{ML} models. 

% Brief description of PIML
Instead of relying solely on data, \ac{PIML} incorporates physics such as governing differential equations, conservation laws, or symmetries to ensure that model predictions remain physically consistent, which is especially valuable in regimes where data are sparse or noisy \citep{karniadakis2021physics}. This makes \ac{PIML} particularly attractive in scientific and engineering domains, where high-fidelity measurements can be expensive or limited by experimental feasibility. Operationally, \ac{PIML} can be implemented through a variety of complementary strategies, including data-centric approaches, design-level interventions, and regularization-based constraints. Yet developing \ac{PIML} remains a nontrivial task, involving challenges such as balancing data fidelity with physics constraints, managing computational costs, and ensuring efficient training \citep{jahani2024enhancing}. Nonetheless, \ac{PIML} has become a prominent paradigm for developing models that are both data-driven and physically informed, balancing accuracy, interpretability, and robustness.

% Principles of physics-informed learning: introduction of observational, inductive and learning biases
Three main pathways have been defined for embedding physics into \ac{ML} models, following the principles outlined by \cite{karniadakis2021physics}. Each is characterized by the introduction of an appropriate bias:

\begin{itemize}
    \item \textbf{Observational bias} can be introduced directly through the training data.
    \item \textbf{Inductive bias} can be introduced by tailored interventions to the model design.
    \item \textbf{Learning bias} can be introduced through modifications to the learning algorithm.
\end{itemize}

% Tangible explanation of the effects of these biases on learning
The hypothesis space provides a useful lens for understanding how these biases affect learning. It denotes the set of all functions a model could, in principle, choose to map inputs to outputs; for example, all functions realizable by a \ac{NN} with a given architecture. Observational bias, stemming from the training data, does not alter the space itself. It merely biases the training process toward functions that better fit the observed data, leaving the set of representable functions intact. Inductive bias, in contrast, acts directly on the model design and explicitly shapes the hypothesis space. Embedding symmetry constraints, conservation laws, or other physical principles into the model restricts the hypothesis space to functions consistent with these principles, actively enforcing physical plausibility. Learning bias exerts a more subtle influence: choices in optimization algorithms, regularization, or training strategies such as early stopping guide the search toward certain solutions, making some regions more likely to be explored while leaving the space itself unchanged. In summary, observational bias influences the selection of hypotheses within the existing space (not imposing constraints), inductive bias modifies the structure of the space itself (imposing hard constraints), and learning bias guides the learning process toward particular regions, thereby effectively prioritizing certain solutions over others (imposing soft constraints).

% - - - - - - - - - - subsection - - - - - - - - - - %
\subsection{Related Work}
\label{subsec:related_work}

% Short overview of related work
Extensive reviews have emerged around both \ac{PHM}~\citep{tsui2015prognostics, atamuradov2017prognostics, hu2022prognostics, zio2022prognostics} and \ac{PIML}~\citep{karniadakis2021physics, cai2021physics, cuomo2022scientific} individually. However, while both fields have advanced significantly, the synergy between them is increasingly recognized as crucial for addressing the intricate challenges of effectively managing system health. Accordingly, several reviews have sought to synthesize the state of research at the intersection of \ac{PHM} and \ac{PIML}, among which the following were known to the authors prior to conducting this systematic literature review: \cite{deng2023Physicsinformedmachinelearning}, \cite{kundu2020Areviewon}, and the review by~\cite{meng2019Areviewon}---all of which were also identified through the systematic approach employed in this work. Hence, a detailed description is omitted here, since these (along with eight additional reviews) will be examined in detail in Section~\ref{subsec:reviews}. Furthermore, a broader perspective on \ac{PIML} in the context of intelligent manufacturing is provided by~\cite{leng2026physics}.

% - - - - - - - - - - section - - - - - - - - - - %
\section{Methodology}
\label{sec:methodology}
\begin{table*}[t]
\centering
\caption{Keywords used to identify the initial set of records. A wildcard operator (*) accounts for morphological variations.}
\label{tab:keywords}
\begin{tabular}{p{2cm} p{13.2cm}}
\hline
\textbf{Category} & \textbf{Keywords} \\
\hline

\raggedright Prior Physical Knowledge &
(\enquote{physics-inform*} OR \enquote{physics-infus*} OR \enquote{physics-guid*} OR \enquote{physics-induc*} OR \enquote{physics-constrain*} OR \enquote{physics-promot*} OR \enquote{physics-enhanc*} OR \enquote{physics-inherit*} OR \enquote{physics-aware} OR \enquote{physics-based} OR \enquote{theory-guid*} OR \enquote{algebraic equation*} OR \enquote{differential equation*} OR \enquote{ODE*} OR \enquote{PDE*} OR \enquote{conservation law*} OR \enquote{spatial invariance*} OR \enquote{probabilistic relation*} OR \enquote{simulat* data} OR \enquote{prior knowledge}) \\

% \rowcolor{gray!20}
& AND \\

\raggedright Machine Learning 
& (\enquote{artificial intelligence} OR \enquote{pattern recognition} OR \enquote{data-driven} OR \enquote{machine learning} OR \enquote{deep learning} OR \enquote{scientific machine learning} OR \enquote{hybrid machine learning} OR \enquote{supervised learning} OR \enquote{unsupervised learning} OR \enquote{self-supervised learning} OR \enquote{reinforcement learning} OR \enquote{active learning} OR \enquote{automated learning} OR \enquote{meta-learning} OR \enquote{transfer learning} OR \enquote{few-shot learning} OR \enquote{multi-task learning} OR \enquote{ensemble learning} OR \enquote{probabilistic modeling} OR \enquote{sequence modeling} OR \enquote{time series forecasting} OR \enquote{Markov model*} OR \enquote{hidden Markov model} OR \enquote{Gaussian process*} OR \enquote{support vector machine*} OR \enquote{support vector regression} OR \enquote{decision tree*} OR \enquote{random forest*} OR \enquote{k-nearest neighbors} OR \enquote{k-means clustering} OR \enquote{hierarchical clustering} OR \enquote{Gaussian mixture model} OR \enquote{gradient boosting} OR \enquote{boosting algorithms} OR \enquote{AdaBoost} OR \enquote{XGBoost} OR \enquote{LightGBM} OR \enquote{CatBoost} OR \enquote{neural network*} OR \enquote{long short-term memory} OR \enquote{gated recurrent unit} OR \enquote{autoencoder} OR \enquote{generative adversarial network*} OR \enquote{Bayesian network*} OR \enquote{transformer*} OR \enquote{foundation model} OR \enquote{CNN*} OR \enquote{RNN*} OR \enquote{LSTM*} OR \enquote{pretext task} OR \enquote{downstream task} OR \enquote{Q-learning} OR \enquote{deep Q-network} OR \enquote{policy gradient methods} OR \enquote{actor-critic methods} OR \enquote{model-free learning} OR \enquote{model-based learning}) \\

% \rowcolor{gray!20}
& AND \\

\raggedright Prognostics and Health Management &
(\enquote{prognostics and health management} OR \enquote{PHM} OR \enquote{diagnos*} OR \enquote{prognos*} OR \enquote{health assessment} OR \enquote{health management} OR \enquote{health monitoring} OR \enquote{condition monitoring} OR \enquote{predictive maintenance} OR \enquote{remaining useful life} OR \enquote{RUL} OR \enquote{state of health} OR \enquote{end of life} OR \enquote{degrad*} OR \enquote{fault isolation} OR \enquote{fault detect*} OR \enquote{anomaly detect*} OR \enquote{damage detect*} OR \enquote{failure detect*} OR \enquote{failure prediction}) \\

% \rowcolor{gray!20}
\centering & AND NOT \\

\raggedright Healthcare &
(\enquote{healthcare} OR \enquote{medic*} OR \enquote{clinic*} OR \enquote{nursing} OR \enquote{cancer*} OR \enquote{illness*} OR \enquote{disease*} OR \enquote{sickness}) \\

\hline
\end{tabular}
\end{table*}
% Outlook of the section to follow
The following section describes the methodology employed to conduct the systematic literature review on \ac{PIML} in \ac{PHM}. Starting from the research questions formulated in Section~\ref{sec:introduction}, the process encompasses the selection of relevant keywords and databases, a rigorous screening procedure to identify all relevant studies, and a quality-based refinement to distill the final selection. By systematically analyzing the literature, this work provides a solid foundation for synthesizing current physics-informed approaches to diagnostics and prognostics, identifying methodological gaps, and guiding future research directions.

% - - - - - - - - - - subsection - - - - - - - - - - %
\subsection{Keywords}
\label{subsec:keywords}

% Categories of keywords, their purpose and how they were gathered
The keywords used to identify the initial set of records are listed in Table~\ref{tab:keywords}. In total, 105 unique terms were defined across four categories: \textit{prior physical knowledge}, \textit{machine learning}, \textit{prognostics and health management}, and \textit{healthcare}. The first three categories ensure relevance to the scope of this work. The last category serves as an implicit filter to rule out irrelevant healthcare-related records resulting from ambiguities in \ac{PHM}-related keywords.

% Identification and refinement of keywords
Relevant keywords were identified by examining the research questions' core concepts and corresponding synonyms. This also included screening prominent studies explicitly addressing the scope of this review for their author-assigned keywords, employing large language models to generate additional keyword suggestions, and consulting peers to review and validate the keyword collection. Furthermore, keywords were iteratively refined by performing exploratory searches in the selected databases, enabling the identification and exclusion of keywords of lower importance or higher ambiguity (e.g., \enquote{energy} or \enquote{force}
within the category of \textit{prior physical knowledge}; \enquote{fault} or \enquote{failure} within the category of \textit{prognostics and health management}). This approach ensured that the final keyword set was both precise and comprehensive.

% - - - - - - - - - - subsection - - - - - - - - - - %
\subsection{Databases}
\label{subsec:databases}

% Selection of databases and rationale on why arXiv was incorporated
Scopus and Web of Science (WoS) were selected as the primary databases due to their comprehensive coverage of peer-reviewed literature across various disciplines, ensuring a robust and thorough search. Additionally, the preprint server arXiv was included to capture the most recent research developments. This decision was made to provide a more accurate depiction of the current state of research, acknowledging that many cutting-edge studies are first disseminated through preprints before formal publication, especially in the realm of \ac{ML}. 

% Rationale regarding the time period covered
\ac{TGDS}, a precursor to \ac{PIML}, was formally established around 2017 (as stated in Sec.~\ref{subsec:physics_informed_machine_learning}), marking a significant milestone in integrating scientific knowledge with \ac{ML} techniques. However, the search covered the period from 2012 onward---a year widely recognized within the \ac{ML} community as a pivotal moment due to the breakthrough results of AlexNet~\citep{AlexNet} in the ImageNet competition. This choice reflects the consensus that 2012 represents the inception of modern \ac{ML} and \ac{DL}~\citep{mienye2024comprehensive}. Yet the search on arXiv was limited to the period from 2023 to the present, under the assumption that high-quality studies submitted to arXiv before 2023 would have already been published in peer-reviewed journals or conference proceedings.

% Explanation on how search strings were built
The complete search string was constructed by connecting the keywords of the various categories with the appropriate logical operators (see Tab.~\ref{tab:keywords}). The syntax for the logical operators was adapted to the specific requirements of each database. The fields examined for relevant records included the title, abstract, and author-assigned keywords. For arXiv, which does not provide searching capabilities for author-assigned keywords, only the title and abstract were considered. All keywords were enclosed in quotation marks in order to ensure exact phrases in the search queries. Moreover, both hyphenated and non-hyphenated variants of keywords were automatically retrieved by the databases. 

% - - - - - - - - - - subsection - - - - - - - - - - %
\subsection{Screening Procedure}
\label{subsec:screening_procedure}

% The purpose of using ASReview
The screening procedure was aided by ASReview~\citep{asreview}, which is an \ac{ML}-based tool designed to streamline the systematic review process. It leverages active learning algorithms to prioritize the most relevant records from a large corpus, significantly reducing the manual effort required for initial screening. Records are presented iteratively, with their order continually updated based on reviewer feedback (researcher-in-the-loop) to facilitate the identification of relevant records. Additionally, ASReview supports bias-free screening with respect to author names, affiliations, and other details by providing only the title and abstract of each record.

% Five-step process (from constructing a dataset to obtaining the final set of relevant records)
ASReview was employed to manage the substantial number of records initially retrieved from the searched databases. A structured approach to screening was implemented, consisting of the following steps:

\begin{enumerate}
    \item Constructing a dataset containing the records to screen.
    \item Specifying inclusion and exclusion criteria to determine whether records are considered relevant.
    \item Defining a stopping criterion to determine at which point screening will conclude.
    \item Conducting screening in two cycles:
    \begin{enumerate}[label=4.\arabic*]
        \item Configuring the active learning algorithm and providing a subset of records for its initial training.
        \item Performing the screening based on the specified criteria until the stopping criterion is met.
    \end{enumerate}
    \item Extracting the records identified as relevant to compile the subset of records for further analysis.
\end{enumerate}

% Constructing a dataset and defining inclusion/exclusion criteria 
The set of records initially retrieved from the designated databases constitutes the dataset, with all duplicates removed. Only records containing both a title and an abstract are retained. To determine the relevance of identified records, inclusion and exclusion criteria are specified. The inclusion criterion requires that a record substantively addresses all three areas: prior physical knowledge, \ac{ML}, and \ac{PHM}. While the search string (see Tab.~\ref{tab:keywords}) ensures that matching terms appear in the title, abstract, or author-assigned keywords of every retrieved record, syntactic matching alone is insufficient to guarantee genuine topical relevance---records may, for instance, reference related concepts in a negating or merely peripheral context. The reviewer therefore objectively assesses whether all three areas are adequately addressed in substance, rather than relying solely on keyword occurrence. Regarding the exclusion criterion, a record is excluded if the studied system, component, material, or process is not investigated with respect to degradation during operation, as diagnostics and prognostics fundamentally rely on the asset’s current health state. The scope of eligible assets further excludes transportation systems and their infrastructure, unmanned aerial vehicles, consumer electronics, and other technical systems used in residential settings. 

% Defintion of the stopping criterion
\cite{asreview} state that, based on their simulation studies, reviewing 8--33\,\% of the total number of records using ASReview is typically sufficient to identify 95\,\% of the relevant records. Based on these findings, the stopping criterion is defined as follows: at least 8\,\% of the records must be reviewed, and screening is terminated at the latest once 33\,\% have been screened. Within these bounds, screening is considered complete when 50 consecutive irrelevant records are encountered---a heuristically set threshold.

% Explanation of the two-stage screening process
Screening comprises two cycles, with the first cycle leveraging a rather simple but fast configuration of the active learning algorithm. The minimum requirement for labeled training data is to include at least one record labeled as relevant and one as irrelevant. The aim during this cycle is to gather all relevant records that are more easily distinguishable from the rest of the dataset. At a certain point, however, this configuration reaches its limitations, and inevitably, the stopping criterion is met. Nevertheless, it can be assumed that the dataset still contains relevant records, though identifying them will require a more nuanced approach. This necessitates a second cycle of screening, where a more sophisticated model is employed to capture all remaining relevant records that are more difficult to detect. These records often elude initial screening due to subtle semantic nuances and complex contextual variations that a simpler model might overlook. The result (i.e., all labeled records) of the first cycle serves as the initial training data for the more complex model. Upon reaching the stopping criterion for the second time, it is assumed that a representative subset of relevant records has been identified.

% Explanation of the reviewers part during screening
Two reviewers screened records independently in alternating one-hour sessions. Records of debatable relevance were discussed and decided upon through mutual consultation. Periodic discussions ensured a shared understanding and consistent application of the criteria outlined earlier.

% - - - - - - - - - - subsection - - - - - - - - - - %
\subsection{Quality-Based Refinement}
\label{subsec:quality_based_refinement}

\begin{table*}[t]
\centering
\caption{Conferences considered for the quality-based refinement of conference papers (in alphabetical order).}
\label{tab:conferences}
\begin{tabular}{p{1.8cm} p{13.4cm}}
\hline
\textbf{Acronym} & \textbf{Conference} \\
\hline
CMS    & CIRP Conference on Manufacturing Systems \\
ESREL  & European Safety and Reliability Conference \\
ETFA   & IEEE International Conference on Emerging Technologies and Factory Automation \\
ICPHM  & IEEE International Conference on Prognostics and Health Management \\
INDIN  & IEEE International Conference on Industrial Informatics \\
PHMSC  & Annual Conference of the PHM Society (including its counterparts in Europe and Asia) \\
RAMS   & International Conference on Reliability and Maintainability \\
SMC    & IEEE International Conference on Systems, Man and Cybernetics \\
\hline
\end{tabular}
\end{table*}

% Refining the collection of identified records (journals)
After the screening procedure is completed, the collection of identified records is refined by retaining only those that meet specified quality criteria. With respect to journal articles, the quality-based refinement depends on both the impact factor (taken from the Journal Citation Reports provided by \cite{clarivate}) and the SCImago Journal Rank (SJR) (provided by \cite{scimago}).
The former is required to be 3 or greater, while the SJR is required to be Q2 or better. If a journal is assigned to multiple categories within the SJR, it must maintain a minimum ranking of Q2 across all categories. Both the impact factor and SJR are referenced according to the publication year of the article, with the most recent available values used when the corresponding year’s metrics are not yet released. An article will still be considered if only one of the two metrics is available, provided the journal meets the required standard. Articles from journals for which neither metric is available are excluded from the final selection.

% Refining the collection of identified records (conferences)
For conference papers, no established metric or ranking system with broad interdisciplinary applicability comparable to those used for journals exists. Consequently, the quality-based refinement for these records is applied differently. Only papers presented at conferences recognized for their relevance and impact in the field of \ac{PHM} are considered (see Tab.~\ref{tab:conferences}).

% Handling of arXiv papers and other forms of publication
Contributions from the preprint server arXiv cannot undergo quality-based refinement. Accordingly, arXiv papers are incorporated into the final selection of relevant records without additional assessment if they are identified as relevant during the screening procedure. In contrast, other forms of publication, such as book chapters or technical reports, are excluded due to the lack of a reliable strategy for assessing their academic rigor and impact.

% - - - - - - - - - - subsection - - - - - - - - - - %
\subsection{Results}
\label{subsec:results}

\begin{figure}[t]
    \includegraphics[width=\linewidth]{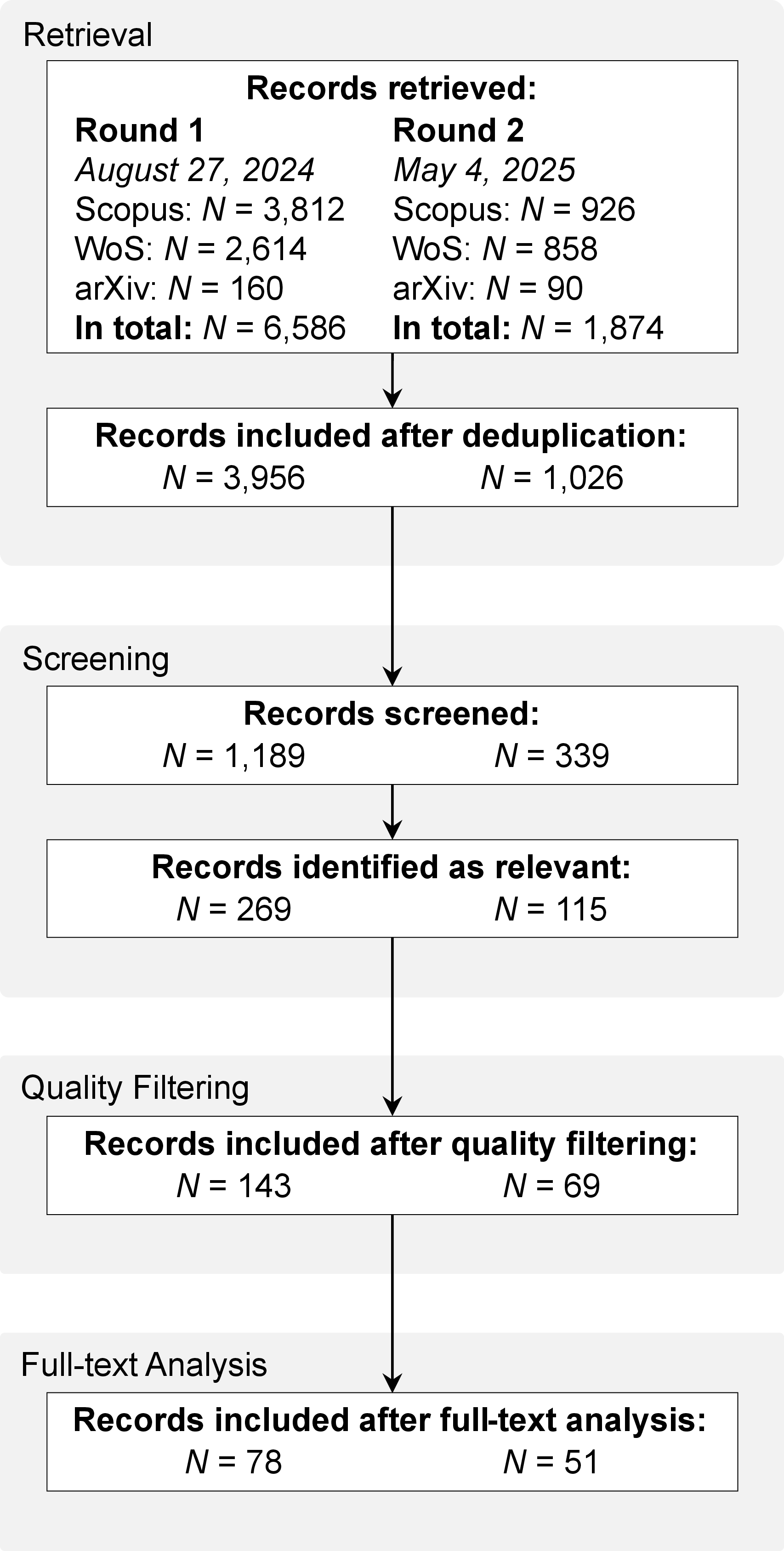}
    \caption{This flowchart illustrates the key stages of the methodology and the corresponding number of records at each stage for both rounds.}
    \label{fig:flowchart}
\end{figure}

% Retrieval of potentially relevant records
The methodology was applied twice (referred to as two rounds), with database searches conducted on each occasion to systematically gather relevant records and provide an up-to-date depiction of the state of research. This approach revealed a notable increase in publications over time, reflecting heightened research activity and growing interest at the intersection of \ac{PIML} and \ac{PHM}. Among the sources, Scopus yielded the highest number of records, followed by WoS, while arXiv produced the fewest due to the more restricted time frame applied to that search. While the period from January 2012 to August 27, 2024, yielded 6,586 records, the subsequent period from August 27, 2024, to May 4, 2025, produced 1,874 records. This indicates a marked acceleration in contributions, with nearly 30\,\% of the previous 12.5 years’ output occurring within this brief interval. After retrieval, these records were preprocessed, resulting in a reduction to 3,956 and 1,026 records, respectively. Thereafter, systematic screening and filtering yielded the final set, which formed the basis for analyzing the most current and relevant literature on \ac{PIML} in \ac{PHM} (see Fig.~\ref{fig:flowchart}). 

% Configuration of the active learning algorithms
The two rounds comprised a total of three cycles, each utilizing a specifically configured active learning algorithm within the ASReview framework. In the first cycle of round one, a comparatively simple yet computationally efficient configuration was adopted. Feature extraction was performed using \textit{term frequency-inverse document frequency}, and a Naive Bayes classifier served as the underlying model for prioritizing records. The query strategy and the balancing strategy were set to \textit{mixed} and \textit{dynamic resampling}, respectively. Upon reaching the stopping criterion, a more sophisticated configuration was introduced for the second cycle. Specifically, feature extraction was switched to \textit{sBERT}, which produces contextually richer embeddings, and the classifier was replaced by an \ac{NN}, while the query and balancing strategies remained unchanged. The second round of screening (cycle three) adopted the same advanced configuration as cycle two, leveraging all previously labeled records from the first round as initial training data.

% Initial training of the first active learning algorithm
In preparation for the first cycle, 20 labeled records were provided for initialization of the active learning algorithm. Although the minimum requirement is only two labeled records, this provided the model with a more informative starting point. Records initially labeled as relevant were not guaranteed inclusion in the final review, as some may have been excluded during quality-based refinement or full-text analysis. The initial labeled set comprised the following records:

\begin{itemize}
    \item Ten relevant records: \cite{sun2018AHybridApproach, ma2024Accurateandefficient, ellis2022Ahybridframework, chen2022PhysicsInformedLSTMhyperparameters, badora2023Usingphysicsinformedneural, yucesan2019Windturbinemain, garpelli2023Physicsguidedneuralnetworks, gareev2021Improvedfaultdiagnosis, gurgen2022Developmentandassessment, zhou2023PHYSICSINFORMEDMACHINELEARNING}
    \item Ten irrelevant records: \cite{gao2021sensing, he2024training, gong2022deep, pan2019aqlearning, sajedi2023twin, mcmahon2024river, chen2020augmenting, hu2017semi, zjavka2017nwp, wanasundara2023detecting}
\end{itemize}

% Screening
Two reviewers assessed a total of 1,189 (30.1\,\%) and 339 (33.0\,\%) records for relevance during the two rounds of screening, respectively (see Tab.~\ref{tab:screening_results}). Screening was performed in an alternating fashion, requiring 43 iterations overall. During round one (i.e., cycle one and two), screening was terminated upon encountering 50 consecutive irrelevant records, in accordance with the stopping criterion, whereas round two concluded automatically upon reaching the maximum screening threshold of 33\,\%. The difference in the total number of records screened per reviewer can be attributed to several factors, such as encountering longer abstracts, varying complexity in determining whether all three areas (prior physical knowledge, \ac{ML}, and \ac{PHM}) were substantively addressed, and varying cognitive load. The screening yielded a total of 384 records across both rounds, a number considered excessive even for a thorough review.

% Final selection of studies 
The application of the quality-based refinement effectively reduced the number of records. The final set comprised 212 records, including 180 journal articles, 18 conference papers (ESREL, ICPHM, PHMSC, and RAMS), and 14 preprints sourced from arXiv. Accordingly, this work constitutes the most comprehensive review to date in this area of research. All 212 records were subsequently subjected to full-text analysis, after which 83 records were excluded due to insufficient alignment with the scope of this review. All exclusions were transparently documented with their corresponding rationale (see \nameref{appendix}). Hence, 129 records will be presented and discussed in this review.

\begin{table*}[t]
\centering
\caption{This table provides a detailed overview of the screening results across all three cycles, where the first two authors correspond to reviewer A and B (in no particular order).}
\label{tab:screening_results}
\begin{tabular}{ p{2.9cm}  p{2.1cm} p{2.1cm} p{2.1cm} p{2.1cm} p{2.1cm} }
\hline
& \textbf{Round 1} & & & \textbf{Round 2} & \\
\hline
& Cycle 1 & Cycle 2 & \raggedright In total \mbox{(Round 1)} & Cycle 3 & {\raggedright \mbox{In total} \mbox{(Round 1+2)}} \\
\hline

Iterations & 18 & 17 & 35 & 8 & 43 \\
\raggedright Records \mbox{screened} & 613 & 576 & 1189 & 339 & 1528 \\
\hspace{5mm} Reviewer A & 372 & 346 & 718 & 174 & 892 \\
\hspace{5mm} Reviewer B & 241 & 230 & 471 & 165 & 636 \\
\raggedright Relevant records & 149 & 120 & 269 & 115 & 384 \\ 
\raggedright Irrelevant records & 464 & 456 & 920 & 224 & 1144 \\ 

\hline
\end{tabular}
\end{table*}

% - - - - - - - - - - subsection - - - - - - - - - - %
\subsection{Limitations}
\label{subsec:limitation}

% Limitations regarding ASReview
Given that a reproducible systematic literature review depends on transparency, all potential limitations related to the outlined methodology are clearly articulated, mainly concerning the aided screening procedure and the subsequent quality-based refinement. Screening a large number of records required incorporating an additional tool---ASReview---aimed at facilitating the identification of those that are actually relevant. Although ASReview relies on iterative reviewer feedback, semantic nuances may still pose challenges for the underlying active learning algorithms. As a result, some relevant records may not have been surfaced by the algorithm before the stopping criterion was met. This constitutes an inherent limitation of active learning-based screening that cannot be fully eliminated without exhaustive manual review of the entire corpus of nearly 5,000 records. However, the stopping criterion, defined in accordance with~\cite{asreview}, inherently implies that approximately 5\,\% of relevant studies may remain unidentified by design.

% Limitations regarding the quality-based refinement
The decision to apply a quality-based refinement was partly driven by the substantial number of records identified as relevant during screening. The set of 384 records was considered impractical for full review, necessitating a reduction to a more manageable number. Retaining only high-quality studies not only reduced the number of records but also enhanced both the relevance of the findings and the reliability of the conclusions drawn. Nonetheless, the thresholds for the impact factor and SJR were established based on the authors' expert judgment, as was the selection of high-impact conferences, which may be considered a limitation of the review.

% - - - - - - - - - - section - - - - - - - - - - %
\section{Literature Review}
\label{sec:literature_review}

% Outlook on the section to follow
The following section provides a synthesis of all studies included in the systematic literature review, classified according to their methodological approach to combining physics and \ac{ML}. The classification scheme adopts and further refines the three pathways outlined by~\cite{karniadakis2021physics}, imposing a more stringent framework. Additionally, hybrid approaches are recognized as a distinct class, with the rationale discussed in detail below. After outlining the classification scheme, a summary of the included review studies is presented, emphasizing the motivation for conducting the current review. This is followed by a detailed depiction of the current state of research based on the remaining studies. 

% - - - - - - - - - - subsection - - - - - - - - - - %
\subsection{Classification}
\label{subsec:classification}

\begin{figure*}[h]
    \centering
    \includegraphics[width=\textwidth]{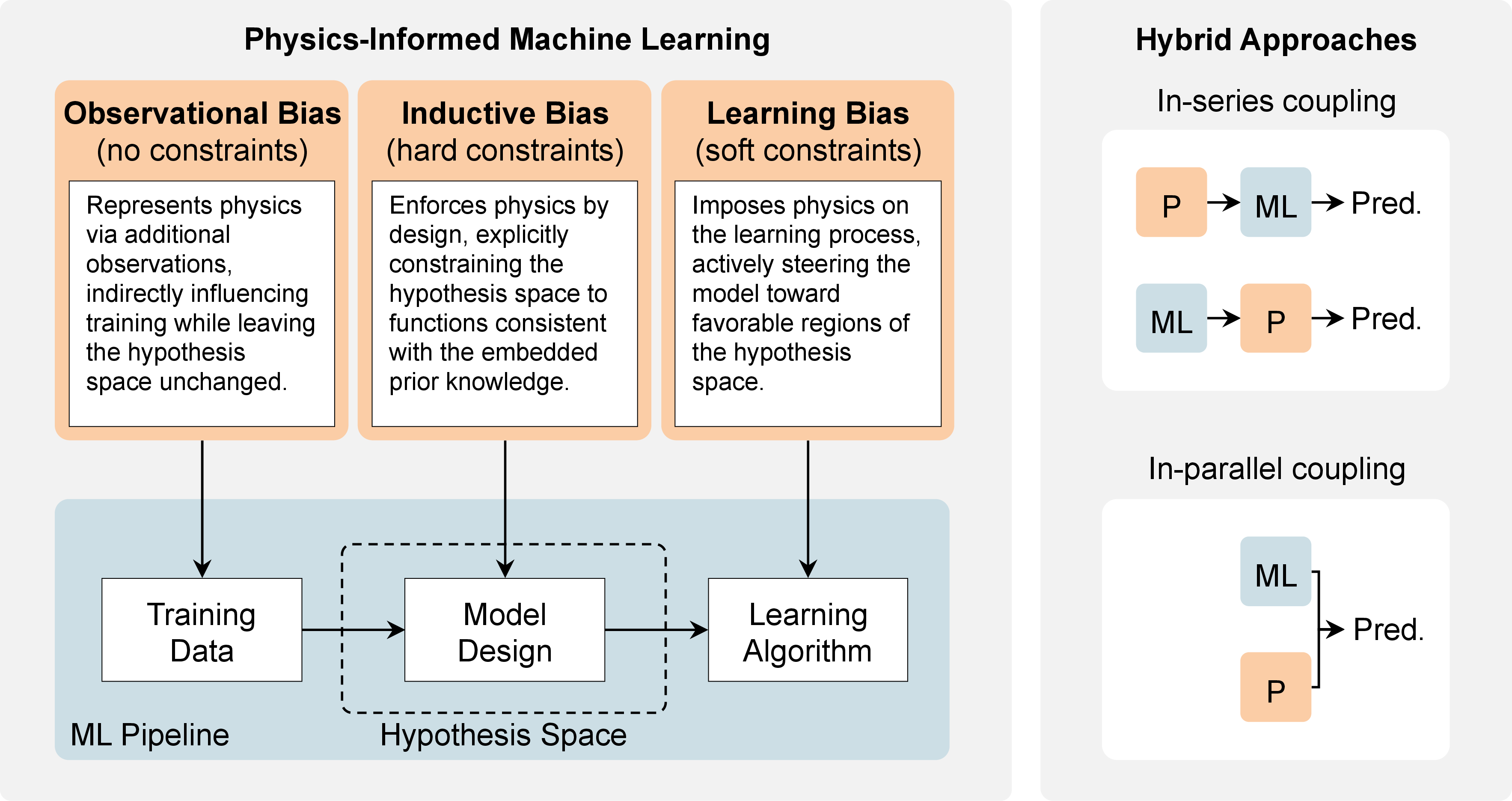}
    \caption{The classification scheme underlying this review comprises four classes, with the first three classes representing genuine physics-informed approaches and the fourth capturing hybrid approaches.}
    \label{fig:classification_scheme}
\end{figure*}

% General overview, including a "foreshadowing" regarding the separate consideration of PIML and hybrid approaches
Approaches to implementing \ac{PHM} are generally subdivided into model-based, data-driven, and hybrid approaches~(see Sec.~\ref{subsec:prognostics_and_health_management}). In this context, \ac{PIML} is generally regarded as a paradigm within hybrid approaches. Nonetheless, a key observation motivated distinguishing \ac{PIML} from hybrid approaches, resulting in a classification scheme comprising four classes: observational bias, inductive bias, learning bias, and hybrid approaches (see Fig.~\ref{fig:classification_scheme}). While all four approaches involve physics and \ac{ML}, the last category is conceptually distinct with respect to the role that physics plays within the overall model. Crucially, it departs from the core principle of \ac{IML} that prior knowledge is explicitly integrated into the  \ac{ML} pipeline~\citep{vonrueden2021informed}, since hybrid approaches couple two independent models. In light of this, studies in the fourth class could technically be considered outside the scope of this review. However, to foster understanding of these conceptual differences and to provide a comprehensive overview of the intersection between physics and \ac{ML} in \ac{PHM}, this class has been deliberately included. In doing so, the distinct role of \ac{PIML} within the broader landscape becomes more evident.

% Adoption of PIML definition according to Karniadakis et al., with further refinement regarding observational bias (i.e., regarding "data that embody underlying physics")
The first three classes are based on the well-estab\-lished three pathways for \ac{PIML}. While inductive and learning bias are defined sufficiently to allow accurate classification, observational bias requires a stricter interpretation to avoid conflating physics-informed approaches with conventional \ac{ML}. In the original work, \cite{karniadakis2021physics} describe two ways of introducing observational bias, namely \enquote{through data that embody the underlying physics or carefully crafted data augmentation procedures.} While the former applies to virtually any \ac{ML} problem involving real-world systems whose behavior follows physical principles, the latter provides little practical guidance, as \enquote{carefully crafted} is not further defined. To resolve the former ambiguity, the definition of prior knowledge as formulated by~\cite{vonrueden2021informed} is adopted, which specifies that it \enquote{exist[s] in an external, separated way from the learning problem and the usual training data.} Hence, observational bias is not introduced by the mere use of empirical data obtained from physical systems. As a result, the inclusion of additional (physics-informed) data becomes obligatory to incorporate this form of bias. Primary approaches include the use of simulated data, either to enrich the training data or as the source domain data within a \ac{TL} setting---both strategies essentially addressing data scarcity. This also implies, however, that training (and testing) exclusively on simulated data does not qualify as \ac{PIML}, as it fails to meet the requirement that prior knowledge is separate from the training data.

% Adoption of PIML definition according to Karniadakis et al., with further refinement regarding "carefully crafted data augmentatation procedures"
The ambiguity surrounding carefully crafted data augmentation procedures is resolved as follows. On the one hand, it is virtually impossible to determine what qualifies as carefully crafted, which is why conventional data augmentation techniques are categorically excluded from the scope of \ac{PIML}. By the same reasoning, feature engineering (regardless of its complexity) does not constitute \ac{PIML} either. Ultimately, such measures remain conventional \ac{ML} in that they operate on or derive from existing training data. On the other hand, the integration of entire physics-based models in conjunction with an \ac{ML} model forms the basis of the fourth class. While some studies employing a hybrid approach could, in principle, have been classified under observational bias according to the broad definition given by~\cite{karniadakis2021physics}, it was established earlier that they violate the requirement of explicit integration of prior knowledge into the \ac{ML} pipeline. Instead, the physics-based and \ac{ML} components interact largely independently.

\begin{figure*}[h]
    \centering
    \includegraphics[width=\textwidth]{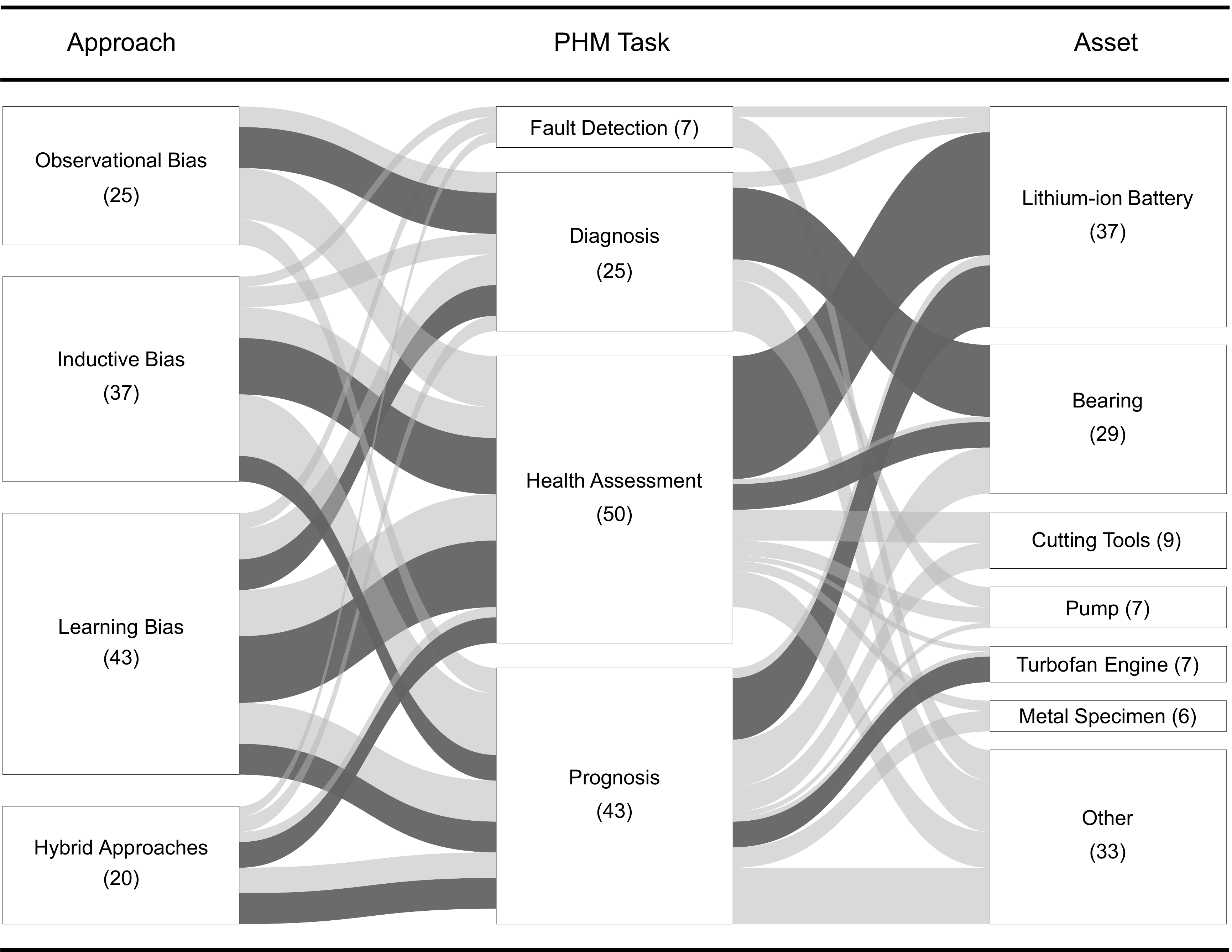}
    \caption{This Sankey diagram shows three categories---\textit{approach}, \textit{\ac{PHM} task} and \textit{asset}---each containing multiple items, with connections between them representing the state of research based on the 118 studies discussed in Sections~\ref{subsec:observational_bias}--\ref{subsec:hybrid_approaches}. The width of each path connecting two items is proportional to the number of studies supporting that link, reflecting the relative strength of evidence. To highlight areas of focus, paths supported by five or more studies across all three categories (i.e., studies that employ the same \textit{approach}, target the same \textit{PHM task} and address the same \textit{asset}) are shown in darker gray. Within the \textit{asset} category, only those studied by five or more contributions are represented as individual items, while all remaining assets are aggregated under \enquote{Other} to preserve visual clarity. Moreover, the number of studies corresponding to each item is indicated in brackets. As some studies contribute to multiple items within a category (e.g., by employing different approaches), category totals may exceed 118.}
    \label{fig:sankey}
\end{figure*}

% As a result of that: class four - hybrid approaches
Based on a thorough examination of studies in this fourth class, two types of approaches emerged, referred to as in-parallel and in-series, respectively. Drawing on terminology from electrical engineering, these labels aptly capture the relationships between the physics-based and \ac{ML} models. Primary in-parallel approaches include residual learning and ensemble methods. In-series approaches involve integrating the physics-based and \ac{ML} models sequentially, so that the output of one model directly informs the input of the other. Conceptually, this sequential integration can occur in two variants, with the \ac{ML} model directing the physics-based model, or vice versa. Importantly, these hybrid approaches differ in two key aspects with respect to observational bias and feature engineering, respectively. Unlike approaches that incorporate observational bias, in which physics-based simulators are used only offline and are no longer required once the \ac{ML} model is trained, hybrid approaches retain a physics-based model as an integral component of the prediction pipeline during both training and inference. Additionally, they go beyond feature engineering, where fixed algebraic relations are used to transform input variables but do not constitute a model with meaningful physical parameters. Hence, studies following a hybrid approach involve a physics-based component corresponding to a parameterized model, combined either in parallel or in series with the \ac{ML} model. In addition, a clear distinction is drawn with respect to inductive-bias approaches that appear to employ an in-series coupling. If a physics-based model is inserted as a fixed, differentiable module within the computational graph (i.e., the model output is passed through the physics-based model prior to computing the loss, and gradients are propagated through this module), it is referred to as being a part of the \ac{ML} model. This links back to the hypothesis space: all admissible predictions are, by construction, consistent with the employed physics module.

% Detailed description of the structure of the results chapter
All studies (excluding reviews) are classified and presented following the classification scheme outlined above. Each class is further subdivided by the four \ac{PHM} tasks: fault detection, diagnosis, health assessment, and prognosis~(see Sec.~\ref{subsec:prognostics_and_health_management}). Within these task-specific sections, studies are grouped and reported according to related use cases, where possible. Studies that incorporate multiple mechanisms for embedding physics into \ac{ML}, or that target multiple \ac{PHM} tasks, are addressed as follows. If mechanisms or tasks are clearly separable, they are reported individually and may therefore appear multiple times across the following sections. Studies with tightly coupled mechanisms are assigned to the class that best reflects the primary driving mechanism. The same logic applies to \ac{PHM} tasks. For each class of approaches, a representative example is presented in a dedicated figure, highlighting the prior knowledge employed and its incorporation. The schematics illustrating each method are adapted from the corresponding study and simplified to highlight the essential information, where blue denotes the \ac{ML} components and orange the physics components. Lastly, the Sankey diagram in Figure~\ref{fig:sankey} offers a visual overview, capturing overall trends regarding the approaches employed for specific \ac{PHM} tasks across various assets, thereby offering context for subsequent analysis of methodological patterns.

% - - - - - - - - - - subsection - - - - - - - - - - %
\subsection{Reviews}
\label{subsec:reviews}

% Overview
The relevant literature comprises eleven review studies published between 2019 and 2025, exploring how the integration of physics into data-driven methods has advanced the broader field of \ac{PHM}. These reviews offer valuable insights into the field's evolution and present unique perspectives. The following summaries are presented in order of relevance to the current work, highlighting each review's contributions, methodologies, and individual strengths and limitations.

\setlength{\tabcolsep}{3pt}
\begin{table}[ht]
    \centering
    \caption{Overview of all identified reviews, listed in order of relevance. Dashes indicate omitted information.}
    \label{tab:review_overview}
    \begin{tabular}{ 
        >{\raggedright\arraybackslash}p{2.2cm} 
        >{\centering\arraybackslash}p{1.5cm}
        >{\centering\arraybackslash}p{1.2cm} 
        >{\centering\arraybackslash}p{1.8cm} 
    }
        \hline
        \textbf{Reference} & 
        \textbf{Publication Year} & 
        \textbf{No. of Studies} & 
        \textbf{Time Period} \\ \hline

        \cite{deng2023Physicsinformedmachinelearning} & 2023 & 122 & 2013--2023 \\

        \cite{wu2024Physicsinformedmachinelearning} & 2024 & 107 & -- \\

        \cite{li2024Areviewon} & 2024 & 106 & 2003--2022 \\

        \cite{khan2024Areviewof} & 2024 & -- & -- \\

        \cite{yan2025KnowledgeDrivenMachine} & 2025 & -- & -- \\
        
        \cite{fassi2024TowardPhysicsInformedMachineLearningBased} & 2024 & -- & -- \\ 

        \cite{zhao2024Batterysafety:Machine} & 2024 & -- & -- \\

        \cite{cuesta2025Areviewof} & 2025 & 87 & 2018--2024 \\

        \cite{zhu2023Physicsinformedmachinelearning} & 2023 & -- & -- \\

        \cite{kundu2020Areviewon} & 2020 & -- & -- \\

        \cite{meng2019Areviewon} & 2019 & -- & 2009--2018 \\

        \hline
    \end{tabular}
\end{table}

% Deng et al.
\cite{deng2023Physicsinformedmachinelearning} present a broad overview of \ac{PIML} in \ac{PHM}, without focusing on any specific domain or task. The review is well-motivated, highlighting the advantages of \ac{PIML} over physical model-based and purely data-driven methods, respectively. While the methodology is the most detailed among the reviews discussed here, it lacks specifics on the screening procedure, only stating that it was done manually without explaining the criteria used to differentiate relevant from irrelevant studies. The identified studies are organized in a manner reminiscent of the three pathways outlined by~\cite{karniadakis2021physics}, albeit using different terminology. However, a significant concern is that, despite identifying 122 relevant studies, only 69 are categorized into the three approaches, leaving 53 unaddressed without any explanation from the authors.

% Wu et al.
The comprehensive review by~\cite{wu2024Physicsinformedmachinelearning} focuses on the application of \ac{PIML} to anomaly detection and condition monitoring, but fails to establish a connection to \ac{PHM}. With the methodology only briefly outlined, key details on screening and selection are missing. Supported by informative figures, the various strategies for integrating physics into ML provide a clear structure for the review. Overall, it offers a valuable contribution with a thorough analysis of recent developments, though the level of detail can be excessive.

% Li et al.
\cite{li2024Areviewon} thoroughly review methods for predicting \ac{RUL}, focusing on \ac{PIML} while also identifying and discussing the fusion of physics-based and data-driven models and the development of stochastic degradation models. The review demonstrates technical depth and includes striking figures to clarify various approaches, but the repeated subdivision of \ac{PIML} methods with inconsistent terminology impedes comprehension. Although the methodology for identifying relevant literature is briefly outlined, the authors themselves acknowledge that an exhaustive collection of studies remains elusive.

% Khan et al.
\cite{khan2024Areviewof} review physics-based learning for system health management but fail to establish a clear connection to \ac{PHM}. Although the study provides a transparent background, the methodology is vaguely described, offering only examples of databases and keywords and failing to detail the screening procedure. The authors reference \cite{karniadakis2021physics} when introducing \ac{PIML}, yet merely distinguish between physics-based loss functions and \enquote{various architectures}---a category that lacks clear definition and fails to meaningfully differentiate between the diverse approaches emerging in the field. Rather than offering a comprehensive analysis of the identified literature, the review discusses some \ac{PIML} approaches in general terms, with only a limited selection of examples provided. 

% Yan et al.
The review by~\cite{yan2025KnowledgeDrivenMachine} sets out to introduce a \enquote{universal concept, knowledge driven machine learning, for integrating diverse knowledge into machine learning pipeline [sic] in PHM domain.} However, it largely reproduces the \ac{IML} taxonomy by~\cite{vonrueden2021informed} without demonstrable novelty. While the authors attempt to synthesize their findings systematically and provide tangible case studies, the article’s contribution is weakened by conceptual ambiguity, methodological omissions, and linguistic inaccuracies, limiting its value as a rigorous or original synthesis of the field.

% Fassi et al.
\cite{fassi2024TowardPhysicsInformedMachineLearningBased} provide a comprehensive overview of \ac{PdM} for power converters, structured around model-based, data-driven, and hybrid approaches, with notable emphasis on \ac{PIML}. However, the review overlooks \ac{PHM} as the foundational framework for \ac{PdM} and follows a narrative rather than systematic methodology. Moreover, the discussion of studies employing \ac{PIML} partly diverges from power converters, drawing substantially on adjacent domains. Lastly, the absence of an in-depth discussion of the reviewed studies limits broader insights, leading to a brief and generic outlook on future work.

% Zhu et al.
\cite{zhu2023Physicsinformedmachinelearning} provide a brief review of the application of \ac{PIML} in structural integrity, including failure mechanism modeling and \ac{PHM}. However, the review’s completeness is undermined by the lack of a reported methodology. Instead of offering a critical analysis, it mainly reports the identified studies, missing opportunities to extract valuable insights, such as the prior physical knowledge used, resulting in a weak foundation for discussing current challenges.

% Zhao et al.
The overview of \ac{ML}-based battery safety by~\cite{zhao2024Batterysafety:Machine} serves as a solid entry point, outlining the core mechanisms driving battery faults and failures. Yet it provides no information on the methodology, rendering the review narrative rather than systematic and leaving its completeness uncertain. Despite the title emphasizing prognostics, studies spanning the entire \ac{PHM} spectrum are covered. With respect to \ac{PIML}, the structure is somewhat diffuse: \ac{ML} in conjunction with battery models is first presented (mainly as a source of domain-specific features), yet several of these approaches effectively fall under \ac{PIML}, with a later section specifically dedicated to \ac{PIML} further blurring conceptual boundaries.

% Cuesta et al.
The review by~\cite{cuesta2025Areviewof} provides a comprehensive overview of \ac{PHM} in wind energy, focusing on \ac{RUL} estimation for key turbine components such as gearboxes, generators, blades, and bearings. It outlines three main degradation modeling approaches: physics-based, data-driven, and hybrid models. In discussing hybrid models, the authors emphasize the integration of physical knowledge into \ac{ML} frameworks but do not explicitly situate this discussion within the emerging field of \ac{PIML}. The review further identifies challenges related to uncertainty quantification, integration of physical knowledge, environmental variability, and system complexity. However, the absence of a clear methodology for study selection raises concerns about the completeness and representativeness of the reviewed literature.

% Kundu et al.
\cite{kundu2020Areviewon} present a review of diagnostic and prognostic approaches for gears. Offering both technical depth and clarity, this review is an excellent starting point for researchers and practitioners developing or applying \ac{PHM} in this area. However, the review lacks information on the process used to gather the relevant literature. Additionally, the authors report only a few studies on hybrid approaches, explaining that few such methods have been developed, and attempt to cover all major approaches up to and including 2020---none of which employ \ac{PIML}.

% Meng and Li
The review on \ac{PHM} of lithium-ion batteries by~\cite{meng2019Areviewon} is structured into physics-based, data-driven, and hybrid approaches, with the latter focusing primarily on \ac{PF}-based and Kalman filter-based methods. Given the lack of methodological rigor, the work constitutes a descriptive survey rather than a systematic analysis. Moreover, the absence of studies employing \ac{PIML} as a tool for \ac{PHM}, likely due to the review's 2019 publication date, underscores the rapid progress of the field.

% Conclusion
While acknowledging the significant contributions of previous reviews at the intersection of \ac{PHM} and \ac{PIML}, it is concluded that a more comprehensive and systematic review is needed to fully address the research questions outlined in Section~\ref{sec:introduction}. Although these reviews provide valuable insights, they often offer only partial coverage, focusing on particular domains, e.g., mechanical components~\citep{kundu2020Areviewon}, batteries~\citep{meng2019Areviewon, zhao2024Batterysafety:Machine}, power converters~\citep{fassi2024TowardPhysicsInformedMachineLearningBased}, or wind energy~\citep{cuesta2025Areviewof}, or specific \ac{PHM} tasks, such as \ac{RUL} prediction~\citep{li2024Areviewon}, and in certain instances, extending more broadly to related areas, such as condition monitoring~\citep{wu2024Physicsinformedmachinelearning} or \ac{PdM}~\citep{fassi2024TowardPhysicsInformedMachineLearningBased}. Moreover, in some cases, \ac{PIML} is not the central focus but is instead addressed only as a peripheral topic. Furthermore, a recurring issue across all reviews is the lack of a systematic approach and detailed descriptions, which undermines scientific rigor, as transparency and reproducibility are essential to research integrity. Notably, most reviews fail to report the number of included studies or the time period they cover (see Tab.~\ref{tab:review_overview}). In response, this review seeks to bridge existing gaps and advance the understanding and application of \ac{PIML} in \ac{PHM} by providing a comprehensive, focused exploration of the state of the art, guided by a rigorous methodology.

% - - - - - - - - - - subsection - - - - - - - - - - %
\subsection{Observational Bias}
\label{subsec:observational_bias}

\begin{figure*}[t]
    \centering
    \includegraphics[width=\textwidth]{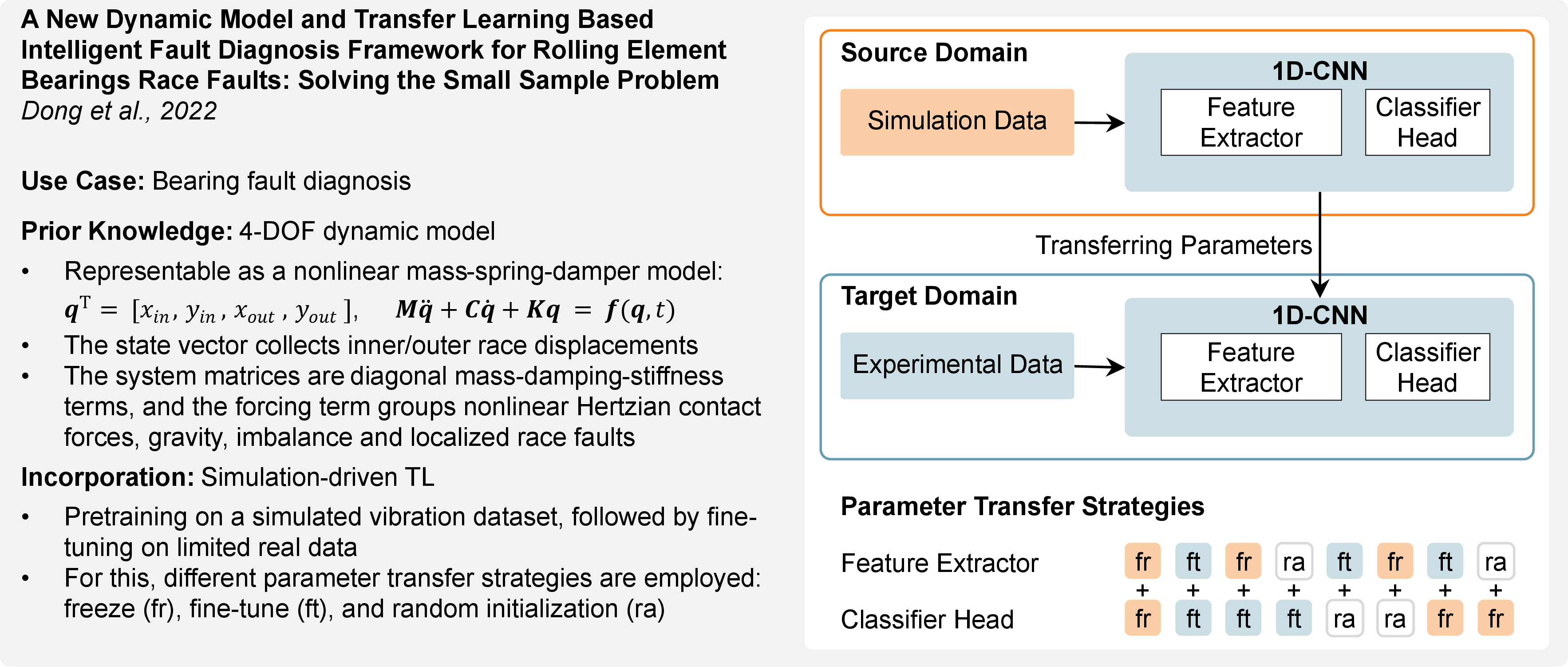}
    \caption{A representative example of observational-bias approaches is proposed by~\cite{dong2022Anewdynamic}, demonstrating how prior physical knowledge can be introduced as additional training data. Own illustration based on the corresponding study. See the original work for full technical details.}
    \label{fig:prime_example_observational_bias}
\end{figure*}

% Outlook for the section to follow
Observational bias is characterized by introducing prior physical knowledge directly through the training data. The identified studies are organized 
by PHM task in 
Tables~\ref{tab:observational_bias_diagnosis}--\ref{tab:observational_bias_prognosis}. A representative example is illustrated in Figure~\ref{fig:prime_example_observational_bias}.

% Prime example

\subsubsection{Fault Detection}
\label{subsub:fault_detection_observational_bias}

No studies have been identified that introduce observational bias for the purpose of fault detection.

\subsubsection{Diagnosis}
\label{subsub:diagnosis_observational_bias}

\begin{table*}[ht]
\centering
\caption{All studies employing \textit{observational bias} to address \textit{diagnosis}, listed in alphabetical order.}
\label{tab:observational_bias_diagnosis}
\begin{tabular}{@{}
  >{\RaggedRight\arraybackslash}p{0.24\textwidth}%
  >{\RaggedRight\arraybackslash}p{0.31\textwidth}%
  >{\RaggedRight\arraybackslash}p{0.23\textwidth}%
  >{\RaggedRight\arraybackslash}p{0.18\textwidth}%
  @{}}
% \rowcolor{gray!20}
\multicolumn{4}{@{}l@{}}{\bfseries Observational Bias for Diagnosis} \\
\hline
\textbf{Reference} 
& \multicolumn{2}{l}{\textbf{Prior Physical Knowledge}} 
& \textbf{Use Case} \\
& Type 
& Representation
& \\
\hline

\cite{dong2022Anewdynamic} & 4-\acs{DOF} dynamic bearing model & \acs{ODE} system & Bearing \\

\cite{li2024Asimulationdatadriven} & Numerical simulation of bearing fault vibration, based on impulse-response signal model & Algebraic equation & Bearing \\

\cite{liu2024Enhancingmultitypefault} & Discretized first-order \acs{ECM} & Recursive algebraic equation & Lithium-ion battery \\

\cite{ma2025Aphysicsbasedsample} & Multi-\acs{DOF} rotor-bearing dynamic model & Coupled nonlinear \acsp{ODE} & Bearing \\

\cite{pettorossi2024AddressingDataScarcity} & Proton exchange membrane fuel cell simulation & \acs{ODE} system & Fuel cell \\

\cite{qin2024Inversephysicsinformed} & 2-\acs{DOF} dynamic bearing model & \acs{ODE} system & Bearing \\

\cite{qin2025SimulationdataDrivenGeneralized} & 2-\acs{DOF} dynamic bearing model & \acs{ODE} system & Bearing \\

\cite{song2022Researchonfault} & Rigid-flexible multibody dynamics simulation of a planetary gearbox & Differential-algebraic equation system & Planetary gearbox \\

\cite{zhang2024AdversarialDomainAdaptation} & System level axial piston pump model (rotor, bearings, fluid), bearing dynamic model & \acs{ODE} system & Axial piston pump, bearing \\

\cite{zhu2025DataGenerationApproach} & 4-\acs{DOF} dynamic bearing model & \acs{ODE} system & Bearing \\

\hline
\end{tabular}
\end{table*}

% Lithium-ion battery
A total of ten studies employing observational bias for diagnosis have been identified (see Tab.~\ref{tab:observational_bias_diagnosis}). \cite{liu2024Enhancingmultitypefault} aim to solve the problem of high similarity among different faults in lithium-ion battery voltage signatures. Considering faults, such as internal short circuits, capacity anomaly, and \ac{SOC} anomaly occurring in a series-connected battery system, the authors calculate fault features by applying a mean difference model to terminal voltages. These serve as input to a vision Transformer, followed by a \ac{MLP} head for fault diagnosis. They use a discretized first-order \ac{ECM} to generate simulation data for pretraining the model, while fine-tuning is performed on experimental data, leading to improved accuracy. 

% Bearing
\cite{qin2024Inversephysicsinformed} focus on the challenge of accurately diagnosing bearing faults given imbalanced fault samples. An inverse \ac{PINN} is employed to estimate the parameters of a 2-\ac{DOF} dynamic bearing model, specifically stiffness and damping ratio, from real vibration data. This enables the generation of simulated fault data whose frequency-domain characteristics closely match those of real measurements. The simulated data are then used to supplement imbalanced datasets, significantly enhancing the accuracy of a \ac{CNN}-based diagnostic model for bearing fault detection. In another work, \cite{qin2025SimulationdataDrivenGeneralized} address the challenge of unseen compound faults in bearing fault diagnosis. The approach employs a composed model consisting of a \ac{CNN}-based feature extractor, a Cycle-\ac{GAN}-based semantic mapping model (trained on simulated data) and a Cycle-\ac{GAN}-based feature generator (trained on measured single-fault data), and a multi-agent deep \ac{RL} model for the final diagnosis. Simulated vibration signals for various single and compound fault types are generated with a 2-\ac{DOF} bearing dynamic model. These signals are employed in conjunction with their respective envelope spectra as semantics to train the semantic mapping model. Hence, zero-shot learning regarding real-world compound faults can be performed. The proposed approach demonstrates superior performance in comparison to other state-of-the-art methods for three bearing datasets.
\cite{dong2022Anewdynamic} present a fault diagnosis framework for bearing race faults that addresses the small-sample problem using a dynamic model and \ac{TL} strategies (see Fig.~\ref{fig:prime_example_observational_bias}). Simulated data generated from a 4-\ac{DOF} bearing dynamic model are used to pretrain a \ac{CNN}, and parameter transfer strategies (i.e., selectively freezing or fine-tuning different parts of the network) adapt the model to limited real-world data. By aligning simulation and real-world conditions, the approach enables effective feature transfer and reduces distribution mismatch. Compared to conventional \ac{ML} models (i.e., \ac{SVM}, \ac{kNN}, \ac{RF}, and \ac{MLP}), both with and without simulation-based \ac{TL}, the proposed \ac{CNN}-based \ac{TL} approach consistently outperforms all alternatives in bearing fault diagnosis under small-sample conditions.
\cite{li2024Asimulationdatadriven} address the challenge of bearing fault diagnosis with unlabeled data. Both completely unlabeled real data and labeled simulation data from a bearing failure simulation are combined in a semi-supervised method. The simulation covers time-domain vibration signals for faults occurring in the outer race, inner race, and rolling elements. Specifically, a multi-kernel \ac{kNN} graph connecting both data sources is built and edge weights are refined with both layer attention and dot-product attention. The resulting node embeddings are fed into fully connected layers for fault classification. The proposed approach outperforms advanced \ac{GNN}-based methods.
The domain gap problem regarding simulation and real data in bearing small-sample fault diagnosis is studied by~\cite{zhu2025DataGenerationApproach}. A bearing dynamic model is used to generate time-series vibration data and corresponding fault labels, which are used as inputs for the generator of a conditional deep convolutional \ac{GAN} instead of random noise. The discriminator is trained to distinguish real vibration data from \ac{GAN}-generated data, encouraging the generator to produce synthetic signals that both respect the underlying fault mechanism and closely match the distribution of real measurements. This generated data can then be used to train a fault diagnostic model alongside real data. The presented method achieves higher accuracy and lower variance compared to a scenario in which the original simulation and real data are used directly to train the fault diagnostic model.
\cite{ma2025Aphysicsbasedsample} address few-shot bearing fault diagnosis by using a nonlinear rotor-bearing system model to simulate mechanistic vibration signals for different fault types and severities. These signals are combined with equipment-specific characteristics extracted from healthy data, denoised via wavelet packet decomposition, and further refined using a cosine-similarity-guided noise injection, yielding a mixed dataset that comprises simulated and scarce real fault samples. Across multiple \ac{DL} architectures, training on the mixed dataset results in superior diagnostic accuracy compared to training solely on measured, simulated, or \ac{GAN}-generated data. 

% Pump
% Axial piston pump/bearing
The small-sample problem, resulting from the limited availability of fault data for mechanical components, is addressed by~\cite{zhang2024AdversarialDomainAdaptation} through the use of simulation models. Time-series data from the simulation are used as the source domain, along with unlabeled experimental data as the target domain, to train an adversarial domain adaptation model with a feature extractor, healthy pattern recognizer, and domain discriminator. Experiments on both axial piston pump and bearing datasets demonstrate that the proposed method significantly enhances fault-diagnosis accuracy and generalization under small-sample conditions, outperforming other domain-adaptation methods.

% Gearbox
\cite{song2022Researchonfault} present a method for planetary gearbox fault diagnosis based on \ac{TL}. In this study, gearbox dynamics are simulated to generate labeled training data, addressing the challenge of limited real-world fault samples. A deep \ac{TL} framework is then applied to generalize the fault detection model across different operating conditions. The framework employs a \ac{CNN}-based feature extractor with shared parameters for both simulated source domain data and experimental target domain data, as well as a classifier. Besides the classification loss, a domain adaptation loss is employed to guide domain-invariant features. Experimental results for different \ac{TL} tasks demonstrate the effectiveness of this approach in identifying cracked gear and missing tooth faults. The proposed method performs slightly better than comparable \ac{TL} baselines.

% Fuel cell (proton exchange membrane fuel cell)
Both the small-sample problem present in proton exchange membrane fuel cell fault diagnosis and the distribution mismatch between simulated and real data are studied by~\cite{pettorossi2024AddressingDataScarcity}. They use a supervised domain-adversarial adaptation model, combining a \ac{LSTM}-based feature extractor, a fully connected fault classifier, and a gradient-reversal-based domain discriminator. A simulated dataset produced by a calibrated physics-based proton exchange membrane fuel cell model is used as the source domain, whereas measurements from a real stack serve as the target domain. In experiments incorporating varying amounts of real data into the training process, the proposed approach consistently outperforms a baseline trained exclusively on real data without the domain discriminator. Notably, these performance gains become more pronounced as the quantity of real data decreases. 

\subsubsection{Health Assessment}
\label{subsub:health_assessment_observational_bias}

\begin{table*}[ht]
\centering
\caption{All studies employing \textit{observational bias} to address \textit{health assessment}, listed in alphabetical order.}
\label{tab:observational_bias_health_assessment}
\begin{tabular}{@{}
  >{\RaggedRight\arraybackslash}p{0.24\textwidth}%
  >{\RaggedRight\arraybackslash}p{0.31\textwidth}%
  >{\RaggedRight\arraybackslash}p{0.23\textwidth}%
  >{\RaggedRight\arraybackslash}p{0.18\textwidth}%
  @{}}
\multicolumn{4}{@{}l@{}}{\bfseries Observational Bias for Health Assessment} \\
\hline
\textbf{Reference} 
& \multicolumn{2}{l}{\textbf{Prior Physical Knowledge}} 
& \textbf{Use Case} \\
& Type 
& Representation
& \\
\hline

\cite{bachar2024AMultidisciplinaryFramework} & Gear vibration model & Nonlinear second-order \acs{ODE} & Gear \\

\cite{kohtz2022Physicsinformedmachinelearning} & 1D \acs{FE} \acs{SEI}-growth model & Coupled \acsp{PDE} & Lithium-ion battery \\

\cite{li2024PhysicsGuidedDeepLearning} & Empirical flank wear evolution model, mechanistic milling force model & Algebraic equations & Milling \\

\cite{yishengliu2024HybridFusionfor} & Extended \acs{SPM} & Nonlinear state-space system derived from coupled \acs{PDE}-based electrochemical equations & Lithium-ion battery \\

\cite{matania2023Onefaultshotlearningfor} & Gear vibration model & Nonlinear second-order \acs{ODE} & Gear \\

\cite{mei2024Ahybridphysicsinformed} & Physics-of-failure-based relay degradation model & Coupled first-order \acsp{ODE} with time-varying stochastic input parameters, evaluated via \acs{FE}-based numerical simulation & Electromagnetic relays \\

\cite{navidi2024PhysicsInformedMachineLearning} & Half-cell degradation model & Algebraic equations & Lithium-ion battery \\

\cite{ren2025Healthassessmentof} & Electromagnetic and electromechanical model & Coupled \acsp{ODE} implemented via 2D \acs{FE} simulation & Brushless direct-current motor \\

\cite{sun2022MicrocrackDefectQuantification} & Ultrasonic guided-wave propagation and scattering model & \acsp{PDE} solved via \acs{FE} simulation & Microcrack quantification in aluminum plate \\

\cite{thelen2022Integratingphysicsbasedmodeling} & Half-cell degradation model & Algebraic equations & Lithium-ion battery \\

\hline
\end{tabular}
\end{table*}

% Lithium-ion batteries
As shown in Table~\ref{tab:observational_bias_health_assessment}, ten studies employ observational bias for health assessment. The study by~\cite{yishengliu2024HybridFusionfor} focuses on accurately estimating the \ac{SOH} of lithium-ion batteries under realistic fast-charging conditions using minimal labeled data. A reduced-order electrochemical model is calibrated on laboratory cycling tests, expanded via stochastic perturbation of aging parameters, and then used to generate partial-charge profiles for pretraining a \ac{CNN}. The pretrained network is subsequently adapted to specific batteries via \ac{TL} with only a small number of real partial-charge segments. Ablation studies (pre- and post-\ac{TL}, varying real-data volume and distribution) demonstrate that incorporating simulated data significantly improves generalization and enables extrapolation from early-life data to mid- and late-life degradation states.
\cite{kohtz2022Physicsinformedmachinelearning} introduce a multi-fidelity framework for estimating the \ac{SOH} of lithium-ion batteries from a single short partial charging segment. A 1D \ac{FE} \ac{SEI}-growth model is first used to simulate capacity fade and \ac{SEI} thickness, which are subsequently fused with experimental data to train a co-kriging multi-fidelity model that links \ac{SEI} thickness to capacity (\ac{SOH}). Separately, nested \ac{GPR} models are trained on \ac{FE} data to map operating conditions and \ac{SEI} levels to voltage points and charging time, which are then used to infer \ac{SEI} thickness---and thus \ac{SOH}---from a single partial charging segment without any prior usage history. The authors validate their proposed framework by comparing the \ac{FE} model, single-fidelity surrogates, and the co-kriging multi-fidelity model, with the latter attaining the smallest capacity estimation errors. However, it is not quantitatively benchmarked against alternative \ac{SOH} estimation methods.
\cite{thelen2022Integratingphysicsbasedmodeling} address the online health assessment of lithium-ion batteries by estimating both capacity (\ac{SOH}) and three internal degradation modes. Prior physical knowledge is incorporated via a half-cell model, which is used to generate degradation data and subsequently combined with limited early-life experimental data. Two approaches are compared: data augmentation, in which simulated and experimental data are merged to train a single model, and delta learning, in which an estimator trained on simulated data is corrected by a second model trained on experimental data. Across both scenarios, data augmentation consistently outperforms delta learning for all four lightweight \ac{ML} models tested, with the elastic net (a linear regression model with combined L1/L2 regularization) achieving the highest overall performance. As a follow-up study, \cite{navidi2024PhysicsInformedMachineLearning} compare four approaches, including the elastic net-based data augmentation and delta learning methods from the previous study~\citep{thelen2022Integratingphysicsbasedmodeling}. Additionally, delta learning employing \ac{GPR} is examined, alongside another method that introduces a learning bias (see Sec.~\ref{subsub:health_assessment_learning_bias}). Among the approaches incorporating observational bias, the \ac{GPR}-based method yields the lowest errors for capacity estimation. A qualitative comparison of all four methods, considering aspects such as model flexibility, data requirements, and ease of implementation, provides a broader perspective on their applicability.

% Milling (cutting tools)
\cite{li2024PhysicsGuidedDeepLearning} propose a method for online tool condition monitoring in milling, where the goal is to estimate flank wear from cutting-force signals under varying operating conditions. A mechanistic tool wear model and a milling force model are first calibrated on a small set of offline measurements and then used to generate synthetic data. Simulated force signals are used to pretrain a \ac{DL} model that combines a residual network with learnable soft-threshold attention (for denoising and feature extraction) and a \ac{BiLSTM} (for mapping force features to simulated wear labels). With a strong initialization provided by pretraining on simulated data, the model is subsequently fine-tuned on a limited set of real samples. In rigorous comparisons with its purely physics-based and purely data-driven counterparts, as well as with \ac{PIML} approaches and several standard \ac{ML} baselines, the proposed method consistently achieves lower errors and more robust wear estimation across multiple milling conditions, while requiring substantially fewer real labels.

% Aluminum plate (crack estimation)
By quantifying existing microcracks from ultrasonic guided-wave measurements obtained via a specially designed, unidirectionally focusing electromagnetic acoustic transducer and a circular array of receivers around the suspected defect location, \cite{sun2022MicrocrackDefectQuantification} tackle the health assessment of metallic plate structures. To compensate for the small set of experimental measurements, synthetic guided-wave signals are generated using an \ac{FE} simulation that spans a broad range of crack geometries. The architecture includes two branches with shared weights that process experimental and simulated signals in parallel, enabling the model to compare them and infer crack length. The method additionally incorporates a learning bias (see Sec.~\ref{subsub:health_assessment_learning_bias}). While the proposed model achieves substantially lower errors compared to conventional \ac{NN} variants, the study does not include ablation studies to infer the corresponding contribution of the various biases incorporated.

% Gear
\cite{matania2023Onefaultshotlearningfor} address fault severity estimation for spur gears when only one faulty experimental sample is available. A dynamic gear model is used to generate vibration signals. The vibration data undergo several preprocessing steps, such as angular resampling, synchronous averaging, propagation through an estimated transfer function, and extraction of time-domain features. The authors then mix the preprocessed simulation data with real data to serve as input for a \ac{kNN} regressor. The ablation studies conducted do not specifically study the effect of incorporating simulation data or comparisons against state-of-the-art baselines. Building on this work, some of the authors propose an unsupervised framework with a broader scope that can perform health assessment for multiple fault types~\citep{bachar2024AMultidisciplinaryFramework}.

% Motor
In their study on health assessment of brushless direct-current motor stators, \cite{ren2025Healthassessmentof} use a stator damage matrix derived from torque, speed, and load signals, which is then mapped to a scalar \ac{HI}. To this end, an \ac{LSTM} is trained on a mixed dataset of experimental and \ac{FE}-simulated torque sequences to predict a damage matrix describing turn-level open-circuit and short-circuit faults. Based on the predicted damage matrix, a scalar \ac{HI} is computed via cosine similarity between the current damage state and a health reference state, where a bootstrap procedure provides confidence intervals for the \ac{HI}. Compared with four model-based methods and a conventional \ac{LSTM}, the non-invasive approach improves damage estimation performance, especially under small-sample conditions.

% Electromagnetic relays
\cite{mei2024Ahybridphysicsinformed} employ a variational \ac{AE} to model degradation of the operating voltage signal of electromagnetic relays and perform time-dependent reliability (failure-probability) assessment. A high-fidelity physics-of-failure \ac{FE} simulation of the relay's electromagnetic and elastic subsystems is used to generate synthetic degradation trajectories, which are combined with a small set of experimental trajectories to train the variational \ac{AE}. This then functions as a generative degradation model: operating voltage trajectories are sampled over time, and at each time point, the fraction of generated samples exceeding a voltage threshold is used to estimate the time-dependent failure probability of a batch of relays. From a data perspective, training on the combined simulation and experimental trajectories outperforms training on either source alone. From a modeling perspective, the variational \ac{AE} achieves lower reliability-assessment error and computational cost than both \ac{GP}- and \ac{LSTM}-based variants.

\subsubsection{Prognosis}
\label{subsub:prognosis_observational_bias}

\begin{table*}[ht]
\centering
\caption{All studies employing \textit{observational bias} to address \textit{prognosis}, listed in alphabetical order.}
\label{tab:observational_bias_prognosis}
\begin{tabular}{@{}
  >{\RaggedRight\arraybackslash}p{0.24\textwidth}%
  >{\RaggedRight\arraybackslash}p{0.31\textwidth}%
  >{\RaggedRight\arraybackslash}p{0.23\textwidth}%
  >{\RaggedRight\arraybackslash}p{0.18\textwidth}%
  @{}}
\multicolumn{4}{@{}l@{}}{\bfseries Observational Bias for Prognosis} \\
\hline
\textbf{Reference} 
& \multicolumn{2}{l}{\textbf{Prior Physical Knowledge}} 
& \textbf{Use Case} \\
& Type 
& Representation
& \\
\hline

\cite{hervedebeaulieu2024RemainingUsefulLife} & Stiction model combined with a generic exponential degradation model & Nonlinear difference equation with a time-varying parameter prescribed by an exponential function & Air distribution system \\

\cite{deng2023ACalibrationBasedHybrid} & 5-\acs{DOF} dynamic bearing model & \acs{ODE} system & Bearing \\

\cite{zhang2023DynamicModelAssistedBearing} & 5-\acs{DOF} dynamic bearing model & \acs{ODE} system & Bearing \\

\cite{zhang2024Modeldatahybriddriven} & Cutting model (accounting for geometry, material, friction, and wear) & \acs{FE}-based numerical simulation & Milling \\

\cite{zhu2024RemainingUsefulLife} & Vibration degradation model & Analytic model defined by algebraic equations & Bearing \\

\hline
\end{tabular}
\end{table*}

% Bearing
Five studies employing observational bias for prognosis have been identified, which are listed in Table~\ref{tab:observational_bias_prognosis}. \cite{deng2023ACalibrationBasedHybrid} leverage a 5-\ac{DOF} dynamic bearing model for \ac{RUL} prediction across different machines. The dynamic model describes vibration under crack propagation and spall growth, with the associated damage parameters inferred via \ac{PF}-based calibration. Both calibrated simulated data and real measurements are fed into a Bayesian \ac{NN}, which processes one-dimensional time-series features through a \ac{GRU} branch and two-dimensional time-frequency features through a \ac{CNN} branch, jointly predicting the \ac{RUL}. Within an adversarial \ac{TL} setup, multiple domain discriminators are weighted according to the similarity of calibrated physical parameters, so that source bearings with degradation dynamics closer to the target bearing have a stronger influence during adaptation. Experiments on two benchmark bearing datasets demonstrate more accurate \ac{RUL} predictions than purely data-driven \ac{TL} baselines.
Similarly, \cite{zhang2023DynamicModelAssistedBearing} generate full life-cycle vibration data using a 5-\ac{DOF} dynamic bearing model. The resulting simulated data serve as the source domain for a multilayer Transformer-based network with maximum mean discrepancy-based domain alignment, which is trained to transfer degradation knowledge to measured vibration signals. Experimental results show that the proposed model substantially reduces \ac{RUL} prediction errors compared to several state-of-the-art methods, particularly when only limited measured run-to-failure data are available.
\cite{zhu2024RemainingUsefulLife} tackle \ac{RUL} prediction for rolling bearings, proposing a framework that incorporates physics in three ways. In terms of observational bias, a physics-based degradation model is used to simulate full life-cycle vibration data. The simulated signals are subsequently fused with sensed data for feature extraction, thereby enriching the training set used to train the \ac{BiLSTM}. In combination with the other two mechanisms (see Sec.~\ref{subsub:prognosis_learning_bias} and Sec.~\ref{subsub:prognosis_hybrid}, respectively), the proposed framework outperforms purely data-driven baselines by achieving lower prediction errors. However, due to the absence of ablation studies, it is not possible to isolate how much of the improvement is attributable to this particular strategy.

% Milling (cutting tools)
Cutting-tool \ac{RUL} prediction under scarce degradation data is studied by~\cite{zhang2024Modeldatahybriddriven}, who propose combining a physics-based cutting model with an improved inverse \ac{GP}. An \ac{FE} cutting model is used offline to generate full-life wear trajectories, which are used for initial parameter estimation of the inverse \ac{GP}. During operation, its parameters are updated online with measured wear via Bayesian inference. Experiments on three tool-wear datasets indicate improved predictive accuracy over a purely data-driven baseline. 

% Air distribution system (airplane cockpit)
In their study, \cite{hervedebeaulieu2024RemainingUsefulLife} focus on predicting the \ac{RUL} of industrial systems without relying on labeled run-to-failure data. Linear closed-loop models of an aircraft cockpit temperature control system are identified from flight data and coupled with an exponential valve-stiction degradation model to generate nominal and degraded time series. An \ac{AE} is trained only on nominal data to derive an unsupervised \ac{HI} from reconstruction error, which is then forecast using an \ac{LSTM}. The approach is validated through a reliability-based assessment using Weibull and Kolmogorov-Smirnov tests, showing internal consistency of predictions. However, the study does not report experimental comparisons to alternative methods.

% - - - - - - - - - - subsection - - - - - - - - - - %
\subsection{Inductive Bias}
\label{subsec:inductive_bias}

% Outlook for the section to follow
Inductive bias is characterized by tailored interventions to the model design. The identified studies are organized 
by PHM task in 
Tables~\ref{tab:inductive_bias_fault_detection}--\ref{tab:inductive_bias_prognosis}. A representative example is illustrated in Figure~\ref{fig:prime_example_inductive_bias}.

\subsubsection{Fault Detection}
\label{subsub:fault_detection_inductive_bias}

\begin{table*}[ht]
\centering
\caption{All studies employing \textit{inductive bias} to address \textit{fault detection}, listed in alphabetical order.}
\label{tab:inductive_bias_fault_detection}
\begin{tabular}{@{}
  >{\RaggedRight\arraybackslash}p{0.24\textwidth}%
  >{\RaggedRight\arraybackslash}p{0.31\textwidth}%
  >{\RaggedRight\arraybackslash}p{0.23\textwidth}%
  >{\RaggedRight\arraybackslash}p{0.18\textwidth}%
  @{}}
\multicolumn{4}{@{}l@{}}{\bfseries Inductive Bias for Fault Detection} \\
\hline
\textbf{Reference} 
& \multicolumn{2}{l}{\textbf{Prior Physical Knowledge}} 
& \textbf{Use Case} \\
& Type 
& Representation
& \\
\hline

\cite{feng2022FullGraphAutoencoder} & Topological prior (sensor information regarding hardware layout or shared subsystems) & Relation matrix fused with adjacency matrix & Liquid rocket engine \\

\cite{liu2025Graphembeddedpatchsense} & Topological prior (sensor information) & Adjacency matrix & Liquid rocket engine, train transmission system \\

\hline
\end{tabular}
\end{table*}

% Liquid rocket engine system
Two studies have been identified that employ inductive bias to address fault detection (see Tab.~\ref{tab:inductive_bias_fault_detection}). \cite{liu2025Graphembeddedpatchsense} address anomaly detection in large-scale multi-component systems with complex interdependencies. The authors propose a model based on an \ac{AE} and graphs that learns unit-level patch embeddings and models global relationships in multivariate time series in an unsupervised manner. Prior knowledge of system structure is encoded as a graph and incorporated via an adjacency matrix, guiding the model to produce representations consistent with known system topology for both anomaly detection and component-level localization. Experiments on liquid rocket engine and train transmission datasets demonstrate improved accuracy and reliability, at a higher computational cost. 
In a closely related study also considering the liquid rocket engine system, \cite{feng2022FullGraphAutoencoder} encode prior knowledge into the adjacency matrix as well. Topological relationships of physical sensors in terms of hardware layout or shared subsystems are leveraged. Improvements in both accuracy and generalizability are achieved, compared to strong non-graph-based and graph-based baselines.

\subsubsection{Diagnosis}
\label{subsub:diagnosis_inductive_bias}

% Prime example
\begin{figure*}[t]
    \centering
    \includegraphics[width=\textwidth]{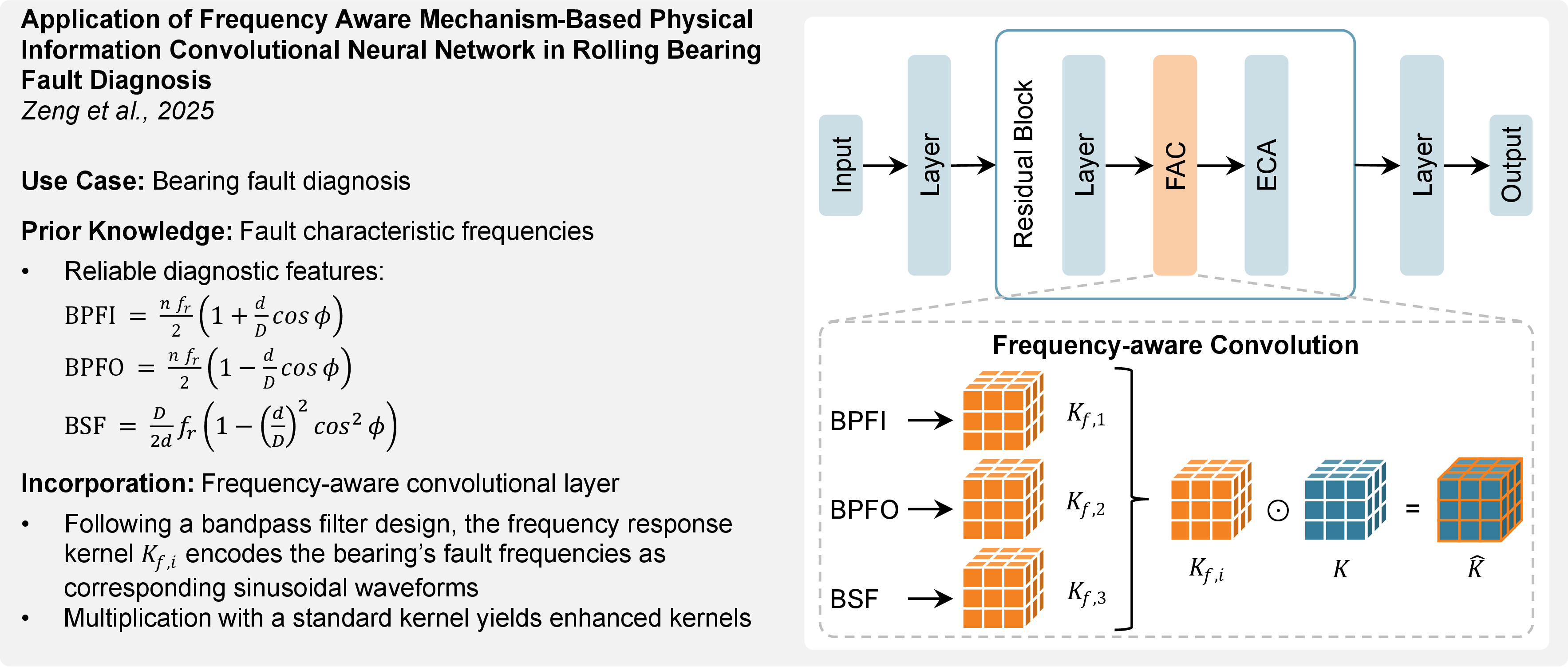}
    \caption{A representative example of inductive-bias approaches is proposed by~\cite{zeng2025ApplicationofFrequency}, demonstrating how prior physical knowledge can be introduced into the model design. Own illustration based on the corresponding study. See the original work for full technical details.}
    \label{fig:prime_example_inductive_bias}
\end{figure*}

\begin{table*}[ht]
\centering
\caption{All studies employing \textit{inductive bias} to address \textit{diagnosis}, listed in alphabetical order.}
\label{tab:inductive_bias_diagnosis}
\begin{tabular}{@{}
  >{\RaggedRight\arraybackslash}p{0.24\textwidth}%
  >{\RaggedRight\arraybackslash}p{0.31\textwidth}%
  >{\RaggedRight\arraybackslash}p{0.23\textwidth}%
  >{\RaggedRight\arraybackslash}p{0.18\textwidth}%
  @{}}
\multicolumn{4}{@{}l@{}}{\bfseries Inductive Bias for Diagnosis} \\
\hline
\textbf{Reference} 
& \multicolumn{2}{l}{\textbf{Prior Physical Knowledge}} 
& \textbf{Use Case} \\
& Type 
& Representation
& \\
\hline

\cite{gao2024MPINet:MultiscalePhysicsInformed} & Vibration model for a localized single-point defect, bearing fault characteristic frequencies & Analytical time-domain signal model, algebraic equations & Bearing \\

\cite{jin2024GraphSpatioTemporalNetworks} & Structural topology of the wind turbine and qualitative causal relationships & Directed graph over selected sensor signals (adjacency matrix) & Wind turbine \\

\cite{zeng2025ApplicationofFrequency} & Fault characteristic frequencies & Algebraic equations & Bearing \\

\hline
\end{tabular}
\end{table*}

% Rolling bearing
As shown in Table~\ref{tab:inductive_bias_diagnosis}, three studies employ inductive bias to target diagnosis. \cite{zeng2025ApplicationofFrequency} tackle rolling bearing fault diagnosis by embedding bearing fault frequency information into a \ac{CNN} via a novel \ac{FAC} layer (see Fig.~\ref{fig:prime_example_inductive_bias}). Vibration signals are first transformed into time-frequency maps via \ac{CWT}, which serve as inputs to the network. The architecture is built from multiple stacked units, each comprising a residual block, the \ac{FAC} layer, and an \ac{ECA} module~\citep{wang2020eca}. Using the characteristic fault frequencies of the inner ring (\acs{BPFI}), outer ring (\acs{BPFO}), and rolling body (\acs{BSF}), the \ac{FAC} layer generates a set of sinusoidal waveforms to form frequency response kernels (one per frequency) with a bandpass profile. A learnable shift parameter, initialized to zero and constrained within a physically meaningful range, allows these kernels to dynamically adapt to varying vibration signals across different operating conditions while remaining close to the analytical fault frequencies. Each frequency response kernel is multiplied element-wise with a standard convolutional kernel to form enhanced kernels that selectively emphasize the corresponding fault-related frequency bands while suppressing irrelevant components. The proposed model demonstrates superior accuracy, noise robustness, and generalization across varying loads compared with conventional \ac{CNN}-based models, as validated on two datasets. Additionally, feature-space visualization (based on t-distributed stochastic neighbor embedding) reveals physically consistent clusters aligned with bearing fault mechanisms.
A conceptually similar, yet structurally different approach is proposed by~\cite{gao2024MPINet:MultiscalePhysicsInformed}, who use a set of subnetworks to target small-sample learning. Each subnetwork is tailored to a specific failure mode (class). Based on the corresponding bearing fault characteristic frequency, a kernel that simulates an ideal bearing fault vibration replaces the data-driven kernel of the first convolutional layer of an otherwise conventional \ac{CNN}, while the healthy-condition subnetwork retains a standard data-driven first layer. These subnetworks are trained independently as binary classifiers to specialize in fault-specific feature extraction from vibration signals. Fusing the extracted features and feeding them to a unifying classifier enables multi-class diagnosis of rolling bearings. Validated on two datasets, the proposed model improves small-sample accuracy compared to both classical baselines and \ac{CNN}-based models.

% Wind turbine
In their work, \cite{jin2024GraphSpatioTemporalNetworks} propose a spatio-temporal \ac{GNN} for wind turbine condition monitoring. They construct a directed graph based on prior knowledge of turbine structure and inter-component causal relationships, with nodes representing sensor signals and edges encoding structural and causal connections. On this prior graph, a graph attention network layer captures spatial dependencies between signals, while global-local attention and recurrent layers model their temporal evolution under healthy operation. The network is trained to predict the next-step values of all signals, and deviations between predicted and actual values are monitored at both the graph and node levels. By doing node-level anomaly detection, it does perform component-level fault localization, which is a practically meaningful form of diagnosis. A multi-node fault propagation chain, constrained by the prior graph, is used to distinguish true faults from scattered false alarms and to identify the origin of abnormal behavior. Case studies on a real turbine with a generator bearing failure and several false alarms show comparable early warning capability, fewer false alarms, and more interpretable monitoring compared with purely data-driven \ac{DL} baselines.

\subsubsection{Health Assessment}
\label{subsub:health_assessment_inductive_bias}

\begin{table*}[ht]
\centering
\caption{All studies employing \textit{inductive bias} to address \textit{health assessment}, listed in alphabetical order.}
\label{tab:inductive_bias_health_assessment}
\begin{tabular}{@{}
  >{\RaggedRight\arraybackslash}p{0.24\textwidth}%
  >{\RaggedRight\arraybackslash}p{0.31\textwidth}%
  >{\RaggedRight\arraybackslash}p{0.23\textwidth}%
  >{\RaggedRight\arraybackslash}p{0.18\textwidth}%
  @{}}
\multicolumn{4}{@{}l@{}}{\bfseries Inductive Bias for Health Assessment} \\
\hline
\textbf{Reference} 
& \multicolumn{2}{l}{\textbf{Prior Physical Knowledge}} 
& \textbf{Use Case} \\
& Type 
& Representation
& \\
\hline

\cite{kristupasbajarunas2024HealthIndexEstimation} & Causal relationship (sensors, operating conditions, degradation) & Architectural constraint (and corresponding regularization terms) & Turbofan engine, lithium-ion battery \\

\cite{cheng2024Researchongas} & Thermodynamic and structural relations of the gas turbine & Adjacency matrix & Gas turbine \\

\cite{ellis2022Ahybridframework} & Model relating crack length and first natural frequency & Non-stationary \acs{GP} prior & Turbomachine rotor blade \\

\cite{fu2024PhysicsInformedNeuralNetwork} & Second-order \acs{ECM} & Parametric state-space model & Lithium-ion battery \\

\cite{hao2023Anoveldeep} & Monotonicity assumption & Activation function & Milling \\

\cite{huang2022AnEnhancedDataDriven} & Empirical degradation model & Algebraic equation & Lithium-ion battery \\

\cite{lehmann2024LearningtheAgeing} & Cyclic stress model, calendric stress model and aging curve model & Algebraic equations & Lithium-ion battery \\

\cite{li2025Applicationofphysicsguided} & Monotonicity assumption & Architectural constraints & Milling \\

\cite{ma2023Ahybriddrivenprobabilistic} & Degradation dynamics (Wiener process) & Stochastic discrete-time state-space model & Milling \\

\cite{qin2025ManagingBatteryPerformance} & First-order \acs{ECM} with hysteresis and Coulomb counting model & \acs{ODE}-based layers & Lithium-ion battery \\

\cite{xie2024DegradationStateAssessment} & Monotonically increasing one-dimensional degradation state in the range $[0,1]$ & Architectural constraint (with associated loss terms) & Semiconductor (insulated gate bipolar transistor) \\

\cite{yucesan2019Windturbinemain, yucesan2020Ahybridmodel, yucesan2021Hybridphysicsinformedneural, yucesan2022Ahybridphysicsinformed, yucesan2023Physicsinformeddigitaltwin} & Cumulative fatigue damage model with lubricant influence & First-order \acs{ODE}, algebraic equations & Bearing \\

\cite{zhu2023PhysicsinformedGaussianprocess} & Empirical tool wear models & Algebraic equations & Milling \\

\hline
\end{tabular}
\end{table*}

% Lithium-ion batteries
Inductive bias is used in 17 studies on health assessment (see Tab.~\ref{tab:inductive_bias_health_assessment}). To estimate the \ac{SOH} of lithium-ion batteries, \cite{huang2022AnEnhancedDataDriven} design a custom kernel for \ac{GPR} based on an empirical degradation model that captures both linear and exponential decay over cycling. The kernel hyperparameters are initialized using the parameters of the empirical model (identified via least squares), facilitating subsequent optimization. Benchmarking against standard Gaussian and Matérn kernels shows that the proposed kernel consistently achieves lower prediction errors on two experimental datasets.
\cite{fu2024PhysicsInformedNeuralNetwork} propose a \ac{RNN} whose architecture is derived from a discretized second-order \ac{ECM}, in which the open-circuit voltage is modeled by an \ac{MLP}. Treating the \ac{ECM} parameters as trainable variables enables accurate online parameter identification and terminal-voltage prediction with low computational complexity. Experimental results demonstrate that the proposed method outperforms both a white-box neural circuit model and an enhanced \ac{ECM}. Further analysis reveals a linear relationship between the identified ohmic resistance and remaining capacity, enabling accurate \ac{SOH} estimation.
Also addressing \ac{SOH} estimation of lithium-ion batteries, \cite{lehmann2024LearningtheAgeing} propose a stress-factor-based aging model that is parameterized using both laboratory aging tests and data from an electric bus fleet. The model consists of cyclic and calendric stress maps and a square-root aging curve. The stress maps are represented by \acp{NN} that share this fixed aging-curve structure with a separately fitted empirical model. Capacities predicted by the empirical model at collocation points are included in the training objective, which ties the learned stress maps to the established functional relationships in regions that are sparsely covered by data. Using \ac{TL} to adapt the aging-curve parameters to different cell types and to the fleet, the approach improves capacity and \ac{SOH} estimates at check-up tests compared with an equal-stress baseline and with the coupled model without adaptation.
For lithium-ion battery health management, \cite{qin2025ManagingBatteryPerformance} propose a predictor-estimator framework that jointly handles \ac{SOC}, internal resistance, and maximum available capacity. A linear-exponential two-stage degradation model with recursive least-squares updating and Bayesian knee-point detection predicts the evolution of several health indicators (\ac{SOH}, a resistance-based \ac{HI}, efficiency, and \ac{SOC} at discharge start), from which future capacity and resistance trajectories are derived. Three lightweight neural estimators then infer \ac{SOC}, resistance, and capacity from operational time series, where physics is embedded by hard-wiring Coulomb counting dynamics with learnable efficiency and initial-\ac{SOC} corrections for \ac{SOC} estimation. The second estimator couples an \ac{MLP} with an equivalent-circuit \ac{ODE} model whose parameters are learned under physical bounds for resistance estimation, while the capacity estimator exploits these latent variables via channel-attention and temporal convolutions. Experiments on two public battery datasets show that this framework provides more accurate degradation prediction and state estimation than state-of-the-art baselines, generalizes well across different chemistries, temperatures, and loading conditions, and achieves these benefits with very compact estimator networks and low computational overhead.

% Bearings
\cite{yucesan2019Windturbinemain} propose a framework for estimating wind-turbine main bearing fatigue life by combining a physics-based cumulative-damage model for bearing fatigue with a data-driven model for grease degradation, organized within a recurrent model that models damage accumulation over time. Bearing fatigue is determined incrementally at each timestep by computing the fatigue damage rate from the current load and speed using an L10-based bearing life relation and accumulating damage via Palmgren-Miner’s rule. Simultaneously, an \ac{MLP} predicts increments in grease degradation---a hidden variable affecting fatigue calculations through viscosity and contamination factors---but is only indirectly supervised through periodic grease observations. By embedding these components within a custom recurrent cell, the model captures both the well-understood physics of bearing fatigue and the complex dynamics of grease degradation that are difficult to model from first principles. While promising, the results provide no quantitative performance metrics and lack baseline comparisons, limiting the ability to fully assess the approach's effectiveness. Even though the model is elaborated upon in subsequent works~\citep{yucesan2020Ahybridmodel, yucesan2021Hybridphysicsinformedneural, yucesan2022Ahybridphysicsinformed, yucesan2023Physicsinformeddigitaltwin}, the fundamental approach remains unchanged from a \ac{PIML} perspective.

% Milling (cutting tools)
Studying tool wear estimation in high-speed milling, \cite{li2025Applicationofphysicsguided} propose embedding an architectural constraint into a \ac{GRU} network. To account for the irreversible nature of tool wear, monotonicity is enforced via constrained hidden-state updates and using \ac{ReLU} activations. This ensures that predicted wear values cannot decrease over time, which improves both physical plausibility and predictive accuracy compared to unconstrained data-driven models. See Section~\ref{subsub:health_assessment_learning_bias} for a description of the learning bias incorporated by~\cite{li2025Applicationofphysicsguided}.
A similar approach is adopted by~\cite{hao2023Anoveldeep}, where a softplus activation function is incorporated into the network architecture prior to the output layer. This design choice explicitly accounts for the monotonic degradation behavior of milling tool wear, ensuring nondecreasing wear predictions over time. As a result, the proposed model likewise achieves improved predictive accuracy compared to purely data-driven baseline approaches. 
\cite{zhu2023PhysicsinformedGaussianprocess} propose a \ac{GPR}-based approach for predicting tool wear. Prior knowledge is incorporated through three physics-based tool wear models (generalized Taylor formula, a cubic polynomial wear-time law, and a generic flank wear model) that describe the evolution of tool flank wear over cutting time. By adopting these wear laws as the prior mean function, the model supports small-sample training, continual online updating, and substantially improved extrapolation. Comparative experiments confirm that the proposed approach significantly reduces prediction error and improves robustness relative to both standalone physics-based models and purely data-driven approaches.
\cite{ma2023Ahybriddrivenprobabilistic} also address tool wear monitoring in milling. A physics-based state-space model based on a Wiener process-inspired wear law serves as prior knowledge within a \ac{GP}-based probabilistic state-space model. A \ac{PF} estimates the unknown posterior distribution of the degradation state. Compared with data-driven baselines such as \ac{CNN}, \ac{LSTM}, and \ac{SVR}, as well as purely physics-based approaches, the method yields a better predictive performance and tighter confidence intervals for tool replacement decisions. The method is also suitable for prognosis tasks and \ac{RUL} prediction.

% Turbofan engine + battery
\cite{kristupasbajarunas2024HealthIndexEstimation} aim to leverage general knowledge about degradation to broaden the applicability of unsupervised \ac{HI} estimation across various systems. They use assumed causal relationships between sensor readings, operating conditions, and the degradation to design the architecture of a convolutional \ac{AE}: the encoder maps sensor readings to a scalar latent variable, while the decoder reconstructs sensor readings from this latent variable and operating conditions, thereby forcing the latent variable to encode degradation information. To further shape this latent variable into a suitable \ac{HI}, they complement this inductive-bias approach with loss terms that guide \ac{HI} properties desired in \ac{PHM}, such as monotonicity and trendability, as well as an optional term that encourages consistency with degradation trends derived from reliability theory. The approach is validated on both turbofan engine data and lithium-ion battery data, leading to improvements in prediction performance and out-of-distribution robustness compared to residual-based baselines. With a subsequent \ac{CNN}, the approach is extended for \ac{RUL} prediction, yet the \ac{PIML} aspects in this paper solely correspond to health assessment.

% Electrical components (semiconductor)
To assess the degradation state of insulated gate bipolar transistor modules under varying operating conditions, \cite{xie2024DegradationStateAssessment} propose an \ac{LSTM}-based \ac{AE} designed to disentangle degradation from operating conditions. The encoder compresses inputs into a low-dimensional latent vector, reserving a single scalar for the degradation index while the remaining dimensions capture variations in operating conditions. Two decoders are then employed: one reconstructs non-degraded behavior from the condition-related latents alone, and the other reconstructs degraded behavior using the full latent vector. This design forces degradation information to flow through a single neuron and yields an interpretable \ac{HI} suitable for online monitoring. Complementing this inductive-bias approach, the authors employ additional loss terms that guide this single health-indicator neuron to satisfy both monotonicity and range constraints. This leads to better results in terms of prediction performance and physical consistency compared to the inductive-bias approach alone.

% Gas turbine
\cite{cheng2024Researchongas} assess gas turbine health via a spatio-temporal \ac{GNN}. Thermodynamic and structural knowledge is encoded by constructing a temporal graph over key monitored parameters, whose edges combine \ac{kNN}-based data correlations with links derived from small-deviation compressor, combustor, and turbine equations. This topology effectively constrains how information propagates between variables, enabling accurate health assessment by mapping multivariate time series to discrete health stages, as corroborated by corresponding ablation studies.

% Turbomachine rotor blade
As part of a broader framework, \cite{ellis2022Ahybridframework} address the health assessment of turbomachine rotor blades by estimating root crack length from blade tip timing-derived natural frequencies under scarce inspection data. A physics-based model (\ac{FE} simulations with an unscented transform) is first built to map natural frequency to crack length, and this ensemble is then used as the prior mean and covariance of a \ac{GPR} model. The latter is conditioned on crack-length measurements obtained via non-destructive testing during routine maintenance. The proposed model preserves physically plausible behavior in data-sparse regions while correcting systematic errors near observed non-destructive testing points, and it consistently outperforms both the pure physics-based model and several purely data-driven regressors.

\subsubsection{Prognosis}
\label{subsub:prognosis_inductive_bias}

\begin{table*}[ht]
\centering
\caption{All studies employing \textit{inductive bias} to address \textit{prognosis}, listed in alphabetical order.}
\label{tab:inductive_bias_prognosis}
\begin{tabular}{@{}
  >{\RaggedRight\arraybackslash}p{0.24\textwidth}%
  >{\RaggedRight\arraybackslash}p{0.31\textwidth}%
  >{\RaggedRight\arraybackslash}p{0.23\textwidth}%
  >{\RaggedRight\arraybackslash}p{0.18\textwidth}%
  @{}}
\multicolumn{4}{@{}l@{}}{\bfseries Inductive Bias for Prognosis} \\
\hline
\textbf{Reference} 
& \multicolumn{2}{l}{\textbf{Prior Physical Knowledge}} 
& \textbf{Use Case} \\
& Type 
& Representation
& \\
\hline

\cite{abiria2025Highcycleandveryhighcycle} & Basquin's law, nonnegativity assumption & Algebraic equations, differential equations, architectural constraints & Additive manufacturing \\

\cite{badora2023Usingphysicsinformedneural} & Paris' law & Differential equation & High-pressure nozzle of an industrial gas turbine \\

\cite{bai2023PrognosticsofLithiumIon} & Bounded and monotonically decreasing capacity fade over cycles & Inequality constraints & Lithium-ion battery \\

\cite{cai2025Knowledgeembeddedspatial} & System/sensor topology & Knowledge graph & Turbofan engine, milling \\

\cite{dourado2019Physicsinformedneuralnetworks, dourado2022Ensembleofhybrid} & Walker model for fatigue crack propagation & Recurrent algebraic equation & Aircraft fuselage panels \\

\cite{jiang2025PhysicsinformedGaussianprocess} & Paris' law for fatigue crack growth & \acs{ODE} discretized to a damage accumulation model & Aluminum specimens \\

\cite{nascimento2021Hybridphysicsinformedneural} & Nernst and Butler-Volmer equations & Algebraic equations & Lithium-ion battery \\

\cite{nguyen2023Physicsinfusedfuzzygenerative} & Spall-growth model, modified Eyring model & Algebraic equation & Bearing, turbofan engine \\

\cite{qiang2023Integratingphysicsinformedrecurrent} & Empirical linear tool wear model & Algebraic equation & Milling \\

\cite{qin2024AnInterpretableNeuroDynamic} & Power equations of low- and high-pressure compressor, high-speed shaft dynamics & Algebraic equation, \acs{ODE} & Turbofan engine \\

\cite{yin2025Physicsguideddegradationtrajectory} & Monotonicity assumption & Architectural constraints & Bearing \\

\cite{zhang2025Applicationofphysicsinformed} & Diamond-shaped wear particle model, Archard wear model and wear model by \cite{zou1996abrasivewearmodel} & Algebraic equations & Axial piston pump \\

\cite{zhou2023Timevaryingtrajectorymodeling} & Nonnegativity assumption & Architectural constraint & Turbofan engine, bearing \\

\cite{zhou2025Physicsinformedspatiotemporalhybrid} & Thermal cycle model & Fixed sensor association graph (adjacency matrix) & Turbofan engine \\

\hline
\end{tabular}
\end{table*}

% Lithium-ion batteries
Table~\ref{tab:inductive_bias_prognosis} provides an overview of the 15 identified studies regarding inductive bias for prognosis. \cite{nascimento2021Hybridphysicsinformedneural} propose a method for lithium-ion batteries that embeds core electrochemical relations to predict voltage discharge curves (and thereby end-of-discharge time) under varying loads and to forecast aging-induced capacity fade and resistance growth. A reduced-order model based on the Nernst and Butler-Volmer equations is implemented as a recurrent cell, while \acp{MLP} replace the non-ideal voltage (activity) terms that are difficult to capture analytically. The framework treats lumped internal resistance and maximum available charge as cell- and age-dependent parameters, and models their evolution with cumulative discharged energy using variational ensemble learning to obtain quantitative aging indicators and uncertainty-aware forecasts. Despite extensive experiments, no quantitative results are reported that demonstrate improvements over purely data-driven baselines. Methodologically, this approach is closely related to the work of~\cite{yucesan2019Windturbinemain} and its follow-up studies (see Sec.~\ref{subsub:health_assessment_inductive_bias}).
\cite{bai2023PrognosticsofLithiumIon} address lithium-ion battery capacity prognostics using a two-stage framework. In the first stage, an \ac{NN} combined with a dual \ac{EKF} is employed for online estimation of the \ac{SOC} and capacity from voltage and current measurements, producing capacity trajectories. In the second stage, these trajectories are used as inputs to a \ac{GPR}-based degradation model to forecast future capacity evolution. Two inequality constraints are imposed on the \ac{GPR}, ensuring that capacity predictions remain bounded and monotonically decreasing with respect to the cycle number. The constrained \ac{GPR} outperforms its unconstrained version, as well as three additional baseline methods in terms of both predictive accuracy and reduced uncertainty. 

% Bearings
Addressing \ac{RUL} prediction for rolling bearings, \cite{yin2025Physicsguideddegradationtrajectory} focus on physically consistent modeling of degradation. The paper proposes a method that leverages phase space reconstruction to transform vibration signals into trajectories, turning \ac{RUL} prediction into a variation estimation problem. Prior knowledge about the monotonic nature of bearing degradation is integrated into a 1D-\ac{CNN}'s final activation function, ensuring that the predicted \ac{RUL} cannot increase unrealistically. This improves robustness, smoothness, and physical plausibility of predictions compared to purely data-driven models. Comparative experiments under varying working conditions demonstrate that the approach outperforms state-of-the-art methods in both predictive accuracy and generalizability.
% Bearing + turbofan
\cite{nguyen2023Physicsinfusedfuzzygenerative} address the limitations of purely data-driven \acp{GAN} for \ac{RUL} prediction, including instability, sample inefficiency, and lack of physical consistency. Their proposed architecture attaches a differentiable fuzzy logic module to the output of a conditional \ac{GAN}'s generator. Within this module, the standard product aggregation operator is replaced by a dataset-specific physics model (a spall-growth model for bearings; a modified Eyring model for turbofan engines). The generator thus learns fuzzy implications whose values serve as parameters of the physics model, constraining predictions to physically realistic solutions while the adversarial training signal still flows end-to-end through both the fuzzy and physics layers. Experiments on two different datasets (bearings and turbofan engines) show reduced prediction errors. Additional experiments, in which the dataset size is iteratively reduced, demonstrate improved data efficiency compared to a conventional \ac{LSTM}.
\cite{zhou2023Timevaryingtrajectorymodeling} target \ac{RUL} prediction as a time-varying trajectory modeling problem rather than a point-wise estimation. Building on Neural~ODE~\citep{chen2018Neuralordinarydifferentialequations}, the proposed approach integrates additional prior physical knowledge about smooth degradation behavior. Physically meaningful \ac{RUL} trends are enforced by incorporating a nonnegative bounded function prior to providing the final prediction. Furthermore, a dynamic learning scheme utilizing a super-network \citep{wu2021neuralarchitecturesearchassparsesupernet} and deep \ac{RL} enables adaptive time-dependent network architectures, improving the model’s ability to capture underlying degradation dynamics. This approach results in more stable and interpretable \ac{RUL} predictions that align with real-world degradation processes. The proposed method shows smooth and accurate prediction results compared to common data-driven methods such as \ac{ResNet} and \ac{BiLSTM}, as demonstrated through experiments on both bearing and turbofan engine datasets.

% Milling (cutting tools)
\cite{qiang2023Integratingphysicsinformedrecurrent} study tool wear prediction in milling under varying cutting parameters, where new operating conditions offer only limited labeled wear data. To tackle this challenge, an instance-based regression transfer algorithm (Two-stage TrAdaBoost.R2, proposed by~\cite{pardoe2010boosting}) is combined with a recurrent \ac{GPR} base learner. The \ac{GPR}'s mean function encodes empirical physical relations between flank wear, cutting power, and previous wear to capture time-accumulation and degradation trends, while the kernel models residual nonlinearities. Using only about 30\,\% of early-life wear data, the framework accurately extrapolates the full wear trajectory and yields tight confidence intervals. The proposed framework is evaluated against three alternative approaches: an otherwise identical framework employing a standard (uninformed) \ac{GPR}, a recurrent \ac{GPR}, and an \ac{LSTM}, with the latter two not incorporating \ac{TL}. Experiments show that the proposed framework achieves substantially lower prediction errors, better tracking of late-life wear growth, and more stable performance across different cutting-parameter combinations.

% Turbofan/Milling
Unlike many studies that rely only on temporal sensor sequences, \cite{cai2025Knowledgeembeddedspatial} take spatial interactions among multiple sensors into account. Two use cases (milling and turbofan engines) are considered to leverage system topology and sensor placement information. In a first step, embeddings are learned that represent the real-world topological structure. This is done by employing an energy-based knowledge embedding algorithm. Distances between embeddings are used to construct graph edges and an initial weighted adjacency matrix, whose weights are then dynamically updated via an attention mechanism. These are fed into a model consisting of spatial modules (graph convolutional network and attention mechanism) as well as temporal modules (\ac{LSTM}) and a final fully connected layer for direct \ac{RUL} prediction. In ablation scenarios considering the turbofan use case, it is shown that temporal modules, spatial modules, and the knowledge encoding in the adjacency matrix contribute to the model performance. In comparison with multiple baselines (e.g., \ac{CNN}, \ac{LSTM}, and Bayesian models), the proposed method achieves superior performance in direct \ac{RUL} prediction tasks in the majority of scenarios. Comparable results are also reported for the milling use case.
% Turbofan
\cite{qin2024AnInterpretableNeuroDynamic} address the inadequate interpretability prevalent in \ac{DL} methods applied to \ac{RUL} estimation, with a particular focus on turbofan engines. The proposed model comprises an augmenter for noise filtering, interpolation, and unobservable state estimation, and an estimator for \ac{RUL} prediction. The augmenter is based on a Neural \ac{ODE} framework that integrates physical models, data-driven models, and a Runge-Kutta \ac{ODE} solver, all trained end-to-end. The estimator combines \ac{LSTM}-based encoding with feature and temporal attention. As prior knowledge, power equations of the low-pressure compressor and high-pressure compressor, shaft-speed dynamics, and stall-margin and efficiency-modifier equations are embedded into the augmenter. Compared to a variety of data-driven baselines, the proposed method proves to be superior in terms of predictive performance.
Also targeting the \ac{RUL} prediction of turbofan engines, \cite{zhou2025Physicsinformedspatiotemporalhybrid} embed thermodynamic prior knowledge into a spatio-temporal \ac{NN} via graph construction. In the spatial branch, thermodynamic cycle and engine-structure knowledge are used to build an association graph between sensors, which is fused with a gray-relation graph to form a fixed adjacency matrix for a multilayer graph attention network with pooling. In the temporal branch, an \ac{LSTM} extracts temporal features, while a temporal-pattern attention module derives time-invariant features from its hidden states, which together form the temporal-domain features. An attention module then fuses temporal and spatial features for \ac{RUL} prediction. Experiments on C-MAPSS show that the proposed model achieves consistently lower prediction errors than standard \ac{DL} and other spatio-temporal baselines.

% Pumps
\cite{zhang2025Applicationofphysicsinformed} address the degradation of hydraulic piston pumps, which they characterize as arising from the interplay of several factors, most notably the progressive wear of internal friction pairs together with fluctuating external load and operating conditions. To capture this behavior, they propose a framework for predicting the \ac{RUL} of hydraulic piston pumps built around an \ac{LSTM}. Wear laws for three key friction pairs (valve plate, piston, slipper) are embedded in an end-to-end design, with wear model parameters updated jointly with the \ac{LSTM} weights. These adaptive wear models convert monitoring data into degradation indicators, which the LSTM uses to predict return oil flow. \ac{RUL} is estimated as the time until the predicted flow exceeds a critical threshold. The proposed method outperforms conventional baselines (including \ac{SVR}, \ac{RF}, and \ac{LSTM}), though more advanced prognostics methods were not evaluated.

% Fatigue - materials
Aging aircraft fleets are studied in the works of~ \cite{dourado2019Physicsinformedneuralnetworks, dourado2022Ensembleofhybrid}, with a specific focus on fatigue crack propagation in aircraft fuselage panels. In both contributions, a physical fatigue crack propagation model (the Walker model, an adapted version of the well-known Paris law) is embedded directly into a custom recurrent cell, while a data-driven branch within the cell learns a correction term that accounts for corrosion effects not captured by the Walker model. However, the authors do not benchmark their method against purely data-driven or purely physics-based baselines, making it difficult to quantitatively assess the added value of the proposed approach. Similar to~\cite{nascimento2021Hybridphysicsinformedneural}, this approach is also methodologically related to the work of~\cite{yucesan2019Windturbinemain} and its follow-up studies. 
\cite{jiang2025PhysicsinformedGaussianprocess} study probabilistic prognosis of fatigue crack growth in metallic specimens. Monte Carlo simulations of a physics-based model (Paris' law) are truncated using a standard \ac{GP} fitted to current observations, whereas the resulting trajectories are used to construct a non-stationary prior mean and covariance for the final \ac{GP}. Experiments show that incorporating these priors markedly improves extrapolation performance and predictive accuracy compared to both a standard \ac{GP} and a \ac{PF}.
% fatigue - nozzle
\cite{badora2023Usingphysicsinformedneural} introduce a custom \ac{RNN} cell designed to model fatigue crack growth in a gas turbine nozzle. An \ac{NN} estimates the stress intensity factor range at shutdown, while a physics-based part then applies Paris' law to compute the crack length increment due to fatigue. The model accurately predicts crack growth over multiple cycles, even with limited observed data, and outperforms standard regression models in terms of predictive performance.
% fatigue - additive manufacturing
In their work, \cite{abiria2025Highcycleandveryhighcycle} tackle the challenge of predicting fatigue life in additively manufactured alloys, where cyclic loading leads to microscopic damage accumulation and eventual failure. The authors address this problem by integrating prior knowledge into a conventional \ac{NN} through modified activation functions derived from Basquin’s law, a modified Paris law, and a nonnegativity condition. The proposed model shows better generalization capabilities compared to other physics-informed and purely physics-based variants.

% - - - - - - - - - - subsection - - - - - - - - - - %
\subsection{Learning Bias}
\label{subsec:learning_bias}

% Outlook for the section to follow
Learning bias is characterized by introducing prior physical knowledge into the learning algorithm, frequently realized via a composite loss function. Such a composite loss typically combines a data-fidelity term, which minimizes the discrepancy between predictions and observations, with one or more physics-informed terms that penalize violations of the governing equations, boundary conditions, or other physical constraints, each scaled by a weighting coefficient that controls its relative influence on training. The identified studies are organized 
by PHM task in 
Tables~\ref{tab:learning_bias_fault_detection}--\ref{tab:learning_bias_prognosis}. A representative example is illustrated in Figure~\ref{fig:prime_example_learning_bias}, which includes a composite loss function.

\subsubsection{Fault Detection}
\label{subsub:fault_detection_learning_bias}

\begin{table*}[ht]
\centering
\caption{All studies employing \textit{learning bias} to address \textit{fault detection}, listed in alphabetical order.}
\label{tab:learning_bias_fault_detection}
\begin{tabular}{@{}
  >{\RaggedRight\arraybackslash}p{0.24\textwidth}%
  >{\RaggedRight\arraybackslash}p{0.31\textwidth}%
  >{\RaggedRight\arraybackslash}p{0.23\textwidth}%
  >{\RaggedRight\arraybackslash}p{0.18\textwidth}%
  @{}}
\multicolumn{4}{@{}l@{}}{\bfseries Learning Bias for Fault Detection} \\
\hline
\textbf{Reference} 
& \multicolumn{2}{l}{\textbf{Prior Physical Knowledge}} 
& \textbf{Use Case} \\
& Type 
& Representation
& \\
\hline

\cite{wang2025Physicallyinformedhierarchical} & Wave-type vibration model of spline shaft & Nonhomogeneous 1D wave \acs{PDE} & Aero-engine involute spline coupling \\

\cite{wang2024Adigitaltwin} & Multi-energy model of a robot joint, multi-joint rigid-body dynamics & Algebraic energy-balance equations and recursive Newton-Euler dynamic equations & Industrial multi-axis robot \\

\cite{xu2024Physicsguideddeeplearning} & Damage index model, monotonic stress-strain relation & Algebraic equations & Carbon fiber reinforced polymer laminates \\

\hline
\end{tabular}
\end{table*}

% Composites (carbon fiber reinforced plastics)
Table~\ref{tab:learning_bias_fault_detection} lists the three studies that employ learning bias for fault detection. \cite{xu2024Physicsguideddeeplearning} target Lamb-wave-based fatigue damage detection in carbon fiber reinforced polymer laminates by augmenting a \ac{CNN} with an additional branch that estimates global stiffness degradation from time-frequency images of guided-wave signals obtained via \ac{CWT}. Their approach incorporates additional loss terms based on a damage index model and a stress-strain constraint, thereby regularizing training toward physically consistent progressive degradation. The damage index model relates stiffness degradation and off-axis angle to normalized power spectral density changes of Lamb-wave responses, converting the network’s predicted stiffness into pseudo-damage labels. The stress-strain constraint enforces monotonically increasing strain under constant load, penalizing non-monotonic strain trajectories derived from the predicted stiffness. Using only data from one carbon fiber reinforced polymer layup, the method generalizes to unseen layups with substantially improved cross-structure detection performance compared to a standard \ac{CNN}, and the resulting path-level damage predictions support accurate delamination localization.

% Aero-engine involute spline coupling
\cite{wang2025Physicallyinformedhierarchical} propose a soft sensing framework for estimating difficult-to-measure aero-engine variables. The approach extends \acp{PINN} to nonhomogeneous \acp{PDE} with unknown, unmeasurable driving terms by training two coupled \acp{NN} in a hierarchical (alternating) optimization scheme: one approximates the \ac{PDE} solution, the other the unknown source term. By employing a joint loss in which the learned source is embedded in the \ac{PDE} residual, the solution is regularized toward \ac{PDE}-consistent behavior, while the source is simultaneously constrained to produce driving terms that are compatible with both the measurements and the governing equation. A recurrent-prediction term further refines the solution using delayed hard-sensor and soft-sensor outputs to mitigate information loss due to sparse sampling and unmeasured sources. Applied as a virtual vibration sensor on an aero-engine spline-coupling test rig, the method achieves substantially lower prediction errors than standard \acp{PINN} and supports a proof-of-concept anomaly detection example for spline-coupling health monitoring.

% Industrial robots
By using a convolutional \ac{AE}, \cite{wang2024Adigitaltwin} estimate joint electrical current from multivariate sensor data (e.g., motion, temperature, and vibration) for anomaly detection in industrial robot systems. Physical knowledge from energy conservation in the joints (multi-energy model) and Newton-Euler multi-joint dynamics is incorporated via additional loss terms, enforcing energy-flow consistency and torque coupling between joints. Using the Kullback-Leibler divergence between estimated and measured current as a health indicator, the proposed method detects injected motor and reducer faults on real factory robots with superior accuracy, outperforming several state-of-the-art time-series anomaly detection methods. 

\subsubsection{Diagnosis}
\label{subsub:diagnosis_learning_bias}

\begin{table*}[ht]
\centering
\caption{All studies employing \textit{learning bias} to address \textit{diagnosis}, listed in alphabetical order.}
\label{tab:learning_bias_diagnosis}
\begin{tabular}{@{}
  >{\RaggedRight\arraybackslash}p{0.24\textwidth}%
  >{\RaggedRight\arraybackslash}p{0.31\textwidth}%
  >{\RaggedRight\arraybackslash}p{0.23\textwidth}%
  >{\RaggedRight\arraybackslash}p{0.18\textwidth}%
  @{}}
\multicolumn{4}{@{}l@{}}{\bfseries Learning Bias for Diagnosis} \\
\hline
\textbf{Reference} 
& \multicolumn{2}{l}{\textbf{Prior Physical Knowledge}} 
& \textbf{Use Case} \\
& Type 
& Representation
& \\
\hline

\cite{chao2025Physicsinformedneural} & Discharge-pressure and internal leakage model, volumetric efficiency definition & Nonlinear \acs{ODE}, algebraic equation & Axial piston pump \\

\cite{dong2025Innovativefaultdiagnosis} & Mass and momentum conservation in fluid dynamics, valve and periodic boundary conditions & \acsp{PDE}, algebraic equations & Axial piston pump \\

\cite{huang2025Physicsinformedcausallearning} & Structural causal model for gearbox vibration data & Algebraic equation & Planetary gearbox, wind turbine gearbox \\

\cite{li2024Hybridphysicsembeddedrecurrent} & Robot dynamic model & Second-order \acs{ODE} & Industrial robot \\

\cite{qiao2024APriorKnowledge} & Bearing fault frequencies & Numerical values & Bearing \\

\cite{sun2024Contrastivelearningand} & 4-\acs{DOF} multi-body bearing dynamics model & Second-order \acs{ODE} system & Bearing \\

\cite{tang2024Apriorknowledgeenhanced} & Demand for consistent time and frequency domain latent representations & Invariance loss & Bearing, gearbox \\

\cite{xu2024Physicsinformedprobabilisticdeep} & Bearing fault frequencies & Algebraic equations & Bearing \\

\cite{zhu2024PhysiCausalNet:ACausaland} & Bearing dynamic model, domain-invariant features per fault case & Second-order \acs{ODE}, progressive consistency causal factorization loss & Bearing \\

\hline
\end{tabular}
\end{table*}

% Bearing
Nine studies employing learning bias for diagnosis have been identified, as summarized in Table~\ref{tab:learning_bias_diagnosis}. \cite{qiao2024APriorKnowledge} focus on the small-sample problem in bearing fault diagnosis under variable operating conditions. A 1D-\ac{CNN} with a sequential temporal attention module is trained within a contrastive learning framework to obtain discriminative fault representations from vibration signals. Prior knowledge enters as analytically derived fault characteristic frequencies, which are provided as additional targets and predicted from the learned embedding via an auxiliary fully connected head. The loss between predicted and known characteristic frequencies, weighted within a composite loss alongside cross-entropy and contrastive loss, softly enforces that the latent representation encodes these physically meaningful frequency features, thereby guiding the network toward fault-relevant structure in the data. The resulting model outperforms purely data-driven baselines on two publicly available bearing datasets, with ablation studies confirming the benefit of the embedded learning bias, particularly in small-sample scenarios.
To address distribution shifts arising from variations in bearing structure and operating conditions, \cite{zhu2024PhysiCausalNet:ACausaland} propose a domain generalization method capable of extracting domain-invariant features for fault diagnosis. The model consists of a Fourier-based low-pass filtering module with learnable parameters, a \ac{CNN}-based feature extractor and a fully connected classifier. Two kinds of regularization terms are incorporated: a dynamic embedding loss to guide features that account for the state of the target machine's bearing dynamics and a progressive consistency causal factorization loss. The latter guides correlation for features of the same fault case across different machines and operating conditions, while discouraging correlation between features of different fault cases. Therefore, domain-invariant features can be realized. Experiments show that the proposed approach demonstrates superior performance in terms of accuracy, generalizability, and interpretability in comparison to common domain generalization methods.
\cite{xu2024Physicsinformedprobabilisticdeep} also incorporate fault characteristic frequencies of bearings into their approach. They use a dual-branch \ac{AE} to reconstruct bearing vibration data in the frequency domain for both real and imaginary parts of the spectrum. A modified loss function that makes use of masked target data is used for reconstruction. The mask highlights bands relevant to fault frequencies. Thus, a robust latent space that is biased toward fault frequencies relevant to fault cases is obtained as input for a subsequent classifier. Their approach outperforms state-of-the-art methods in terms of accuracy and well-separated feature spaces.
To effectively tackle label-free fault diagnosis for rolling bearings, the proposed framework~\citep{sun2024Contrastivelearningand} combines a contrastive-learning backbone with a dynamics-embedding network based on sparse identification of nonlinear dynamics~\citep{champion2019sindy}: a coordinate encoder reconstructs 4-\ac{DOF} latent states from 1-\ac{DOF} acceleration via delay embedding, and a physics-based equation library derived from a 4-\ac{DOF} multi-body bearing model is used together with a sparse coefficient matrix to infer the fault type. Physics-based constraints are imposed via the loss function, which enforces consistency between latent accelerations, reconstructed measurement signals and the accelerations predicted by the equation library, enabling the network to learn both the latent dynamics and a sparse, interpretable governing equation from raw signals. Experiments on both simulated and experimental bearing data show that the proposed framework can correctly distinguish between inner-race, outer-race and roller faults without labels while providing physically meaningful diagnostic explanations.

% Bearing and gearbox
\cite{tang2024Apriorknowledgeenhanced} propose a self-supervised learning framework that addresses the small-sample problem in rotating machinery fault diagnosis, studying both bearings and gearboxes. Two \acp{CNN} are used as feature extractors for the time domain and frequency domain, respectively. During pretraining, an additional loss function incorporating a distance metric is employed to obtain time-frequency domain invariant latent embeddings. In the downstream task, the network is fine-tuned on the limited amount of labeled data to learn to diagnose the fault underlying the rotating part, showing superiority over alternative approaches both in terms of accuracy and data efficiency.
% Gearbox
Aiming to incorporate prior knowledge into domain generalization methods, \cite{huang2025Physicsinformedcausallearning} propose a causal learning network based on ResNet18. It is used to extract independent and causal features regarding the relation between the sensor data of gearboxes and distinct fault cases for diagnosis. Losses for an adversarial mask and an autocorrelation matrix, both founded on a structural causal model for vibration data, are incorporated to favor the aforementioned properties. Compared to other domain generalization methods, the approach demonstrates superior performance in terms of accuracy and class-distinguishability.

% Pumps
The integration of model-based knowledge into diagnostic methods proves challenging in the field of axial piston pumps. Therefore, \cite{dong2025Innovativefaultdiagnosis} present a \ac{PINN} framework serving as a high-frequency virtual dynamic flow meter in axial piston pumps by predicting pump flow ripple. The framework integrates fundamental physical principles of hydraulic systems, such as mass and momentum conservation, boundary conditions relevant to pump operation, and periodic characteristics of pump behavior into the loss function. The method facilitates robust pump fault diagnosis. Overall, the study validates the effectiveness of the proposed approach through numerical simulations, demonstrating close agreement with reference solutions, and through experimental investigations, showing that predicted flow ripples consistently reflect expected fault characteristics.
Building on \acp{PINN}, \cite{chao2025Physicsinformedneural} tackle wear detection in axial piston pumps by reconstructing the discharge pressure while simultaneously inferring the fluid film thicknesses at the pump’s main friction pairs. Following the standard \ac{PINN} framework, an analytically derived \ac{ODE} for the time derivative of discharge pressure is incorporated into the loss function, with internal leakage flows scaling cubically with film thickness. To stabilize the joint estimation of multiple wear-related parameters with different scales, these parameters are learned as bounded variables via sigmoid-based range constraints. In a sequential step, the identified film thicknesses are used in analytical formulas for volumetric efficiency and Cohen’s $d$ effect size, providing physically interpretable wear indicators and enabling localization of the worn friction pair. Experiments on a real pump with naturally worn components demonstrate accurate pressure reconstruction, physically plausible thickness estimates, and correct identification of the valve plate and cylinder block pair as the worn pair, although comparisons with alternative methods are not reported.

% Industrial robot
Using available proprioceptive signals instead of external measurements, \cite{li2024Hybridphysicsembeddedrecurrent} propose an approach for fault diagnosis of industrial robots. The framework consists of  an encoder based on \ac{GRU} and fully connected layers, while the decoder combines a robot dynamic model and data-driven residual model. The decoder is solely used in training for reconstruction purposes, while a classifier leverages the latent variable for fault diagnosis both in training and inference. Fault cases of increased joint friction, partial loss of actuator effectiveness, and drivetrain mechanical faults are considered. In ablation studies, the approach performed superiorly to non-\ac{PIML} variants, evaluated with both a simulated UR5 dataset and a real industrial robot in-situ dataset.

\subsubsection{Health Assessment}
\label{subsub:health_assessment_learning_bias}

\begin{table*}[ht]
\centering
\caption{All studies employing \textit{learning bias} to address \textit{health assessment}, listed in alphabetical order.}
\label{tab:learning_bias_health_assessment}
\begin{tabular}{@{}
  >{\RaggedRight\arraybackslash}p{0.24\textwidth}%
  >{\RaggedRight\arraybackslash}p{0.31\textwidth}%
  >{\RaggedRight\arraybackslash}p{0.23\textwidth}%
  >{\RaggedRight\arraybackslash}p{0.18\textwidth}%
  @{}}
\multicolumn{4}{@{}l@{}}{\bfseries Learning Bias for Health Assessment} \\
\hline
\textbf{Reference} 
& \multicolumn{2}{l}{\textbf{Prior Physical Knowledge}} 
& \textbf{Use Case} \\
& Type 
& Representation
& \\
\hline

\cite{deng2025ANovelMethod} & Empirical aging trend & Monotonicity constraint (loss term) & Lithium-ion battery \\

\cite{freeman2022Physicsinformedturbulenceintensity} & Turbulence intensity as analytical and empirical definition & Algebraic equation & Ocean current turbines \\

\cite{jang2025Stateofhealth} & Lumped energy-balance model for battery heat generation & First-order \acs{ODE} & Lithium-ion battery \\

\cite{li2025Applicationofphysicsguided} & Empirical flank tool wear model & Algebraic equation & Milling \\

\cite{liu2025Aphysicsguidedapproach} & Mechanistic \acs{SEI}-growth capacity-fade model & Nonlinear \acs{ODE} & Lithium-ion battery \\

\cite{navidi2024PhysicsInformedMachineLearning} & Half-cell degradation model & Algebraic equations & Lithium-ion battery \\

\cite{pan2025Inservicefatiguecrack} & Paris' law & \acs{ODE} & Fatigue crack propagation (aluminum specimens) \\

\cite{singh2023HybridModelingof} & Fick's second law of diffusion, initial/boundary conditions & \acs{PDE}, algebraic equations & Lithium-ion battery \\

\cite{sun2022MicrocrackDefectQuantification} & Analytical relationships between signal-derived features and crack geometries & Closed-form algebraic equations and inequality constraints & Microcrack quantification in aluminum plate \\

\cite{yonastefera2025ConstraintGuidedLearningof} & Monotonicity assumption, boundary constraint & Algebraic equation & Bearing \\

\cite{wang2025ABatteryState} & Semi-empirical Verhulst degradation model with Arrhenius temperature factor & Nonlinear logistic-type \acs{ODE} & Lithium-ion battery \\

\cite{wangz.2023Physicsinformedneural} & Discharge pressure model describing leakage & First-order \acs{ODE} & Axial piston pump \\

\cite{wang2024Physicalknowledgeguided} & Continuous degradation-trend & Rank-N-contrast loss & Lithium-ion battery \\

\cite{wang2024Physicsinformedneuralnetwork} & Monotonic degradation, multivariate degradation trend & Regularization term, learnable dynamic model for \acs{PDE} loss & Lithium-ion battery \\

\cite{wang2025PhysicsInformedNeuralNetwork} & Multivariate degradation trend & Learnable dynamic model for \acs{PDE} loss & Lithium-ion battery \\

\cite{pengfeiwen2023PhysicsInformedNeuralNetworks} & Semi-empirical Verhulst degradation model & Nonlinear logistic-type \acs{ODE} & Lithium-ion battery \\

\cite{xu2024PhysicsConstraintVariationalNeural} & Pressure pulsation response model & Algebraic equation & Gear pump \\

\cite{zhang2025AnElectrochemicalAgingInformed} & Electrochemical aging model & Coupled \acsp{ODE}, algebraic equations & Lithium-ion battery \\

\hline
\end{tabular}
\end{table*}

% Lithium-ion batteries
An overview of all studies (18) employing learning bias for health assessment is provided in Table~\ref{tab:learning_bias_health_assessment}. Focusing on \ac{SOH} estimation in lithium-ion batteries, \cite{deng2025ANovelMethod} train an \ac{NN} on features extracted from charge, discharge, and incremental-capacity curves. An additional loss term penalizes deviations from a monotonic relationship between the peak of the incremental-capacity curve and \ac{SOH}, reflecting their consistently one-directional trend over aging. Trained on two public aging datasets with different chemistries and operating conditions, the resulting model achieves lower \ac{SOH} estimation errors than a standard \ac{NN} and a \ac{CNN}.
The challenge of accurately estimating the \ac{SOH} of lithium-ion batteries under dynamic operating conditions is studied by~\cite{wang2024Physicalknowledgeguided}. Building on \ac{ResNet}, the authors use a constraint that guides relative distances and ranking between the embedding space and the output space (\ac{SOH}) as prior knowledge. This battery degradation property was integrated by means of the Rank-N-Contrast loss. Validated on two datasets, the approach demonstrates superior performance in terms of predictive accuracy and structured latent representations compared to traditional and \ac{ML}-based methods. Additional aspects of this work that fall into the class of hybrid approaches are reported in Section~\ref{subsub:health_assessment_hybrid}.

% Lithium-ion batteries
\cite{navidi2024PhysicsInformedMachineLearning} compare four approaches for estimating the capacity (\ac{SOH}) and three internal degradation modes of lithium-ion batteries. As a follow-up study to~\cite{thelen2022Integratingphysicsbasedmodeling}, it includes three approaches that incorporate observational bias, which are described in Section~\ref{subsub:health_assessment_observational_bias}. Furthermore, the authors introduce an approach in which a shallow \ac{NN} predicts half-cell model parameters. A differentiable surrogate of the half-cell model is embedded via additional loss terms that weakly enforces consistency of the predicted parameters and the degradation behavior implied by the half-cell model. Trained on early-life experimental data together with simulation data, the regularized network outperforms an otherwise identical purely data-driven network and all approaches incorporating observational bias.
Explicitly accounting for electrochemical parameter inconsistencies across cells, \cite{zhang2025AnElectrochemicalAgingInformed} present an approach to estimating the \ac{SOH} of lithium-ion batteries. Dedicated subnetworks jointly estimate lithium-ion concentration dynamics and cell-specific electrochemical parameters by processing initial-state features from an early-life discharge to encode parameter variability, and a capacity-difference sequence and sampled time coordinates to capture degradation behavior. The estimated internal states and parameters are then passed through a reduced electrochemical aging model (enhanced \ac{SPM} with polynomial solid-phase diffusion, Butler-Volmer kinetics, Padé-approximated electrolyte diffusion, and electrode stoichiometry shifts) to reconstruct terminal voltage and capacity, with its governing equations and boundary conditions enforced via additional loss terms during training. Across multiple battery chemistries and operating conditions, the method attains very low \ac{SOH}-prediction errors even with scarce training data and supports fast inference, outperforming baselines such as \ac{PINN}, \ac{CNN}, and enhanced \ac{SPM}, though at the cost of higher training complexity.

% Prime example
\begin{figure*}[t]
    \centering
    \includegraphics[width=\textwidth]{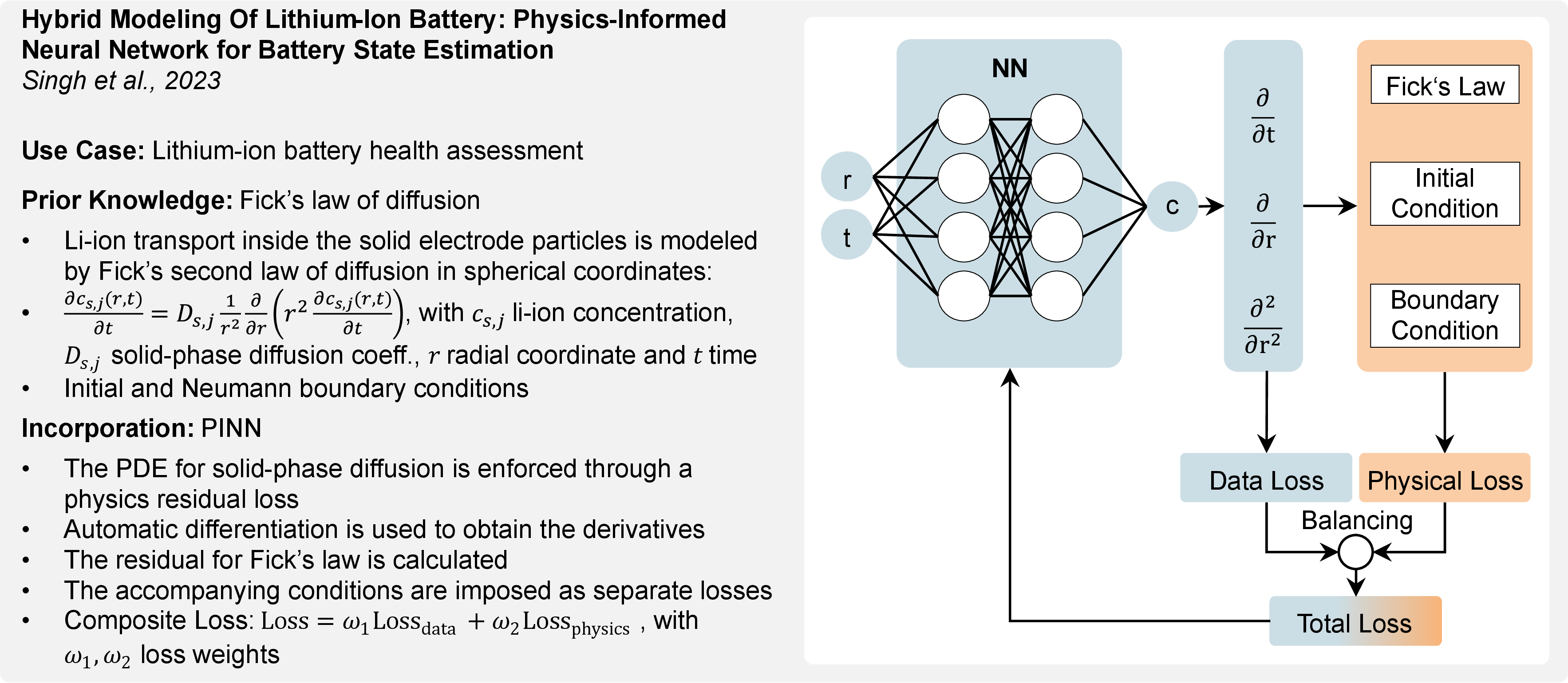}
    \caption{A representative example of learning-bias approaches is proposed by~\cite{singh2023HybridModelingof}, demonstrating how prior physical knowledge can be introduced in the loss function. Own illustration based on the corresponding study. See the original work for full technical details.}
    \label{fig:prime_example_learning_bias}
\end{figure*}

% Lithium-ion batteries - PINN-based approaches
\cite{singh2023HybridModelingof} address the joint estimation of \ac{SOC} and \ac{SOH} for lithium-ion cells operating under varying temperatures and limited measurement data (see Fig.~\ref{fig:prime_example_learning_bias}). Their method incorporates the \ac{PDE} governing solid-phase lithium diffusion from an \ac{SPM}, along with Neumann flux boundary conditions driven by the applied current, into the loss function. To balance the data and physics-informed loss terms, the authors employ gradient normalization for adaptive loss balancing based on the work of \cite{chen2018gradnorm}, although no further details on its implementation are given. The \ac{PINN} predicts spatio-temporal lithium concentration fields within the anode, from which SOC and capacity-based SOH are inferred via established concentration-to-SOC and capacity relations. Results on three cells cycled at different temperatures show that the method accurately tracks both \ac{SOC} and \ac{SOH} over time. Additionally, an ablation study on one cell, where the model is trained on early-life check-up cycles and then applied to subsequent cycles, provides a limited demonstration of short-horizon \ac{SOH} prognosis.
Given that the heat generation rate varies significantly with the \ac{SOH}, a \ac{PINN} is adopted by~\cite{jang2025Stateofhealth}, incorporating a lumped energy-balance equation into its loss function to infer both temperature and the time-varying heat generation rate in a manner consistent with battery thermodynamics. Subsequently, three methods are proposed that leverage the resulting \ac{PINN}-derived profiles to estimate the \ac{SOH}: an \ac{MLP}, a 1D-\ac{CNN}, and an \ac{LSTM}-\ac{CNN}-based approach. However, the evaluation is limited to comparisons among these three methods, leaving their comparative performance relative to established electrical-based or data-driven \ac{SOH} estimation approaches unclear.
\cite{pengfeiwen2023PhysicsInformedNeuralNetworks} also adopt a \ac{PINN}-based approach by embedding a semi-empirical Verhulst degradation model~\citep{xian2013prognosticsverhulstmodel} into the loss function of an \ac{NN} for lithium-ion battery \ac{SOH} estimation. An uncertainty-based weighting scheme adaptively balances the losses during training, allowing the \ac{PINN} to exploit the Verhulst model without manual tuning of loss weights. Experiments demonstrate that this approach achieves lower \ac{SOH} estimation errors compared to both a purely data-driven counterpart and a \ac{GPR}-based method.
Adopting a similar approach, \cite{wang2025ABatteryState} leverage the same Verhulst model~\citep{xian2013prognosticsverhulstmodel} (while additionally integrating the Arrhenius equation) as a soft constraint in the loss function, resulting in a \ac{PINN}-based approach. \ac{SOH} labels are derived from multi-stage constant-current charging via an incremental capacity-based method. Driving-style-related features, together with key operating variables, are used by the \ac{PINN} to estimate the \ac{SOH} of lithium-ion batteries in electric vehicles. Compared with both purely data-driven baselines and alternative \ac{PINN}-based approaches, the proposed model achieves the lowest \ac{SOH} estimation errors.
Addressing \ac{SOH} estimation for lithium-ion batteries via capacity loss, \cite{liu2025Aphysicsguidedapproach} aim to obtain accurate and physically plausible degradation trajectories from charge-discharge data. They first compute entropy-based health indicators from voltage and current, and use these features together with cycle count as inputs to a feedforward \ac{NN}. A mechanistic \ac{SEI}-growth model for capacity fade is then embedded as an additional physics-based loss term that penalizes mismatches between the network’s time derivative of capacity loss and the \ac{SEI} model, thereby regularizing the network toward smooth, mechanistically consistent \ac{SOH} evolution and suppressing non-physical fluctuations. Across three datasets with different chemistries and cycling regimes, this physics-regularized model achieves substantially lower errors than purely data-driven \ac{MLP}- and \ac{CNN}-based models.
Also covering \ac{SOH} estimation for lithium-ion batteries, \cite{wang2024Physicsinformedneuralnetwork} employ a \ac{PINN}-based approach, in which one \ac{NN} estimates the \ac{SOH} based on the current cycle and features generated from sensor readings in the charge phase, and a second \ac{NN} estimates the battery degradation dynamic behavior solely needed for loss construction. From a \ac{PIML} perspective, besides a monotonicity loss, a \ac{PDE}-based loss is also integrated. The latter accounts for the requirement that the degradation trajectory depends on charging rate, discharging rate, temperature, etc., rather than being merely a time-dependent univariate function. In comparison with an \ac{MLP} and a \ac{CNN}, the proposed method performs better in terms of predictive performance, data efficiency and generalizability, as demonstrated in experiments covering \ac{TL} scenarios and data-sparse scenarios. Except for the monotonicity loss, the main aspects of this work are also adopted by~\cite{wang2025PhysicsInformedNeuralNetwork}, where they are specialized for 2-minute data segments obtained from laboratory experiments simulating satellite batteries.

% Bearing
\cite{yonastefera2025ConstraintGuidedLearningof} address the problem of loss balancing in regularization approaches. Considering a convolutional \ac{AE}, the latent variable is fed into multiple fully connected layers for extraction of a \ac{HI} suitable for bearing health assessment. Three constraints are directly integrated into the gradient descent algorithm instead of loss terms. Besides a monotonicity constraint, a boundary constraint leading to a normalized \ac{HI} and a constraint that guides consistency between the signal energy and the \ac{HI} are employed. The approach shows better predictive performance in the majority of experiments compared to other convolutional \ac{AE} baselines.

% Milling (cutting tools)
In this study, \cite{li2025Applicationofphysicsguided} leverage an empirical tool flank wear model relating wear to milling time as prior knowledge for health assessment in high-speed milling. This knowledge is incorporated via a loss function that penalizes deviations between the model predictions and the empirical model during the training of a \ac{GRU}. The regularization enforces physically consistent degradation behavior during training. As a result, the model achieves improved physical consistency while maintaining low prediction errors compared to baselines. See Section~\ref{subsub:health_assessment_inductive_bias} for a description of the inductive bias incorporated by \cite{li2025Applicationofphysicsguided}.

% Axial piston pump
\cite{wangz.2023Physicsinformedneural} incorporate a discharge pressure model for axial piston pumps as a dynamic loss into a \ac{PINN}. Hence, geometrical parameters related to leakage that represent the current health state of the piston-cylinder interfaces can be fitted. Determining these parameters is challenging using traditional methods. As the leakage of the aforementioned intersections can be quantified for multiple piston-cylinder pairs independently, this can also be considered a diagnosis approach.
% Gear pump
Studying gear pumps, \cite{xu2024PhysicsConstraintVariationalNeural} tackle the black-box nature of \ac{DL} models. The output of a compound \ac{NN} (consisting of \ac{CNN}, \ac{BiLSTM} and attention) is guided to match the parameters of a physical pressure model regarding the outlet pressure pulsation. The physics-based model can be used to reconstruct the input signal to perform unsupervised training. The estimated physical parameters are used to construct a health indicator based on distance metrics. The proposed method is superior to purely data-driven approaches in terms of interpretability (the authors show this qualitatively by means of t-distributed stochastic neighbor embedding), though not necessarily in terms of reconstruction capabilities.

% Material - fatigue crack propagation
\cite{pan2025Inservicefatiguecrack} address both the small-sample problem and the poor generalization ability in \ac{ML}-based fatigue crack quantification. Using aluminum specimens representative of aircraft structures, the authors model fatigue crack propagation with an \ac{LSTM} network. A Paris law-based crack-growth model is utilized to formulate an additional loss term aiming to improve the predictive accuracy for crack growth under complex environmental conditions. The positive contribution of the incorporated physics to the model’s overall performance is demonstrated through dedicated ablation studies.
With the incorporated observational bias described in Section~\ref{subsub:health_assessment_observational_bias}, \cite{sun2022MicrocrackDefectQuantification} also introduce a learning bias by augmenting the training objective with additional physics-based loss terms that penalize inconsistencies between the network outputs and analytical relationships derived from guided-wave scattering theory. For crack length, an additional loss term penalizes pairs of samples whose predicted lengths violate the required monotonic relationship between a width-like feature (constructed from neighboring reflection amplitudes) and the angular spread of the reflected lobe. For depth and direction, further penalty terms enforce analytical formulas that link the ratio of reflected to transmitted energy and the corrected reflection angle, respectively, to the corresponding crack parameters given the current length prediction. A single weighting factor controls the influence of all physics-based penalties; sensitivity studies show that choosing this weight appropriately yields substantially lower quantification errors than training without these constraints.

% Turbine (ocean current turbine)
The method of \cite{freeman2022Physicsinformedturbulenceintensity} enables the classification of rotor blade pitch imbalance faults in ocean current turbines according to their level of severity. An \ac{ML} pipeline with the single-phase power output of the turbine as input consists of principal component analysis and multinomial logistic regression for flow-speed classification, an \ac{NN} with an augmented loss for turbulence-intensity classification and a final \ac{NN} for fault severity classification. Both an empirical and an analytical expression of turbulence intensity are used to define a monotonicity constraint as an additional loss. In ablation studies, the proposed method outperforms its purely data-driven counterpart in terms of predictive performance.

\subsubsection{Prognosis}
\label{subsub:prognosis_learning_bias}

\begin{table*}[t]
\centering
\caption{All studies employing \textit{learning bias} to address \textit{prognosis}, listed in alphabetical order.}
\label{tab:learning_bias_prognosis}
\begin{tabular}{@{}
  >{\RaggedRight\arraybackslash}p{0.24\textwidth}%
  >{\RaggedRight\arraybackslash}p{0.31\textwidth}%
  >{\RaggedRight\arraybackslash}p{0.23\textwidth}%
  >{\RaggedRight\arraybackslash}p{0.18\textwidth}%
  @{}}
\multicolumn{4}{@{}l@{}}{\bfseries Learning Bias for Prognosis} \\
\hline
\textbf{Reference} 
& \multicolumn{2}{l}{\textbf{Prior Physical Knowledge}} 
& \textbf{Use Case} \\
& Type 
& Representation
& \\
\hline

\cite{badora2023Usingphysicsinformedneural} & Fracture-mechanics relation between load ratio of thermal stresses and corresponding stress intensity factors & Algebraic ratio constraint & High-pressure nozzle of an industrial gas turbine \\

\cite{e2025Aphysicsinformedneural} & Empirical aging law relating \acs{SOH} to cycle number & Algebraic equation & Supercapacitors \\

\cite{fassi2024PhysicsInformedMachineLearning} & Empirical aging trend & Monotonicity and boundedness constraints (loss terms) & Metal-oxide-semiconductor field-effect transistor \\

\cite{he2025Physicsinformedneuralnetwork} & Impedance-based degradation mechanism, stochastic degradation behavior (Wiener process) & Intermediate physical variables, Gaussian increment model & Lithium-ion battery \\

\cite{najeraflores2023APhysicsConstrainedBayesian} & Empirical aging trend & Monotonicity constraint (loss term) & Lithium-ion battery \\

\cite{pugalenthi2024RemainingUsefulLife} & Semi-empirical \acs{SEI}-film-based capacity fade model & Two-term exponential equation & Lithium-ion battery \\

\cite{ramirez2024ResidualbasedAttentionPhysicsinformed} & Heat-diffusion model of transformer oil, standard thermal-aging model for winding insulation & 1D heat-diffusion \acs{PDE}, algebraic equations & Transformer \\

\cite{wang2024Phyformer:Adegradation} & Monotonicity assumption & Single-term exponential function & Bearing, transformer, electromechanical servo system \\

\cite{wang2025KoopmanInformedNeuralNetwork} & Koopman operator theory & Linear operator with algebraic consistency loss & Turbofan engine, bearing \\

\cite{wang2025Aremaininguseful} & Reliability model of bearing failure process & Weibull cumulative distribution function & Bearing \\

\cite{xu2022Aphysicsinformeddynamic} & First-order Thevenin model & \acsp{ODE} & Lithium-ion battery \\

\cite{zhang2025APhysicsInformedHybrid} & Enhanced \acs{SPM} & \acs{ODE} system with algebraic voltage relation & Lithium-ion battery \\

\cite{zhu2024RemainingUsefulLife} & Inverse monotonic relationship between crack surface area and \acs{RUL} & Monotonicity constraint (loss term) & Bearing \\

\hline
\end{tabular}
\end{table*}

% Lithium-ion batteries (SOH prognosis)
Thirteen studies employ learning bias for prognosis (see Tab.~\ref{tab:learning_bias_prognosis}), several of which focus on lithium-ion battery prognostics. \ac{SOH} prognosis of lithium-ion batteries is addressed by~\cite{xu2022Aphysicsinformeddynamic}. A \ac{ResNet}-based encoder-decoder model uses capacity and secondary variables (i.e., temperature, voltage, and current) from the current cycle to predict capacity and full secondary-variable profiles for the subsequent cycle. This one-step-ahead predictor is iterated from the first discharge cycle to generate long-horizon degradation trajectories over the battery’s life. Physics is incorporated through additional loss terms that penalize violations of a first-order Thevenin-based state equation and a capacity-balance equation, which are combined into an \ac{ODE} system. The proposed method achieves substantially lower \ac{SOH} prediction errors than several \ac{GP}-based baselines using only the first discharge cycle, with ablation studies attributing the performance gains to the added regularization based on the underlying battery model.
\cite{pugalenthi2024RemainingUsefulLife} present a prognostic framework for lithium-ion batteries, where an \ac{NN} is first identified from a single run-to-failure cell using a \ac{PF} and then used to predict the capacity trajectories of other cells. A semi-empirical \ac{SEI}-film-based capacity fade model is incorporated as an additional loss term to improve physical plausibility and reduce errors in predicted capacity, especially when only limited training data are available. The trade-off is higher computational cost due to the extra overhead of evaluating the physics-based loss on top of the \ac{PF}-based parameter estimation.
\cite{he2025Physicsinformedneuralnetwork} propose a degradation modeling framework that combines \ac{DL} with a Wiener process for lithium-ion batteries. The network takes degradation features (i.e., constant-current charging time, incremental-capacity peak, and temperature peak) and maps them through intermediate physical variables that represent impedance-related quantities, before producing a nonlinear degradation path that serves as the drift of the Wiener process. A composite loss combining mean squared error on these intermediate physical variables with a likelihood-based term for degradation increments is used to train both the network and stochastic-process parameters jointly from historical data. The latter are further updated online via Bayesian inference using real-time \ac{SOH} measurements. Ablation and comparative studies on two battery datasets show that adding the impedance-informed latent layer and the Wiener process module improves not only \ac{SOH} trajectory prediction but also the fidelity of the resulting reliability curves and \ac{RUL} and lifetime distributions, compared with purely data-driven and purely stochastic-process baselines.
\pagebreak
% Lithium-ion batteries (RUL prognosis)
To effectively target \ac{RUL} prediction of lithium-ion batteries, \cite{zhang2025APhysicsInformedHybrid} propose a two-stage approach. In the first stage, an electrochemical-informed generative model, constrained by a reduced-order enhanced \ac{SPM}, reconstructs electrode-level states. The model is trained with a composite loss over the electrochemical governing equations, initial and boundary conditions, and terminal-voltage mismatch, whose weights are adaptively balanced via gradient-norm ratios. In the second stage, incremental-ca\-pac\-i\-ty and differential-voltage curves derived from the reconstructed states serve as electrode-lev\-el features. These are combined with cell-level features (capacity-dif\-ference curves and charging protocols) in an \ac{NN}: two \ac{CNN} branches encode the cell-level inputs, while a \ac{GRU} with self-attention encodes the electrode-level inputs, and the concatenated representations are decoded to predict the \ac{RUL}. Across four datasets with different chemistries and operating conditions, the method outperforms both mechanistic and purely data-driven baselines in terms of prediction error, shows better robustness with limited training data, and offers faster inference, while also enabling identification of electrode-level degradation modes.
Early-stage \ac{RUL} prediction is studied by~\cite{najeraflores2023APhysicsConstrainedBayesian}. To this end, a neural differential operator is learned for the discharge capacity rate from early-life cycling data. The proposed architecture draws inspiration from DeepONet~\citep{lu2021learning}, featuring a Bayesian branch network that encodes cell-specific early-life features and a deterministic trunk network that encodes time. Multiple loss terms are employed to promote accurate modeling of the discharge capacity, including a monotonicity constraint that weakly enforces negative self-acceleration of the capacity trajectory. At inference, the learned operator is sampled from the Bayesian branch, integrated forward in time to reconstruct the capacity curve, and the \ac{RUL} is obtained as the difference between the predicted end-of-life time (at a capacity threshold) and the current cycle. Experiments show that, given the same early-life training data, the proposed physics-constrained Bayesian operator achieves smaller \ac{RUL} prediction errors and more reliable uncertainty quantification than a simplified physics-based failure forecast model, a \ac{GPR}-based method, and a similarity-based method.

% Bearing
\cite{wang2025Aremaininguseful} propose an approach for bearing \ac{RUL} prediction using acoustic emission signals. A novel health indicator is introduced to quantify acoustic emission signal complexity and is shown to outperform standard time-domain and entropy features in monotonicity, robustness, and trendability. This health indicator is fed to an \ac{LSTM} whose loss combines mean squared error with a Weibull-based term derived from reliability engineering, thereby constraining the learned degradation trajectory to be consistent with the expected failure behavior. Experiments demonstrate that the proposed method yields substantially lower \ac{RUL} prediction errors than a conventional \ac{LSTM}.
Already discussed in Section~\ref{subsub:prognosis_observational_bias}, \cite{zhu2024RemainingUsefulLife} propose three mechanisms for incorporating physics to tackle bearing \ac{RUL} prediction. In terms of learning bias, the authors design an inconsistency loss that penalizes pairs of predicted \ac{RUL} values violating the inverse monotonic relation between crack surface area and \ac{RUL}---imposed via a \ac{ReLU}-based penalty. Again, the complete framework proves superior, but without ablation experiments, the specific contribution of the additional loss term cannot be quantified. See Section~\ref{subsub:prognosis_hybrid} for a synopsis focusing on the hybrid aspect of the framework.

% Bearing, turbofan engine
To facilitate prognostics of complex industrial machinery, \cite{wang2025KoopmanInformedNeuralNetwork} present a novel approach grounded in Koopman operator theory. The proposed Koopman-informed \ac{NN} enables accurate \ac{RUL} prediction by learning nonlinear system dynamics through a linear representation in a latent eigenfunction space. Within an encoder-decoder architecture, the encoder maps data spanning the entire operational lifespan into an eigenfunction space, where the system’s evolution is approximated by a finite-dimensional Koopman matrix: multi-step temporal forecasting is performed via repeated applications of the learned forward and backward Koopman operators. The decoder reconstructs future system states from the propagated latent representation. Multiple loss terms (including a consistency loss that penalizes discrepancies between forward and backward evolution) regularize the learned dynamics and promote stable, physically coherent temporal behavior. Lastly, a nonlinear regression head leverages the learned high-level features to produce precise \ac{RUL} estimates. The proposed Koopman-informed \ac{NN} is evaluated on both a bearing and a turbofan engine dataset, where the model consistently outperforms several strong \ac{DL} baselines in terms of predictive performance.

% Bearings, transformer, electromechanical servo system
\cite{wang2024Phyformer:Adegradation} propose a Transformer-based prognostics model designed for scenarios lacking reliable degradation physics. The approach hinges on decomposing time-series data into a slowly varying trend component (capturing the overall degradation trend) and a residual component (capturing higher-frequency fluctuations). Thereafter, a simple monotonic parametric curve is fitted to the trend component in a sliding-window procedure. Rather than constructing accurate, domain-specific degradation models, these local fits (here, single-term exponentials) are treated as \enquote{simple, general and imperfect} approximations. An additional loss term penalizes large deviations from these fitted curves, thereby regularizing the model toward predictions that reflect the assumed irreversibility of degradation. Given its domain-agnostic prior knowledge, the model is applied to three different use cases (bearings, transformers, and an electromechanical servo system), where it consistently reduces long-horizon prediction errors compared to several state-of-the-art \ac{DL} models.

% MOSFET
Given that failure of power semiconductor devices poses a significant reliability challenge for power converter systems, \cite{fassi2024PhysicsInformedMachineLearning} address \ac{RUL} prediction for power metal-oxide-semiconductor field-effect transistors under thermal aging. To improve predictive accuracy and physical consistency, the authors incorporate additional loss terms that weakly enforce monotonic, bounded \ac{RUL} trajectories via \ac{ReLU}-based penalties. A series of experiments across various recurrent network architectures demonstrates that it achieves lower or comparable mean squared error than purely data-driven counterparts, while simultaneously enabling faster convergence.

% Supercapacitor
\cite{e2025Aphysicsinformedneural} present a method for predicting the \ac{RUL}  of commercial supercapacitors by embedding an empirical aging law that models \ac{SOH} as a logarithmic function of cycle number into the loss function of an \ac{LSTM}. A scalar weighting factor between the data and physics losses is tuned via Bayesian optimization, leading to stronger regularization under scarce data and reduced reliance on the physical loss term as more data become available. Experiments demonstrate that the proposed model substantially outperforms its purely data-driven counterpart, with prediction errors comparable to more advanced data-driven methods while using only a fraction of the full life cycle as training data.

% High-pressure nozzle of an industrial gas turbine
In addition to the inductive bias reported in Section~\ref{subsub:prognosis_inductive_bias}, \cite{badora2023Usingphysicsinformedneural} incorporate a learning bias independent of the former. An additional loss term regularizes training with respect to the fracture-mechanics load-ratio relation between applied thermal stresses and the corresponding stress intensity factors, using a large set of synthetically generated stress-crack-length combinations. A dynamically adjusted weighting factor gradually shifts emphasis from this physics term toward the empirical crack-length error as training progresses, ensuring that the final model both respects the underlying fracture mechanics and fits the sparse inspection data. While the need for a dynamic weighting between physics and empirical loss terms is well motivated, the specific piecewise schedule for the weighting coefficient is introduced without justification, making this part of the approach heuristic and potentially hard to generalize or reproduce. Combined with the inductive encoding of Paris' law, the proposed method effectively targets fatigue crack growth modeling in a gas turbine nozzle.

% Distribution transformer in a floating photovoltaic power plant
\cite{ramirez2024ResidualbasedAttentionPhysicsinformed} propose an efficient spatio-temporal model for predicting transformer winding temperature, including the local hotspot, and insulation aging. Embedding a simplified one-dimensional heat-diffusion \ac{PDE} with uniform heating into a \ac{PINN} enables estimating oil temperatures, which are then used to calculate winding temperatures, aging acceleration factors, and the associated loss of life. Tested on a distribution transformer in a floating photovoltaic power plant, the method closely matches numerical \ac{PDE} solutions and improves hotspot and aging estimation compared to a standard analytic hotspot model, with validation against fiber optic sensor measurements. Additionally, a residual-based attention scheme improves convergence and training stability of the \ac{PINN}.

% - - - - - - - - - - subsection - - - - - - - - - - %
\subsection{Hybrid Approaches}
\label{subsec:hybrid_approaches}

\begin{figure*}[t]
    \centering
    \includegraphics[width=\textwidth]{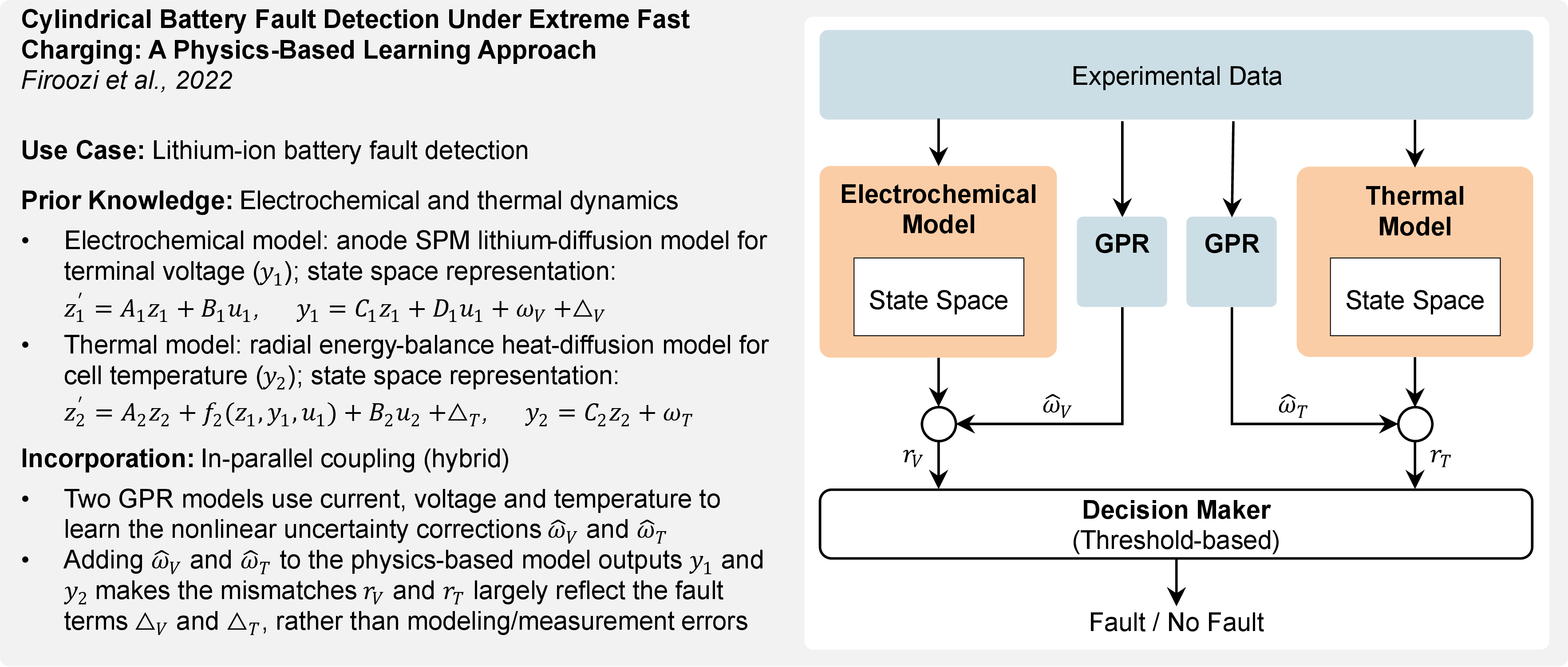}
    \caption{A representative example for hybrid approaches is proposed by~\cite{firoozi2022CylindricalBatteryFault}, demonstrating how physics-based and \ac{ML} models can be combined. Own illustration based on the corresponding study. For brevity, time indices are omitted, with the prime denoting the next state. See the original work for full technical details.}
    \label{fig:prime_example_hybrid_approaches}
\end{figure*}

% Outlook for the section to follow
Hybrid approaches are characterized by combining independent physics-based and data-driven models, either in parallel or in series.  The identified studies are organized 
by \ac{PHM} task in 
Tables~\ref{tab:hybrid_fault_detection}--\ref{tab:hybrid_prognosis}. A representative example is illustrated in Figure~\ref{fig:prime_example_hybrid_approaches}.

% prime example

\subsubsection{Fault Detection}
\label{subsub:fault_detection_hybrid}

\begin{table*}[ht]
\centering
\caption{All studies employing \textit{hybrid approaches} to address \textit{fault detection}, listed in alphabetical order.}
\label{tab:hybrid_fault_detection}
\begin{tabular}{@{}
  >{\RaggedRight\arraybackslash}p{0.24\textwidth}%
  >{\RaggedRight\arraybackslash}p{0.31\textwidth}%
  >{\RaggedRight\arraybackslash}p{0.23\textwidth}%
  >{\RaggedRight\arraybackslash}p{0.18\textwidth}%
  @{}}
\multicolumn{4}{@{}l@{}}{\bfseries Hybrid Approaches for Fault Detection} \\
\hline
\textbf{Reference} 
& \multicolumn{2}{l}{\textbf{Prior Physical Knowledge}} 
& \textbf{Use Case} \\
& Type 
& Representation
& \\
\hline

\cite{firoozi2022CylindricalBatteryFault} & \acs{SPM} model and radial thermal model & Reduced-order \acs{ODE} state-space model & Lithium-ion battery \\

\cite{zhang2024Adaptivefaultdetection} & Second-order \acs{ECM} coupled with a lumped thermal model & \acs{ODE} state-space model, lumped \acs{ODE} & Lithium-ion battery \\

\hline
\end{tabular}
\end{table*}

% Lithium-ion battery
Two studies employing hybrid approaches for fault detection have been identified (see Tab.~\ref{tab:hybrid_fault_detection}), both targeting lithium-ion batteries. \cite{firoozi2022CylindricalBatteryFault} target real-time fault detection in cylindrical lithium-ion batteries under extreme fast charging, aiming for early detection of voltage and thermal faults (see Fig.~\ref{fig:prime_example_hybrid_approaches}). Two detection observers are built on an experimentally identified reduced-order electrochemical-thermal model; a \ac{GPR} is used in parallel to learn additive voltage and temperature uncertainty terms from residuals between model predictions and measurements in no-fault cycles. The learned uncertainty corrections are fed back into the observers to suppress non-fault deviations, and residuals are evaluated against calibrated thresholds to detect voltage and thermal faults. Comparative results with a model-only observer indicate that incorporating the \ac{GPR} can enable detection of smaller faults that the purely physics-based counterpart misses.
Building on a similar concept, \cite{zhang2024Adaptivefaultdetection} (who also reference~\cite{firoozi2022CylindricalBatteryFault} as related work) propose an adaptive fault detection framework for lithium-ion batteries that combines a thermoelectric model-based \ac{EKF} observer with a \ac{BiLSTM}. A second-order \ac{ECM} coupled with a simplified thermal model is used in the \ac{EKF} to estimate voltage, temperature, and \ac{SOC}, while the \ac{BiLSTM} is trained on healthy data to learn the voltage observation error and subsequently compensates the observer to suppress uncertainty-induced residuals. The corrected residuals are compared against a threshold calibrated from healthy operation to detect soft internal short circuit faults. Experiments on driving-cycle data show that the proposed framework yields substantially lower voltage estimation errors than the standalone \ac{EKF}, thereby enabling more reliable fault detection with fewer false alarms.

\subsubsection{Diagnosis}
\label{subsub:diagnosis_hybrid}

\begin{table*}[t]
\centering
\caption{All studies employing \textit{hybrid approaches} to address \textit{diagnosis}, listed in alphabetical order.}
\label{tab:hybrid_diagnosis}
\begin{tabular}{@{}
  >{\RaggedRight\arraybackslash}p{0.24\textwidth}%
  >{\RaggedRight\arraybackslash}p{0.31\textwidth}%
  >{\RaggedRight\arraybackslash}p{0.23\textwidth}%
  >{\RaggedRight\arraybackslash}p{0.18\textwidth}%
  @{}}
\multicolumn{4}{@{}l@{}}{\bfseries Hybrid Approaches for Diagnosis} \\
\hline
\textbf{Reference} 
& \multicolumn{2}{l}{\textbf{Prior Physical Knowledge}} 
& \textbf{Use Case} \\
& Type 
& Representation
& \\
\hline

\cite{pettorossi2025Physicsguidedfaultdiagnosis} & Proton exchange membrane fuel cell simulation & \acs{ODE} system & Fuel cell \\

\cite{singh2024Hybridphysicsinfused1DCNN} & 0D high-fidelity physics-based engine model & Coupled algebraic equations, \acsp{ODE} & Diesel engine \\

\cite{xu2023Physicsguideddatarefinedfault} & Fault hierarchies and statistical fault evolution/propagation mechanisms (time-to-failure behavior and correlations between failure modes) & Probability density functions and cumulative distributions of failure modes, algebraic update formulas & Offshore wind turbine \\

\hline
\end{tabular}
\end{table*}

% Fuel cell (proton exchange membrane fuel cell)
Table~\ref{tab:hybrid_diagnosis} lists the three studies that employ hybrid approaches for diagnosis. \cite{pettorossi2025Physicsguidedfaultdiagnosis} address fault diagnosis for proton exchange membrane fuel cells, focusing on identifying and isolating four different fault types: flooding, drying, air starvation, and hydrogen starvation. To tackle this, the authors propose an approach that combines a physics-based proton exchange membrane fuel cell model with an \ac{LSTM}. The physics-based model provides estimates of unmeasured process variables, including membrane resistance, water content, and current density distribution. These variables are combined with measured stack signals and fed to the \ac{LSTM}, both in training and inference. Hence, more informed and robust fault classification is enabled. This integration improves diagnostic accuracy, reduces detection time, and enhances generalizability compared to purely data-driven approaches.

% Diesel engine
\cite{singh2024Hybridphysicsinfused1DCNN} propose a  fault diagnosis framework for a diesel engine that combines a 0D (lumped, time-dependent) high-fidelity physics-based engine model with an \ac{AE} and a 1D-\ac{CNN}. The \ac{AE} compresses high-dimensional sensor data into a latent feature vector, while the physics-based engine model provides additional simulated variables. These are then concatenated and passed to a 1D-\ac{CNN} that classifies four conditions: nominal operation, and faults due to injection pressure, injection duration, and start of injection. The engine model thus acts as an in-parallel source of physics-based features that complement the data-driven latent representation for fault classification. Experiments on test-bed data show that the proposed model achieves higher diagnostic accuracy than a purely data-driven 1D-\ac{CNN} and exhibits improved robustness to sensor noise and to extrapolation across unseen engine speeds and operating conditions, with notably fewer false positives in nominal conditions.

% offshore wind turbine
Fault root cause tracing in complex electromechanical systems is studied by~\cite{xu2023Physicsguideddatarefinedfault}, demonstrated on an offshore wind turbine experiencing an unscheduled power drop. Common physics- and statistics-based knowledge about fault mechanisms is first encoded into a static hierarchical fault root cause tracing network---a probabilistic graph whose nodes and edges represent functional units, fault modes, and their propagation relationships. This model is then refined with operation data: anomalies detected via a Wasserstein \ac{GAN}, together with statistical laws describing how faults evolve over operating time, update the weights of fault nodes and edges. Finally, a bidirectional probabilistic reasoning scheme combines forward fault propagation and backward tracing information across the hierarchy to rank fault nodes and identify the most likely root cause and fault paths. 

\subsubsection{Health Assessment}
\label{subsub:health_assessment_hybrid}

% Lithium-ion batteries
Table~\ref{tab:hybrid_health_assessment} summarizes the five studies that employ hybrid approaches for health assessment, all but one of which target lithium-ion batteries. Addressing \ac{SOH} estimation, \cite{feng2024comprehensive} leverage a second-order resistor-capacitor \ac{ECM}, whose parameters are identified from post-charge relaxation voltage via nonlinear least squares and then used as inputs to various \ac{ML} models, including \ac{GPR}, XGBoost, \ac{SVR}, elastic net, and a simple \ac{MLP}. For each regressor, the hybrid approach is compared with its purely data-driven counterpart trained either on raw relaxation voltage samples or on statistical features extracted from the relaxation curve. The results indicate that the hybrid \ac{GPR} generally achieves the lowest prediction error, with its advantage particularly evident when training data or relaxation time are limited.
Essentially following the same approach, \cite{lin2025Physicsinformedmachinelearning} use a fractional-order \ac{ECM}, whose parameters are identified via recursive least squares and then passed to an \ac{RF} regressor for \ac{SOH} estimation. Although not investigating further \ac{ML} models, they also benchmark the hybrid \ac{RF} against a standard \ac{RF} (either trained on raw relaxation data or on statistical features), with the former generally outperforming its counterparts in terms of prediction error.
\cite{kohtz2022PhysicsbasedMachineLearning} present a hybrid approach for the online joint \ac{SOC} and \ac{SOH} estimation of lithium-ion batteries. A dual \ac{EKF} is built on a simple empirical voltage measurement function (polynomial open-circuit voltage-type relation in \ac{SOC}, capacity, and current), and an \ac{NN} is trained offline as a residual model to learn the error between this physics-based measurement function and the measured voltage. In online operation, the \ac{NN} correction is added to the empirical measurement function within the dual \ac{EKF} measurement equation. Results show that embedding the residual model significantly reduces capacity estimation error, demonstrating more accurate battery health assessment.
In addition to the learning bias discussed in Section~\ref{subsub:health_assessment_learning_bias}, the method proposed by~\cite{wang2024Physicalknowledgeguided} also falls into the class of in-series hybrid approaches. An \ac{ECM}'s identified parameters are concatenated with data-driven features extracted from current measurements, both in training and inference. The concatenated feature vector forms the input for a subsequent encoder.

\begin{table*}[t]
\centering
\caption{All studies employing \textit{hybrid approaches} to address \textit{health assessment}, listed in alphabetical order.}
\label{tab:hybrid_health_assessment}
\begin{tabular}{@{}
  >{\RaggedRight\arraybackslash}p{0.24\textwidth}%
  >{\RaggedRight\arraybackslash}p{0.31\textwidth}%
  >{\RaggedRight\arraybackslash}p{0.23\textwidth}%
  >{\RaggedRight\arraybackslash}p{0.18\textwidth}%
  @{}}
\multicolumn{4}{@{}l@{}}{\bfseries Hybrid Approaches for Health Assessment} \\
\hline
\textbf{Reference} 
& \multicolumn{2}{l}{\textbf{Prior Physical Knowledge}} 
& \textbf{Use Case} \\
& Type 
& Representation
& \\
\hline

\cite{feng2024comprehensive} & Second-order RC \acs{ECM} & Nonlinear algebraic expressions of exponentials (derived from first-order linear \acsp{ODE}) & Lithium-ion battery \\

\cite{kohtz2022PhysicsbasedMachineLearning} & Empirical voltage-\acs{SOC}-capacity relation (open-circuit voltage-type measurement model) & Algebraic equation & Lithium-ion battery \\

\cite{lin2025Physicsinformedmachinelearning} & Fractional-order \acs{ECM} & Fractional-order state-space model & Lithium-ion battery \\

\cite{wang2024Physicalknowledgeguided} & Second-order RC \acs{ECM} & \acs{ODE} system & Lithium-ion battery \\

\cite{xu2024Wearstateassessment} & System-level lumped-parameter dynamic model & Coupled nonlinear \acsp{ODE} & Gear pump \\

\hline
\end{tabular}
\end{table*}

% Gear pump
\cite{xu2024Wearstateassessment} present a digital twin-based framework for wear state assessment of gear pumps in fuel control systems. A first-principles dynamic model of the fuel system is built in Simulink, while a deep \ac{RL} agent adaptively updates flow correction coefficients so that simulated pressures match measured ones. These learned coefficients, together with normalized operating conditions and model error, form an interpretable wear feature vector whose distance to a healthy reference indicates wear severity.

\subsubsection{Prognosis}
\label{subsub:prognosis_hybrid}

\begin{table*}[t]
\centering
\caption{All studies employing \textit{hybrid approaches} to address \textit{prognosis}, listed in alphabetical order.}
\label{tab:hybrid_prognosis}
\begin{tabular}{@{}
  >{\RaggedRight\arraybackslash}p{0.24\textwidth}%
  >{\RaggedRight\arraybackslash}p{0.31\textwidth}%
  >{\RaggedRight\arraybackslash}p{0.23\textwidth}%
  >{\RaggedRight\arraybackslash}p{0.18\textwidth}%
  @{}}
\multicolumn{4}{@{}l@{}}{\bfseries Hybrid Approaches for Prognosis} \\
\hline
\textbf{Reference} 
& \multicolumn{2}{l}{\textbf{Prior Physical Knowledge}} 
& \textbf{Use Case} \\
& Type 
& Representation
& \\
\hline

\cite{aizpurua2023Integratedmachinelearning} & Arrhenius-based thermal-stress model with Miner's rule for stator winding insulation & Algebraic equation & Electric motor \\

\cite{kundu2024Developmentofdatadriven} & Pit-growth model inspired by Paris' law & Algebraic equation & Gearbox \\

\cite{li2024Particlefilterbasedfatiguedamageprognosisusingprognosticaidedmodelupdating} & Fatigue crack growth model (Paris' law) & Algebraic equation & Fatigue crack growth in aluminum lug joint \\

\cite{liang2024Ahybridapproach} & Double exponential model for capacity prediction & Algebraic equation & Lithium-ion battery \\

\cite{ma2024Accurateandefficient} & Incremental capacity curves expressed as sum of Lorentzian functions & Algebraic equation & Lithium-ion battery \\

\cite{shi2022Batteryhealthmanagement} & Semi-empirical calendar and cyclic aging model & Algebraic equation & Lithium-ion battery \\

\cite{sun2023Adaptiveevolutionenhanced} & Electrochemical-thermal-\acs{SEI} model & Coupled \acsp{PDE} plus \acs{SEI} capacity-fade \acs{ODE}, solved as a high-fidelity numerical simulation & Lithium-ion battery \\

\cite{xu2023ANovelHybrid} & Pseudo-two-dimensional electrochemical model (Doyle-Fuller-Newman model) & \acs{PDE} & Lithium-ion battery \\

\cite{zhu2024PhysicsInformedDeepLearning} & Flank wear model & Algebraic equation & High-speed milling \\

\cite{zhu2024RemainingUsefulLife} & Vibration degradation model & Analytic model defined by algebraic equations & Bearing \\

\hline
\end{tabular}
\end{table*}

% Lithium-ion batteries
A total of ten studies employ hybrid approaches for prognosis, as listed in Table~\ref{tab:hybrid_prognosis}. \cite{sun2023Adaptiveevolutionenhanced} propose a prognostics framework for lithium-ion batteries, in which a multi-physics simulation model (electrochemical-thermal with \ac{SEI}-coupling) and an \ac{LSTM} operate in series. The simulation model uses measured current, voltage and temperature to estimate \ac{SOH}, which is then used both as training targets and as a slower, high-fidelity reference to periodically recalibrate the \ac{LSTM} during operation. The \ac{LSTM} employs a dynamically sized input window whose length is adapted based on the Kullback-Leibler divergence between consecutive windows, so that the network alternately emphasizes long-term trends and short-term fluctuations. An adaptive evolution mechanism retrains and updates the \ac{LSTM} whenever its predictions deviate too much from the simulation, improving long-term prediction performance. Compared to a conventional \ac{LSTM}, the proposed framework consistently achieves lower \ac{RUL} prediction errors across two datasets and laboratory experiments conducted under varying operating conditions.
Specifically tailored for data-scarce scenarios, \cite{liang2024Ahybridapproach} propose a hybrid method for \ac{RUL} prediction and uncertainty quantification in lithium-ion batteries. An ensemble learning approach is adopted to integrate an empirical degradation model and a data-driven component. While the former, a double exponential degradation model, captures the nonlinear degradation trend, the \ac{GRU}-\ac{CNN} network learns to predict short-term fluctuations. The output of these models, along with the preprocessed data, is fed into a Bayesian \ac{NN} that predicts \ac{RUL} and provides uncertainty quantification, showing comparable performance to several alternative approaches and increased data efficiency in ablation studies.
\cite{ma2024Accurateandefficient} present an approach for predicting battery \ac{RUL} from a single constant-current charging curve. The method also addresses the black-box nature of data-driven methods by combining battery physics and \ac{ML}. Characteristic peaks from incremental capacity curves are approximated with Lorentzian functions and integrated to form a smooth, parametric capacity-voltage model. From this model, the peak centers, widths, and areas are extracted as features sensitive to degradation for a lightweight \ac{NN} that maps them to the \ac{RUL}. Across different chemistries and operating conditions, the proposed approach provides more stable and accurate predictions than purely data-driven methods.
The framework of~\cite{xu2023ANovelHybrid} enables predicting lithium-ion battery degradation trajectories and \ac{RUL}. Early-life features are extracted in three forms: the variance of the difference of discharge capacity-voltage curves between cycles 10 and 100; the anode state-of-lithiation change between cycles 90 and 100 obtained from a pseudo-2-dimensional electrochemical model; and a final feature that multiplies the aforementioned to capture both observable and internal degradation signals. Battery cells are clustered using k-means clustering based on the third feature mentioned to group similar aging patterns. Data augmentation techniques are applied to enrich cluster-specific datasets. For each cluster, an \ac{LSTM} encoder-decoder model predicts the full capacity degradation trajectory from limited early cycles. The proposed method enables accurate and early prediction of battery life across different aging conditions and chemistries and performs better than \ac{GPR}, \ac{SVR} and an autoregressive \ac{LSTM} baseline.
\cite{shi2022Batteryhealthmanagement} present a physics-informed framework for lithium-ion battery degradation estimation and \ac{RUL} prediction. Their method combines a calendar-and-cycle-aging model, expressed through five semi-empirical operating stress-factor formulations, with an \ac{LSTM}. The physics-based component represents degradation associated with operating conditions (including cycle duration, rest time, temperature, \ac{SOC}, and load), while the \ac{LSTM} uses this modeled degradation together with cycle-level monitoring signals to learn degradation behavior that is not captured by the former alone. The estimated degradation trajectory is then passed to a separate \ac{LSTM} that forecasts future capacity loss; \ac{RUL} is obtained from the predicted point at which the end-of-life threshold is reached. In the reported experiments, the method achieves better capacity-fade modeling performance than the \ac{CNN} and \ac{BiLSTM} baselines under the tested conditions.

% Bearings
Introducing three distinct mechanisms to inform a \ac{BiLSTM}, \cite{zhu2024RemainingUsefulLife} propose a framework for bearing \ac{RUL} prediction. One of these mechanisms represents an in-series approach: a parametric physics degradation model generates an explicit degradation trajectory, which is subsequently passed to the recurrent model to guide training by a mechanistic representation of health progression. Combined with the observational and learning biases (see Sec.~\ref{subsub:prognosis_observational_bias} and Sec.~\ref{subsub:prognosis_learning_bias}, respectively), the framework consistently outperforms purely data-driven baselines in terms of \ac{RUL} prediction errors. Yet the lack of ablation studies prevents determining which of the three mechanisms yields the greatest benefit.

% Milling (cutting tools)
\cite{zhu2024PhysicsInformedDeepLearning} target tool wear monitoring and \ac{RUL} prediction in high-speed milling. A \ac{BiLSTM}-based architecture processes cutting force signals to extract temporal features, while a physics-based wear model (driven by the milling parameters) runs in parallel and provides a flank-wear estimate that conditions an attention mechanism aggregating the learned features into a shared representation used for prediction. Assuming that tool wear and \ac{RUL} are two manifestations of the same underlying tool state, the network jointly predicts both quantities from this shared representation under a single multi-task loss with two separate regression heads. Experiments over different milling conditions show that the proposed model yields clearly improved  predictive accuracy compared with an otherwise identical purely data-driven model.

% Fatigue - materials 
To address fatigue crack prognosis in an aluminum lug joint monitored by Lamb waves, \cite{li2024Particlefilterbasedfatiguedamageprognosisusingprognosticaidedmodelupdating} targets \ac{RUL} prediction within a hybrid, \ac{PF}-based framework. Based on Paris' law, a nonlinear state-space model is constructed to represent the evolution of fatigue crack length. Additionally, a \ac{GPR} model maps the Lamb-wave feature to lifetime percentage, which is subsequently used to modify \ac{PF} state and parameter samples (prognostic-aided model updating) so that they are consistent with both past measurements and prognostic information. Applied to five specimens across five testing scenarios, the method yields more accurate \ac{RUL} and lifetime-percentage predictions with generally tighter uncertainty bounds, and achieves more reliable prognosis than both the approach without prognostic-aided updating and the standalone \ac{GPR} model.

% Gearbox
In order to perform \ac{RUL} prediction for gearboxes, \cite{kundu2024Developmentofdatadriven} propose multiple approaches. In one approach, \ac{RFR} is used to determine the current pitting area based on a correlation coefficient-based \ac{HI} constructed from both healthy and faulty gearbox vibration data. With the current pitting area known, a physical pit-growth model can be employed to estimate the number of remaining cycles until failure that directly relates to the \ac{RUL}. Additionally, Bayesian inference techniques are used to update the parameters of the physical model during inference.

% Electrical component - electric motor
\cite{aizpurua2023Integratedmachinelearning} address prognostics of permanent magnet motors in a maritime context. With the winding insulation as the degradation quantity of interest, features derived from wind speed and vessel speed are used as inputs to two \ac{ML} models connected in series. The first model predicts torque from operational and meteorological data, and the second predicts winding temperature using the predicted torque along with the same input features. The resulting winding-temperature trajectory is then fed into an Arrhenius-based thermal-stress degradation model with Miner’s rule and Monte Carlo simulation to obtain a probabilistic \ac{RUL} estimate for the insulation. The authors benchmark several alternative \ac{ML} models (including linear regression, gradient boosting, \ac{RF}, and \ac{MLP}) for the torque and temperature prediction tasks and select the best-performing configurations, but the hybrid \ac{RUL} framework itself is only validated on a single case study and is not quantitatively compared to other prognostics approaches.

\section{Discussion}
\label{sec:discussion}

% Outlook for the section to follow
This section discusses the main findings of the review, focusing on recurring methodological patterns, the observed effects of incorporating physics into \ac{ML}-based \ac{PHM}, and implications for different \ac{PHM} tasks. It then highlights contextual limitations, barriers to deployment, and terminological issues in the current literature and concludes by outlining promising directions for future research.

% - - - - - - - - - - subsection - - - - - - - - - - %
\subsection{Methodological Patterns}
\label{subsec:methodological_patterns}

% Outlook
Beyond the classification of individual methods (see Fig.~\ref{fig:sankey}), examining recurring meth\-odological patterns provides insight into conceptual maturity of the field, the emergence of shared modeling principles, and the extent to which physics integration has become methodologically standardized. The following analysis examines each class with respect to the diversity of implementation strategies, the strength of physics enforcement, and the implications for transferability and practical adoption.

% Observational bias 
\subsubsection{Observational Bias}
\label{subsub:methodological_patterns_observational_bias}

Methods incorporating observational bias rely on phys\-ics solely as a data source, while the model and training objectives remain conventional and agnostic to the underlying physics. Two dominant strategies emerge across the reviewed literature: using physics-based simulators to generate simulated data that enrich limited or imbalanced datasets~\citep{li2024Asimulationdatadriven, qin2024Inversephysicsinformed, qin2025SimulationdataDrivenGeneralized, ren2025Healthassessmentof, mei2024Ahybridphysicsinformed, kohtz2022Physicsinformedmachinelearning, deng2023ACalibrationBasedHybrid, zhu2024RemainingUsefulLife}, and pretraining on simulated data followed by fine-tuning on scarce real-world data, typically in a \ac{TL} setting~\citep{dong2022Anewdynamic, liu2024Enhancingmultitypefault, pettorossi2024AddressingDataScarcity, song2022Researchonfault, zhang2024AdversarialDomainAdaptation, li2024PhysicsGuidedDeepLearning, yishengliu2024HybridFusionfor, zhang2023DynamicModelAssistedBearing}. Delta learning, a less common variant in which a model trained on simulated data is corrected by a secondary model trained on experimental data, has also been explored \citep{thelen2022Integratingphysicsbasedmodeling, navidi2023PHYSICSINFORMEDNEURALNETWORKS}.

This apparent methodological uniformity is not a sign of community consensus but rather a consequence of the definitional boundary. Because physics can enter only through additional training data, the set of feasible implementation strategies is inherently small, regardless of the specific \ac{PHM} task or asset. This constraint simultaneously explains both the accessibility and the limitations of observational-bias approaches: they are straightforward to apply, provided that a simulator of sufficient fidelity can be constructed. Once training is complete, however, the simulator is discarded and the deployed model behaves like a conventional black box, retaining no mechanism to enforce physical plausibility at inference time. 

Ultimately, observational-bias approaches are well-positioned to address data scarcity within regimes covered by the simulated data, yet structurally ill-suited to deliver on stronger promises such as physical consistency and reliable extrapolation to unseen regimes---properties that require explicit structural or algorithmic enforcement.

% Inductive bias
\subsubsection{Inductive Bias}
\label{subsub:methodological_patterns_inductive_bias}

Methods incorporating inductive bias embed physical principles directly into the model through tailored architectural interventions. Consequently, they achieve the closest alignment between model structure and prior physical knowledge, albeit at the expense of being highly asset-specific. The resulting methodological landscape is correspondingly heterogeneous. Characteristic implementation patterns include: custom recurrent cells informed by specific degradation laws~\citep{yucesan2019Windturbinemain, dourado2019Physicsinformedneuralnetworks, dourado2022Ensembleofhybrid, nascimento2021Hybridphysicsinformedneural, badora2023Usingphysicsinformedneural}, graph topologies reflecting asset structure~\citep{liu2025Graphembeddedpatchsense, feng2022FullGraphAutoencoder, jin2024GraphSpatioTemporalNetworks, cheng2024Researchongas, zhou2025Physicsinformedspatiotemporalhybrid}, layers tailored to known frequency content~\citep{zeng2025ApplicationofFrequency, gao2024MPINet:MultiscalePhysicsInformed}, informed \ac{GP} priors~\citep{huang2022AnEnhancedDataDriven, zhu2023PhysicsinformedGaussianprocess, qiang2023Integratingphysicsinformedrecurrent}, or activation and output constraints enforcing monotonicity~\citep{hao2023Anoveldeep, li2025Applicationofphysicsguided,  yin2025Physicsguideddegradationtrajectory, zhou2023Timevaryingtrajectorymodeling}. 

While these patterns share broad conceptual goals, their implementations are fundamentally different and tightly coupled to the specific asset under study. Graph-based encodings naturally align with diagnostics in multi-component systems~\citep{liu2025Graphembeddedpatchsense, feng2022FullGraphAutoencoder, jin2024GraphSpatioTemporalNetworks}, whereas custom recurrent cells and informed \ac{GP} priors predominantly serve prognostics~\citep{yucesan2019Windturbinemain, badora2023Usingphysicsinformedneural, zhu2023PhysicsinformedGaussianprocess}, where degradation dynamics must be captured temporally. Monotonicity constraints represent a notable exception: because they encode a domain-agnostic property of irreversible degradation, they transfer readily across assets. Yet they provide only a weak inductive bias compared to the mechanistic models embedded in asset-specific architectures. 

Overall, research on inductive bias largely constitutes a proliferation of problem-specific solutions---the product of tailored model engineering rather than mature, reusable modeling strategies. While such tailoring yields strong performance and structurally enforces adherence to the embedded physical principles, it contributes little to developing reusable design patterns, leaving this class fragmented.

% Learning bias
\subsubsection{Learning Bias}
\label{subsub:methodological_patterns_learning_bias}

Learning-bias approaches are structurally more uniform, with physics incorporated almost exclusively as additional terms in the loss function, regularizing the model toward physically plausible solutions. Three main strategies emerge for introducing learning bias: the first relies on applying \acp{PINN}, as demonstrated across a broad range of assets, such as aero-engine spline couplings~\citep{wang2025Physicallyinformedhierarchical}, power transformers~\citep{ramirez2024ResidualbasedAttentionPhysicsinformed}, axial piston pumps~\citep{dong2025Innovativefaultdiagnosis, chao2025Physicsinformedneural, wangz.2023Physicsinformedneural}, and lithium-ion batteries~\citep{singh2023HybridModelingof, jang2025Stateofhealth, pengfeiwen2023PhysicsInformedNeuralNetworks, wang2025ABatteryState, liu2025Aphysicsguidedapproach, wang2024Physicsinformedneuralnetwork}. The second strategy hinges on embedding mechanistic models as soft constraints in otherwise standard \ac{ML} models, including crack growth and wear laws~\citep{pan2025Inservicefatiguecrack, badora2023Usingphysicsinformedneural}, half-cell or electrochemical aging models and equivalent-circuit dynamics~\citep{navidi2024PhysicsInformedMachineLearning, zhang2025AnElectrochemicalAgingInformed, xu2022Aphysicsinformeddynamic, pugalenthi2024RemainingUsefulLife, zhang2025APhysicsInformedHybrid}, and dynamic models of industrial robots derived from joint multi-energy and rigid-body dynamics~\citep{wang2024Adigitaltwin, li2024Hybridphysicsembeddedrecurrent}. Lastly, a third set of strategies relies on comparatively simple constraints, with the two most prominent patterns either imposing generic degradation properties such as monotonicity on health indicators or degradation trajectories~\citep{xu2024Physicsguideddeeplearning, deng2025ANovelMethod, freeman2022Physicsinformedturbulenceintensity, najeraflores2023APhysicsConstrainedBayesian, wang2024Phyformer:Adegradation, fassi2024PhysicsInformedMachineLearning} or encoding prior knowledge about fault signatures in the frequency domain~\citep{qiao2024APriorKnowledge, tang2024Apriorknowledgeenhanced, xu2024Physicsinformedprobabilisticdeep}. 

Among these, the \ac{PINN} formulation has become a \textit{de facto} standard for learning-bias integration, offering a unified way to weakly enforce the governing dynamics of a system, whether mechanical, electrochemical, or otherwise. Its abstract, problem-agnostic formulation makes it straightforward to implement across diverse domains, which has contributed to their widespread adoption---particularly in \ac{PHM}. Yet this convenience introduces a characteristic challenge: the relative weighting between data-fidelity and physics-residual losses. Across the reviewed literature, loss balancing is predominantly controlled by scalar weighting coefficients that are either fixed or tuned empirically, with only a few exceptions proposing more systematic schemes such as uncertainty-based weighting~\citep{pengfeiwen2023PhysicsInformedNeuralNetworks}, Bayesian optimization-based weighting~\citep{wang2025PhysicsInformedNeuralNetwork}, or gradient-norm-based balancing~\citep{singh2023HybridModelingof, zhang2025APhysicsInformedHybrid}. When these weights are poorly calibrated, the model may overfit the data while failing to satisfy physical constraints, or vice versa---a tension that remains an open practical challenge.

Notably, among all studies incorporating a learning bias, only a single study does so through the optimization procedure itself: \cite{yonastefera2025ConstraintGuidedLearningof} embed monotonicity, boundary, and energy-consistency constraints directly into the gradient-descent updates. Whether optimization-level integration of physics offers practical advantages over loss-based regularization (e.g., in settings with multiple competing constraints) remains an open question, given that only a single study has explored this pathway. However, the near-complete dominance of regularization-based approaches likely reflects practical considerations. Adding penalty terms is trivial in modern \ac{ML} frameworks, whereas implementing custom optimizers demands specialized expertise and is harder to generalize. 

Ultimately, a key trade-off emerges within this class: excluding qualitative degradation properties (e.g., monotonicity or boundedness), introducing learning bias inherently couples the \ac{PIML} formulation to the specific asset, trading the method's transferability across different scenarios for stronger constraints within the intended domain. An example of a more transferable design is the Koopman-informed \ac{NN} by \cite{wang2025KoopmanInformedNeuralNetwork}, which learns an approximately linear evolution in a latent eigenfunction space and regularizes forward-backward consistency of the dynamics without relying on asset-specific mechanistic models.

% Hybrid approaches
\subsubsection{Hybrid Approaches}
\label{subsub:methodological_patterns_hybrid_approaches}

Hybrid approaches couple independent physics-based and \ac{ML} models either in parallel or in series, without embedding physical knowledge within the \ac{ML} pipeline itself. Across all four classes, hybrid approaches exhibit the most pronounced methodological uniformity, arising from the limited structural design space: one model feeds the other, or both run concurrently with combined outputs.

% In-series (physics informs ML)
The dominant pattern is in-series coupling in which physics informs \ac{ML}. Several studies leverage calibrated physics-based models, namely \acp{ECM}~\citep{lin2025Physicsinformedmachinelearning, feng2024comprehensive}, electrochemical and aging models for batteries~\citep{ma2024Accurateandefficient, sun2023Adaptiveevolutionenhanced, shi2022Batteryhealthmanagement, xu2023ANovelHybrid}, fuel-cell models~\citep{pettorossi2025Physicsguidedfaultdiagnosis}, and wear and degradation laws~\citep{zhu2024PhysicsInformedDeepLearning, zhu2024RemainingUsefulLife} to derive intermediate quantities (e.g., model parameters or health indicators) that serve as inputs to an \ac{ML} model performing the actual \ac{PHM} task. In effect, the physics-based models act as \textit{mechanism-aligned} encoders that compress raw sensor data into compact, physically interpretable features. By reducing the input dimensionality to as few as six~\citep{lin2025Physicsinformedmachinelearning} or ten parameters~\citep{ma2024Accurateandefficient}, this in-series coupling enables simplifying the downstream learning task, justifying the use of simple models such as \acp{RF} and shallow fully connected networks.

% In-series (special cases)
Moreover, in three other cases, the physics-based model runs alongside a data-driven encoder, with their outputs concatenated and fed to a single \ac{ML} model. From the viewpoint of the \ac{ML} models performing fault diagnosis~\citep{singh2024Hybridphysicsinfused1DCNN}, health assessment~\citep{wang2024Physicalknowledgeguided}, and \ac{RUL} prediction~\citep{liang2024Ahybridapproach}, respectively, this configuration effectively retains an in-series coupling. This approach arises when the physics-based model is acknowledged as too low-fidelity to serve as the sole encoder (unlike the in-series approaches discussed above), yet still contributes structured information absent from raw data. In each case, the architecture implicitly decomposes the prediction task: the physics branch supplies trend-level or steady-state features while the data-driven branch captures residual dynamics or high-frequency fluctuations. Ablation studies in two of the three studies confirm that the combined approach outperforms either branch in isolation, while the third~\citep{singh2024Hybridphysicsinfused1DCNN} demonstrates clear performance gains over the data-driven branch alone.

% In-series (ML informs physics-based model)
The reverse coupling (\ac{ML} informing physics) is less common but mostly follows a consistent logic: the physics-based model requires as input a quantity that is not directly observable from available sensors, and the \ac{ML} model's sole task is to estimate precisely this quantity. Identified examples include a \ac{GPR} model that infers lifetime percentage to correct particle states in a Paris law-based crack-growth model~\citep{li2024Particlefilterbasedfatiguedamageprognosisusingprognosticaidedmodelupdating}, an \ac{RFR} that maps vibration features to the current pitting area for a Paris law-inspired pit-growth model~\citep{kundu2024Developmentofdatadriven}, a deep \ac{RL} agent that updates flow correction coefficients of a first-principles system-level digital twin of a fuel control system~\citep{xu2024Wearstateassessment}, and a Wasserstein \ac{GAN} that infers local fault evidence which is used to update node weights of the probabilistic fault root cause tracing network~\citep{xu2023Physicsguideddatarefinedfault}. Crucially, the
physics-based model retains responsibility for the mechanistic inference that
constitutes the actual \ac{PHM} task, thereby preserving physically consistent predictions.

% In-parallel coupling
In-parallel approaches are the least studied, with all identified examples addressing lithium-ion batteries, where data-driven residual models run alongside electrochemical and equivalent-circuit battery models~\citep{zhang2024Adaptivefaultdetection, kohtz2022PhysicsbasedMachineLearning, firoozi2022CylindricalBatteryFault}. All three studies share a common residual-correction scheme: a physics-based model (equivalent-circuit or electrochemical) provides a structured baseline prediction, while the \ac{ML} component (\ac{NN}, \ac{BiLSTM} or \ac{GPR}) explicitly learns the discrepancy between that prediction and the measured signal. The concentration of in-parallel approaches in battery applications likely reflects the availability of compact, well-understood physics models whose outputs are directly comparable to sensor measurements---a prerequisite for meaningfully defining a learnable residual.

% Interpretation
Hybrid approaches offer clear practical advantages: they require comparatively low engineering effort provided a physics-based model of sufficient fidelity exists. However, in configurations where the physics model informs a downstream \ac{ML} model, the latter remains a black box, inheriting the limitations of purely data-driven techniques---most notably the risk of physically implausible predictions and unreliable extrapolation. Nevertheless, hybrid approaches are still relatively common in the \ac{PHM} literature (see Fig.~\ref{fig:sankey}).

\subsubsection{Multi-Class Approaches}
\label{subsub:methodological_patterns_multi_class_approaches}

Although the four classes are conceptually distinct, the literature confirms that they are complementary in practice. Although studies have been assigned to multiple classes only if their mechanisms for incorporation are clearly separable, five works nevertheless span multiple classes \citep{sun2022MicrocrackDefectQuantification, li2025Applicationofphysicsguided, badora2023Usingphysicsinformedneural, wang2024Physicalknowledgeguided, zhu2024RemainingUsefulLife}, and in all cases a learning-bias component is present. 
This pattern likely reflects the comparatively low implementation barrier of adding physics-informed regularization, which naturally facilitates its combination with the other approaches. However, none of these multi-class studies report ablation experiments that isolate the contribution of each embedded mechanism. 

To conclude, the multi-class cases should primarily be interpreted as evidence that the four classes are practically compatible, rather than as proof that specific combinations outperform carefully designed single-class approaches. Establishing when and how multi-mechanism designs offer systematic advantages remains an open question.

% - - - - - - - - - - subsection - - - - - - - - - - %
\subsection{The Effects of Incorporating Physics}
\label{subsec:the_effects_of_incorporating_physics}

% --- Claim
Having characterized the methodological landscape in Section~\ref{subsec:methodological_patterns}, this section examines the empirical evidence for the effects of incorporating prior physical knowledge into \ac{ML}-based \ac{PHM}. The analysis is organized along six dimensions: predictive performance, generalizability, robustness, data efficiency, interpretability, and physical consistency. For each dimension, the strength and scope of available evidence is assessed, and---where possible---the observed effects are traced back to the class of incorporation. The section concludes by identifying systematic gaps and biases in how effects are currently reported.

% Predictive performance
\subsubsection{Predictive Performance}
\label{subsub:predtictive_performance}

Improved predictive performance constitutes the most consistently reported benefit and is supported across all four classes of approaches and all four \ac{PHM} tasks. For fault detection and diagnosis, numerous studies report higher classification accuracy relative to baselines~\citep{qiao2024APriorKnowledge, ma2025Aphysicsbasedsample, dong2022Anewdynamic, xu2024Physicsinformedprobabilisticdeep, zeng2025ApplicationofFrequency, gao2024MPINet:MultiscalePhysicsInformed, huang2025Physicsinformedcausallearning, zhu2024PhysiCausalNet:ACausaland}. For health assessment and prognosis, reduced regression errors are typical~\citep{yucesan2022Ahybridphysicsinformed, wang2025ABatteryState, mei2024Ahybridphysicsinformed, najeraflores2023APhysicsConstrainedBayesian}. While this breadth of evidence is encouraging, its interpretation requires caution for two reasons. 

First, the strength of the evidence varies substantially with experimental design. A minority of studies include clear ablation experiments comparing the proposed model against an architecturally identical counterpart from which only the physics component has been removed. The majority instead compare against simple baselines (e.g., standalone \ac{SVR}, \ac{RF}, or \ac{LSTM}), even when the proposed \ac{PIML} model is architecturally far more complex. Under such conditions, it is difficult to disentangle performance gains attributable to the embedded physics from those arising from increased model capacity, additional engineering effort, or more sophisticated training procedures. Comparisons against strong, state-of-the-art baselines are rare~\citep{wang2025KoopmanInformedNeuralNetwork, liu2025Graphembeddedpatchsense}, further limiting the conclusiveness of reported performance gains. 

Second, the magnitude of reported improvements is almost never contextualized with respect to practical significance. In \ac{PHM}, a small reduction in the \ac{RUL} prediction error may or may not alter maintenance decisions depending on the asset's failure consequences, the planning horizon, and the associated prediction uncertainty. Yet no study in the reviewed literature connects reported accuracy gains to downstream decision quality or maintenance cost savings. This omission limits the practical value of reported performance improvements for deployment contexts. 

Moreover, the prevalence of highly problem-specific solutions precludes drawing overarching conclusions about which type of prior knowledge or integration pathway yields the greatest performance gains in general.

\subsubsection{Generalizability}
\label{subsub:generalizability}

% Generalizability
Generalization, the ability to maintain performance under distributional shift, is a less frequently but repeatedly reported benefit.  Evidence emerges predominantly from two experimental protocols: either in \ac{TL} settings~\citep{wang2024Physicsinformedneuralnetwork, zhu2024PhysiCausalNet:ACausaland, zhang2023DynamicModelAssistedBearing, zhang2024AdversarialDomainAdaptation}, or across different operating and environmental conditions (such as varying loads, speeds, or temperatures)~\citep{zhu2024PhysicsInformedDeepLearning, abiria2025Highcycleandveryhighcycle, feng2022FullGraphAutoencoder, zeng2025ApplicationofFrequency, wang2025KoopmanInformedNeuralNetwork}. 

A critical distinction, seldom made explicit in the reviewed literature, is that between in-distribution and out-of-distribution generalization. The former refers to conditions spanned by the training data but held out for evaluation, whereas the latter refers to genuinely novel conditions, environments, or assets. Most reported generalization evidence pertains to the former. True out-of-distribution generalization, which represents the stronger and practically more relevant claim, is almost never targeted directly and remains largely unsubstantiated.

Observational-bias approaches are particularly well-represented among generalization claims. However, their purported generalization advantage must be interpreted carefully: by augmenting the training distribution with simulated data that spans a broader operating regime, the effective training distribution is expanded. Improvements relative to a data-driven baseline trained on less data therefore partly reflect the additional information injected via simulation, rather than a structural capacity to generalize. Without controlling for training-set coverage, such claims risk conflating data augmentation with genuine generalization ability.

% Robustness
\subsubsection{Robustness}
\label{subsub:robustness}

Robustness encompasses a broad range of notions in the \ac{ML} literature, including stability under input perturbations, resilience to label noise, and resistance to distributional shift. \cite{yin2025Physicsguideddegradationtrajectory}, \cite{zeng2025ApplicationofFrequency}, and \cite{singh2024Hybridphysicsinfused1DCNN}, for example, investigate robustness against noisy data. \cite{kristupasbajarunas2024HealthIndexEstimation} evaluate robustness to distributional shift by comparing \ac{HI} quality and \ac{RUL} prediction performance across methods. Other studies interpret robustness differently, including low variance in \ac{RUL} prediction errors across varying battery chemistries and operating conditions~\citep{ma2024Accurateandefficient} and the model’s ability to avoid false alarms~\citep{jin2024GraphSpatioTemporalNetworks}. In the latter case, however, evaluation was conducted on a single turbine, with only one documented true fault and a few false-alarm episodes, without broader tests across turbines, fault types, or operating conditions---limiting the generalizability of the finding. The inconsistent operationalization of robustness across studies precludes aggregating evidence into a coherent assessment. While \ac{PIML} shows promise in this dimension, results are fragmented and often context-specific, leaving the overall evidence inconclusive.

% Data efficiency
\subsubsection{Data Efficiency}
\label{subsub:data_efficiency}

Data efficiency, the ability to achieve a given performance level with fewer real observations, is rarely evaluated through dedicated experimental protocols. Where evidence exists, it typically derives from iteratively reducing dataset sizes and observing performance degradation~\citep{nguyen2023Physicsinfusedfuzzygenerative, wang2024Physicsinformedneuralnetwork, tang2024Apriorknowledgeenhanced, zhang2025APhysicsInformedHybrid}, or from limiting the available history or operational horizon~\citep{fu2024PhysicsInformedNeuralNetwork, zhu2024PhysicsInformedDeepLearning, e2025Aphysicsinformedneural, liang2024Ahybridapproach}. The available results generally suggest that incorporating physics largely preserves performance as data volume decreases.

Improved extrapolation capability can be regarded as a manifestation of data efficiency: by imposing physical constraints, a model can accurately recover solutions with fewer observations in poorly explored regions of the input space, as demonstrated by~\cite{xu2022Aphysicsinformeddynamic}, \cite{singh2023HybridModelingof}, and \cite{singh2024Hybridphysicsinfused1DCNN}. However, the same qualification noted above applies to observational-bias methods: the claim of requiring little real data is only partially justified when the underlying data is effectively augmented with simulated observations. The actual data budget (including simulation data generation, physics-model calibration, and real data collection) is rarely reported transparently, obscuring the true resource requirements.

% Interpretability
\subsubsection{Interpretability}
\label{subsub:interpretability}

Improved interpretability is among the most frequently claimed yet least substantiated benefits. In several instances, studies assert interpretability solely because prior physical knowledge has been incorporated, without providing any empirical evidence. Incorporating prior knowledge does not inherently make a model interpretable. Indeed, many \ac{PIML} models remain functional black boxes that offer negligible insight into how predictions are derived. 

A few exceptions demonstrate that at least parts of the model’s inner workings may exhibit a degree of interpretability, as shown by~\cite{xu2024Physicsinformedprobabilisticdeep}, where internal representations qualitatively align with known fault frequencies and uncertainty patterns. Several studies illustrate how integrating prior physical knowledge enhances the separability of learned representations between regimes, classes, or operating conditions, often visualizing this effect using methods such as t-distributed stochastic neighbor embedding~\citep{zhu2024PhysiCausalNet:ACausaland, cheng2024Researchongas, li2024PhysicsGuidedDeepLearning, huang2025Physicsinformedcausallearning, zeng2025ApplicationofFrequency, dong2022Anewdynamic, liu2025Graphembeddedpatchsense}. While such observations suggest that physics guides the model toward more structured feature spaces, improved cluster separability does not constitute interpretability in a rigorous sense. Consequently, interpretability remains an aspiration rather than a demonstrated outcome of current \ac{PIML} methods in \ac{PHM}.

% Physical consistency
\subsubsection{Physical Consistency}
\label{subsub:physical_consistency}

Physical consistency is frequently claimed but lacks a shared definition, rendering it context-dependent. As used in the reviewed literature, it encompasses: (i) adherence to fundamental physical laws (e.g., conservation principles, thermodynamics, or electrochemistry); (ii) conformity with established degradation properties (e.g., monotonicity, irreversibility, or bounded ranges); and (iii) admissibility of internal model variables (e.g., \ac{SOC} $\in [0, 1]$, crack length $\ge 0$, or temperature within certain limits). As established earlier, the four classes offer fundamentally different structural guarantees that must be considered when formulating or interpreting physical consistency claims (see Sec.~\ref{subsec:methodological_patterns}).

Beyond these class-level differences, several structural limitations deserve emphasis. First, many constraints enforce local behavior (e.g., step-to-step monotonicity), without guaranteeing globally realistic trajectories (e.g., correct knee behavior in battery aging). Second, \acp{PINN} enforce governing equations only at a finite set of collocation points. Third, constraints typically cover only a subset of the physics (e.g., simple wear law), while ignoring multi-physics coupling effects that may dominate in certain operating regimes. Consequently, partial enforcement of physical constraints can create a misleading impression of physically consistent behavior when the unmodeled physics becomes dominant. 

Ultimately, the field lacks a common metric for quantifying the degree of physical consistency of a model, which in turn presupposes consensus on what this notion precisely entails. Without such metrics, the claims of physical consistency remain qualitative and largely unverifiable.

% Evaluation integrity
\subsubsection{Caveats on Reported Effects}
\label{subsub:caveats_on_reported_effects}

Beyond the dimension-specific observations above, several cross-cutting issues affect the reliability of the overall evidence base. First, the absence of ablation experiments in many studies prevents isolating the contribution of incorporated physics from confounding factors such as architecture changes, additional hyperparameter tuning, or increased training data. This is particularly acute in studies spanning multiple classes~\citep{sun2022MicrocrackDefectQuantification, li2025Applicationofphysicsguided, badora2023Usingphysicsinformedneural, wang2024Physicalknowledgeguided, zhu2024RemainingUsefulLife}, where no ablation experiment disentangles the individual mechanisms.

Second, \ac{PIML} introduces a unique and under-recognized risk of evaluation leakage: when physics-model parameters are calibrated on data that overlaps with the test set---as in~\cite{abiria2025Highcycleandveryhighcycle}, where Basquin's and Paris' law parameters are fitted to the complete dataset before train-test splitting---the physics prior becomes partially informed by the test data itself. This systematically favors the physics-informed model over purely data-driven baselines and renders generalization claims overly optimistic. The vulnerability extends beyond this single instance: any \ac{PIML} approach that calibrates embedded physics-model parameters from data is susceptible to this form of leakage unless the calibration is strictly confined to the training partition.

Third, potential drawbacks of incorporating physics are almost never quantified. Computational cost (during training and inference), convergence behavior, sensitivity to loss-weight selection, and implementation overhead relative to purely data-driven alternatives are consistently omitted from evaluations. Among the reviewed studies, only \cite{liu2025Graphembeddedpatchsense} report \acp{FLOP} for all compared models. The widespread absence of such information represents a significant barrier to informed method selection and yields a strongly benefit-skewed evidence base that is insufficient as a foundation for balanced deployment decisions.

% - - - - - - - - - - subsection - - - - - - - - - - %
\subsection{Contextual Limitations}
\label{subsec:contextual_limitations}

% Overview of the most studied assets
Having discussed the effects of incorporating physics into ML in Section~\ref{subsec:the_effects_of_incorporating_physics}, this section highlights the contextual limitations that must be considered when assessing the maturity and generalizability of \ac{PIML} in \ac{PHM}. It first documents a pronounced concentration of the reviewed literature around a narrow set of assets, then traces this imbalance to two reinforcing drivers, namely data availability and the availability of formalized prior physical knowledge, and finally discusses the resulting implications for maturity assessment.

% Asset concentration
Across the reviewed literature, lithium-ion batteries and bearings clearly emerge as the primary focus, accounting for approximately half of the total studies (see Fig.~\ref{fig:sankey}). Only a small subset of other assets (i.e., cutting tools, pumps, turbofan engines, and metal specimens) has been investigated in five or more studies, highlighting their relatively limited attention. The remaining assets are examined sporadically, often in a single study, underscoring a significant gap in research coverage. This concentration likely reflects the convergence of multiple factors: the industrial and commercial significance of batteries and bearings, the maturity of their respective research communities, and (as discussed below) the favorable availability of both public datasets and well-characterized physics-based models for these assets. These factors are mutually reinforcing rather than independent.

% Availability bias
\ac{ML} research is strongly shaped by data availability, which in turn influences both the problems studied and the methods developed. The work of ~\cite{mauthe2025overview_esrel} presents the most comprehensive overview and analysis of publicly available degradation datasets for \ac{PHM}, covering 98 datasets in total. The ongoing updating of this overview, along with complete documentation for each dataset, is available online~\citep{mauthe2024overview_arxiv}. Notably, batteries and bearings form the two largest asset categories in terms of dataset count, with 15 each. By contrast, most other asset types are represented by only one or two datasets. This imbalance means that researchers reliant on publicly available data encounter a markedly richer landscape for batteries and bearings than for other assets. In addition, specific datasets (such as the battery dataset provided by the NASA Prognostics Center of Excellence~\citep{saha2007battery}, XJTU-SY~\citep{wang2020hybrid} and FEMTO~\citep{nectoux2012pronostia} for bearings, or C-MAPSS for turbofan engines~\citep{saxena2008turbofan}) have acquired the status of \textit{de facto} community benchmarks, further concentrating research activity around the assets they represent. The reviewed literature reflects this pattern, drawing heavily on publicly available datasets and exhibiting a similar asset distribution. Consequently, the predominance of studies on lithium-ion batteries and bearings is, at least in part, reinforced by an availability bias.

% Methodological selection bias
The review reveals a dependency between the degree of formalization of available prior physical knowledge and the range of incorporation strategies that become feasible. Lithium-ion batteries and bearings are particularly amenable to \ac{PIML} because their underlying physics is well-characterized and formalized. For batteries, electrochemical models---including \ac{SPM} variants~\citep{yishengliu2024HybridFusionfor, singh2023HybridModelingof, zhang2025AnElectrochemicalAgingInformed, zhang2025APhysicsInformedHybrid}, half-cell~\citep{navidi2024PhysicsInformedMachineLearning, thelen2022Integratingphysicsbasedmodeling} and \ac{SEI}-growth models~\citep{kohtz2022PhysicsbasedMachineLearning, liu2025Aphysicsguidedapproach}---are widely employed, alongside \acp{ECM} for cell voltage and \ac{SOC} dynamics~\citep{liu2024Enhancingmultitypefault, fu2024PhysicsInformedNeuralNetwork, qin2025ManagingBatteryPerformance}. For bearings, multi-\ac{DOF} dynamic models~\citep{dong2022Anewdynamic, qin2024Inversephysicsinformed, qin2025SimulationdataDrivenGeneralized, deng2023ACalibrationBasedHybrid, zhang2023DynamicModelAssistedBearing, sun2024Contrastivelearningand}, analytical vibration signal models~\citep{li2024Asimulationdatadriven, zhu2024RemainingUsefulLife, gao2024FaultDiagnosisof}, and fault characteristic frequencies \citep{zeng2025ApplicationofFrequency, qiao2024APriorKnowledge, xu2024Physicsinformedprobabilisticdeep, gao2024MPINet:MultiscalePhysicsInformed} provide a rich repository of embeddable prior knowledge. Accordingly, the literature appears to be influenced, at least in part, by a methodological selection bias. This creates a self-reinforcing pattern in which assets whose physics is already well-formalized attract disproportionate research attention, while those whose degradation involves poorly understood or multi-physics mechanisms remain underrepresented.

% Implication
These contextual limitations must inform any evaluation of the current maturity of \ac{PIML} in \ac{PHM}. While the mere volume of studies covered in this review may suggest that \ac{PIML} has matured into an established standard for industrial \ac{PHM}, such an interpretation would be misleading. The observed imbalance necessitates a more differentiated assessment. For lithium-ion batteries and bearings, a certain level of methodological maturity can reasonably be claimed. However, even  for these assets, existing studies focus predominantly on specific \ac{PHM} tasks: health assessment and prognosis for batteries; diagnosis and prognosis for bearings. Thus, the apparent maturity is confined to a few asset-task combinations rather than the full \ac{PHM} spectrum. Beyond these focal assets, \ac{PIML} research remains at an early, exploratory stage with only scattered and often isolated evidence. The skew in asset coverage is particularly consequential because \ac{PIML} methods are generally tied to the specific asset under study, owing to the use case-specific nature of the embedded prior knowledge. This entanglement constrains applicability to the studied context, and, by extension, raises the question of how to leverage \ac{PIML} without sacrificing generalizability.

% - - - - - - - - - - subsection - - - - - - - - - - %
\subsection{Barriers to Deployment}
\label{subsec:barriers_to_deployment}

The concentration of the literature around a narrow set of assets and tasks (see Sec.~\ref{subsec:contextual_limitations}) already constitutes a barrier to the broader deployment of \ac{PIML}-based \ac{PHM}. Even within these well-studied domains, however, a substantial gap separates current proofs of concept from industrially deployable solutions. This section examines the practical barriers that collectively account for this gap. These barriers are not independent but cumulative: constructing a \ac{PIML} solution demands significant expertise and engineering effort; even where such effort is invested, the resulting models rarely provide decision-relevant uncertainty estimates; and even if both of the former barriers were overcome, insufficient evidence exists regarding whether these models can operate under real-time and resource constraints.

\subsubsection{Implementation Effort and Expertise}
\label{subsub:implementation_effort_and_expertise}

A prerequisite for any \ac{PIML} solution is the successful identification, formalization, and embedding of appropriate prior physical knowledge---a process that remains largely undocumented and unquantified across the reviewed literature. Unlike purely data-driven pipelines, which can often be constructed by \ac{ML} practitioners with general domain familiarity, \ac{PIML} demands expertise at the intersection of two traditionally separate disciplines: the physics of the asset's degradation mechanisms and the engineering of \ac{ML} architectures and training procedures. This dual requirement manifests at multiple stages: selecting which physics to embed (and, equally importantly, which to omit); translating qualitative physical understanding into a formal, computable representation amenable to integration; choosing an appropriate pathway (observational, inductive, or learning bias); and calibrating the interplay between data-driven and physics-informed components, such as loss-weight tuning or simulator fidelity.

No study among those reviewed reports the human effort, development time, or iterative design cycles required to arrive at the final physics-informed model. Yet this engineering overhead is arguably the most immediate practical barrier, particularly when seeking to deploy \ac{PIML} at scale across heterogeneous asset fleets. The observation from Sections~\ref{subsec:methodological_patterns} and \ref{subsec:contextual_limitations} that nearly all solutions are tightly coupled to the specific asset under study is, in part, a downstream consequence of this barrier: each new asset demands the aforementioned integration effort, which cannot be easily amortized.

Incorporating more general prior knowledge, such as simple monotonicity constraints, can broaden applicability, although the resulting performance gains may often be modest. Developing methods that are both applicable across diverse contexts (or even assets) and capable of delivering substantial improvements thus remains a key challenge. Nevertheless, three patterns in the literature suggest pathways toward reducing this overhead by circumventing the need for intricate physical models: (i) structural and topological knowledge describing an asset's component layout or the relative placement of sensors is particularly suitable for graph-based approaches~\citep{feng2022FullGraphAutoencoder, liu2025Graphembeddedpatchsense, jin2024GraphSpatioTemporalNetworks, cheng2024Researchongas, zhou2025Physicsinformedspatiotemporalhybrid}; (ii) qualitative degradation properties (e.g., monotonicity) require no system-specific physical model and can be imposed via architectural constraints, such as monotone activations, constrained hidden-state updates, and bounded output layers~\citep{hao2023Anoveldeep, li2025Applicationofphysicsguided, abiria2025Highcycleandveryhighcycle, bai2023PrognosticsofLithiumIon, yin2025Physicsguideddegradationtrajectory, zhou2023Timevaryingtrajectorymodeling}, or inequality-type loss terms that penalize local violations~\citep{deng2025ANovelMethod, wang2024Physicsinformedneuralnetwork, fassi2024PhysicsInformedMachineLearning, najeraflores2023APhysicsConstrainedBayesian}; and (iii) causal relationships between operating conditions, sensor signals, and the underlying degradation state, although studied less frequently~\citep{kristupasbajarunas2024HealthIndexEstimation}. Beyond this, very few studies actually demonstrate applicability across multiple use cases without requiring modifications~\citep{tang2024Apriorknowledgeenhanced, kristupasbajarunas2024HealthIndexEstimation, wang2025KoopmanInformedNeuralNetwork, wang2024Phyformer:Adegradation, zhou2023Timevaryingtrajectorymodeling}, and in each case separate training is still required. 

Taken together, these structural, qualitative, and causal forms of prior knowledge represent the most accessible entry points for transferable \ac{PIML} in \ac{PHM}. However, transferable \ac{PIML} solutions remain an aspiration rather than an established practice, with the high implementation cost per asset being a principal inhibitor.

% Uncertainty quantification
\subsubsection{Uncertainty Quantification}
\label{subsub:uncertainty_quantification}

Uncertainty quantification is widely recognized as a core requirement for prognostics in \ac{PHM}, where single-point estimates are \enquote{usually considered meaningless} for industrial applications~\citep{kundu2020Areviewon}. Yet the vast majority of reviewed studies produce exclusively deterministic predictions, offering no calibrated confidence information to inform decision-making.

A smaller subset of studies combines physics with probabilistic models~\citep{bai2023PrognosticsofLithiumIon, jiang2025PhysicsinformedGaussianprocess, ellis2022Ahybridframework}. In addition, Bayesian \acp{NN} are used, though to a lesser extent~\citep{deng2023ACalibrationBasedHybrid, liang2024Ahybridapproach, najeraflores2023APhysicsConstrainedBayesian}. In another study, a Wiener process-based stochastic degradation model is embedded in an \ac{NN}~\citep{he2025Physicsinformedneuralnetwork}. However, among these, very few explicitly investigate how the incorporation of prior physical knowledge improves the quality of uncertainty estimates relative to purely data-driven probabilistic models~\citep{bai2023PrognosticsofLithiumIon, xu2024Physicsinformedprobabilisticdeep}. This represents a missed opportunity, because physics-informed constraints (including inductive and learning bias) have the potential to sharpen predictive distributions, e.g., by ruling out physically implausible predictions and thus yielding tighter confidence bounds.

Conversely, this same mechanism introduces a distinctive risk: in operating regimes where the embedded physics becomes inaccurate (e.g., due to unmodeled multi-physics coupling, degradation-mode transitions, or environmental conditions outside the model's validity range), overly constrained physics-informed predictions may produce dangerously miscalibrated uncertainty estimates. However, neither improvements nor deteriorations in uncertainty estimates resulting from the incorporation of prior physical knowledge have been sufficiently studied in the current literature. As a result, \ac{PIML} models in \ac{PHM} rarely provide the reliable, calibrated uncertainty information needed for decision-making, representing a critical barrier to deployment in settings where decisions carry safety or financial consequences.

% Computational cost
\subsubsection{Computational Cost and Operational Readiness}
\label{subsub:computational_cost_and_operational_readiness}

The transition from offline validation to operational deployment introduces requirements that the current literature leaves largely unaddressed: real-time inference under latency constraints, execution on resource-limited hardware, adaptation to evolving conditions, and integration with existing monitoring and maintenance infrastructure. Assessing the feasibility of meeting these requirements presupposes insight into computational costs, yet such information is largely absent from the reviewed studies.

Among all reviewed studies, only \cite{liu2025Graphembeddedpatchsense} report \acp{FLOP} for all compared models, thereby enabling fully transparent computational assessment. Partial reporting is more common but insufficient: \cite{fu2024PhysicsInformedNeuralNetwork} and \cite{xie2024DegradationStateAssessment} report \acp{FLOP} for their proposed physics-informed models but not for baselines. \cite{zeng2025ApplicationofFrequency} and \cite{pugalenthi2024RemainingUsefulLife} evaluate efficiency solely in terms of training time without reporting \acp{FLOP} or parameter counts, limiting the comparability of their efficiency claims across studies. \cite{wang2025ABatteryState} additionally report inference times, revealing that their \ac{PINN}-based approach incurs training times approximately two orders of magnitude above the fastest baseline, while inference times remain comparable across models. These scattered observations do not enable systematic comparison of performance gains versus computational overhead across the field.

Nevertheless, the inherent characteristics of each class of approaches permit a qualitative assessment of computational trade-offs. Observational-bias approaches shift the computational burden to an offline simulation phase: once the ML model is trained, inference cost is identical to a purely data-driven model. Inductive-bias approaches embed physics directly in the architecture, where the added inference cost depends on the specific mechanism---ranging from negligible (e.g., constrained activations) to non-trivial (e.g., embedded \ac{ODE} solvers). Learning-bias approaches, particularly \acp{PINN}, increase training cost through additional loss evaluations and automatic differentiation but typically incur no overhead at inference, unless online retraining is required. Hybrid approaches are unique in retaining the physics-based model during inference, which can become a bottleneck for real-time deployment when the physics model is computationally expensive. These qualitative distinctions provide initial guidance for practitioners facing real-time engineering decisions, while underscoring the urgent need for future studies to systematically report computational metrics alongside predictive performance.

Beyond computational cost in isolation, several interrelated deployment requirements remain entirely unaddressed. First, virtually no study reports inference latency, throughput, or memory footprint under conditions representative of industrial monitoring systems. Second, experiments investigating whether current \ac{PIML} models can be executed on resource-constrained hardware such as embedded edge devices are missing. Third, the coupling of \ac{PIML} models with operational infrastructure (e.g., programmable logic controllers, supervisory control and data acquisition systems, cloud- and edge-based data pipelines, or maintenance management systems) is never discussed. Fourth, the question of model maintenance after deployment, including adaptation when operating conditions drift, physics assumptions degrade, or new failure modes emerge, is entirely unstudied. Based on the experimental settings reported (i.e., predominantly laboratory datasets, offline evaluations, and controlled conditions), the reviewed methods appear to correspond broadly to early \acp{TRL} (approximately \ac{TRL} 3--4), with no study demonstrating operational deployment.

% - - - - - - - - - - subsection - - - - - - - - - - %
\subsection{Terminology}
\label{subsec:terminology}

% Inconsistent use of terminology regarding the decriptor "physics-informed"
The review highlights a notable inconsistency in the terminology regarding \ac{PIML} across \ac{PHM} studies. Within this paradigm, many researchers introduce novel contributions by combining the descriptor \enquote{physics-informed} with a term denoting their specific model or approach. As a result, the descriptor \enquote{physics-informed} emerges as the most frequently used term among the reviewed studies, effectively signaling that prior physical knowledge is incorporated into the \ac{ML} pipeline in some form. This dominance, however, may have prompted some researchers to use other descriptors better reflecting the particular characteristics of their approach, such as \enquote{physics-guided}~\citep{li2025Applicationofphysicsguided} or \enquote{physics-constrained}~\citep{najeraflores2023APhysicsConstrainedBayesian}---a phenomenon known to the authors prior to conducting this review, which also influenced the keyword selection (see Tab.~\ref{tab:keywords}). While the choice of the respective descriptor may be appropriate in some instances, it is rarely accompanied by an explicit explanation. The wide variety of terms used is, on the one hand, likely a consequence of the relatively recent emergence of \ac{PIML} as a research field, particularly in older studies. On the other hand, in certain cases, the descriptor appears to reflect an arbitrary choice of wording. In milder instances, this results in descriptors other than \enquote{physics-informed} being used consistently within a single study---individually unproblematic, yet collectively contributing to terminological fragmentation across the field. In more extreme cases, however, multiple variants may appear within the same study. For instance, \cite{freeman2022Physicsinformedturbulenceintensity}  refer to their loss function as \enquote{physics-informed,} \enquote{physics-guided,} and \enquote{physics-based,} which likely reflects an attempt to employ synonyms for stylistic variation. While inconsistent terminology across studies is understandable, given that the seminal work by \cite{karniadakis2021physics} was published in 2021, inconsistencies within individual studies remain problematic, as they can obscure the intended message and compromise clarity for the reader. 

% Using "physics-informed" and the issue with the term PINN
When the descriptor \enquote{physics-informed} is used consistently, an ambiguity naturally arises with the term \ac{PINN}---a pattern repeatedly observed across the reviewed studies. Since \cite{raissi2019physics} coined the term to describe \acp{NN} that incorporate known physical laws, expressed as differential equations, into their loss function, the term \ac{PINN} is somewhat constrained in its usage. Several studies make use of this term, yet their approaches do not align with its original definition~\citep{ badora2023Usingphysicsinformedneural, deng2025ANovelMethod, navidi2023PHYSICSINFORMEDNEURALNETWORKS}. In light of the highly influential work of \cite{raissi2019physics}, many readers may reasonably expect the term \ac{PINN} to imply the incorporation of differential equations for regularization, creating potential confusion when used differently. It is acknowledged, however, that establishing clear terminology for novel approaches that are both physics-informed and employ \acp{NN}, yet do not align with the concept of \acp{PINN}, remains a challenge.

% Synonymously using the terms "hybrid" and "physics-informed" in the context of PHM
Since approaches to implementing \ac{PHM} are typically subdivided into model-based, data-driven, or hybrid, it is understandable that some studies use the terms \enquote{physics-informed} and \enquote{hybrid} interchangeably (e.g., \cite{lehmann2024LearningtheAgeing}), given that the latter is generally regarded as a broader category encompassing \ac{PIML}. While this may be technically correct, the classification scheme adopted in this review makes an explicit distinction between \ac{PIML} and hybrid approaches to further enhance clarity in this regard, thereby facilitating more precise communication regarding the methodology employed in each study. In this context, the combined use of these terms within a single phrase can be misleading, such as \enquote{hybrid physics-informed neural network}~\citep{dourado2022Ensembleofhybrid}, \enquote{physics-informed spatio-temporal hybrid neural network}~\citep{zhou2025Physicsinformedspatiotemporalhybrid}, or \enquote{hybrid physics-embedded recurrent neural network}~\citep{li2024Hybridphysicsembeddedrecurrent}. This does not pose a problem when the corresponding study actually combines a hybrid approach that incorporates bias via one of the three \ac{PIML} pathways.

% Inconsistent use of terminology regarding the research field of PIML itself
Apart from the use of descriptors to specify the proposed approach, some studies even apply them to the entire field of research, disregarding the foundational definition of \ac{PIML} by \cite{karniadakis2021physics}. While these alternative terms still refer to research that effectively incorporates prior physical knowledge into \ac{ML}, this practice appears to be related to the earlier-discussed issue of unnecessary lexical variation. Although there is some flexibility in naming a novel approach differently, using alternative terms for the research field may give the impression of a subtle yet notable distinction regarding \ac{PIML}. For example, \cite{kohtz2022PhysicsbasedMachineLearning} propose employing \enquote{physics-based machine learning techniques.} \cite{bai2023PrognosticsofLithiumIon} deviate even further, referring to it as \enquote{knowledge-constrained machine learning.} \cite{yan2025KnowledgeDrivenMachine} claim to introduce a novel taxonomy, named \enquote{knowledge driven machine learning,} intended to synthesize the current state of research at the intersection of prior knowledge and \ac{ML}. They explicitly adopt the taxonomy of \cite{vonrueden2021informed}, yet contribute no novelty and could therefore have been referred to simply as \ac{IML}. \cite{yin2025Physicsguideddegradationtrajectory} reference~\cite{karniadakis2021physics} to outline the three pathways for introducing physics into \ac{ML}. In doing so, they fully adopt the conceptualization of \ac{PIML}, yet deliberately use the term \enquote{physics-guided} throughout their entire study---a choice for which no clear rationale is provided.

% - - - - - - - - - - subsection - - - - - - - - - - %
\subsection{Future Research}
\label{subsec:future_research}

The preceding analysis reveals that \ac{PIML} already delivers tangible performance benefits for \ac{PHM}, yet persistent methodological gaps (see Sec.~\ref{subsec:methodological_patterns}), insufficiently substantiated claims (see Sec.~\ref{subsec:the_effects_of_incorporating_physics}), narrow asset coverage (see Sec.~\ref{subsec:contextual_limitations}), practical deployment barriers (see Sec.~\ref{subsec:barriers_to_deployment}), and terminological inconsistencies (see Sec.~\ref{subsec:terminology}) collectively constrain the field's advancement. The following directions are organized to address each of these gaps in turn.

% 5.1 - benchmarking PIML integration pathways
The review demonstrates the potential of \ac{PIML} for \ac{PHM}, yet the analyzed studies also reveal persistent methodological gaps and open challenges, as discussed in Section~\ref{subsec:methodological_patterns}. Addressing these issues will be essential to translate current approaches into robust, deployable solutions for industrial practice. Building on the synthesis and critical discussion presented above, one key opportunity for future research is the design of benchmark studies that systematically compare observational-, inductive-, and learning-bias approaches under controlled conditions. Wherever feasible, these benchmarks ought to reuse the same prior physical knowledge so that differences in performance can be attributed to the pathway of integration rather than to the physics itself. The resulting evidence would enable well-founded guidelines for selecting an integration strategy based on the type and fidelity of available prior knowledge, the volume and quality of data, and application-level requirements such as robustness and interpretability.

% 5.1 - standardizing inductive-bias approaches, and exploring learning-bias approaches
Tackling the highly heterogeneous landscape of induc\-tive-bias approaches (see Sec.~\ref{subsub:methodological_patterns_inductive_bias}) requires moving from problem-specific designs toward modular building blocks that can be reused with minimal adaptation---analogous to the \ac{PINN} framework for learning-bias approaches. Ultimately, the goal should be to reduce the often-overlooked engineering overhead of \ac{PIML}, making inductive-bias approaches more practical and scalable. Moreover, regularization-based methods dominate the landscape of learning-bias approaches (see Sec.~\ref{subsub:methodological_patterns_learning_bias}), yet reliance on empirically tuned loss balancing suggests a shift toward adaptive methods. While aiming to stabilize training and promote convergence, such methods should ultimately ensure that the physical loss contributes appropriately, preventing the model from over-prioritizing the minimization of data loss. To gain more precise control over both training dynamics and the influence of physics on model training, future research should additionally explore integrating physics directly into the optimizer. Although challenging to implement, physics-informed optimizers can produce solutions that are both accurate and physically plausible by incorporating constraints directly into the optimization step, providing stronger adherence than loss-based regularization.

% 5.2 - sound experiment design to substantiate claims of improvement
While promising, the claimed improvements from incorporating prior physical knowledge largely remain unsubstantiated beyond predictive performance, and require more rigorous experimental design (see Sec.~\ref{subsec:the_effects_of_incorporating_physics}). To enable fair and informative comparisons, models built on simpler architectures should be benchmarked against conventional baselines, whereas more powerful architectures should be compared to state-of-the-art alternatives. In all cases, ablation studies are essential to isolate the specific contribution of the embedded physics. Moreover, experiments need to encompass diverse operating and environmental conditions to rigorously evaluate the validity of the claims---an aspect particularly crucial in the context of industrial \ac{PHM}. In addition, physical consistency often lacks a precise formulation. To facilitate meaningful comparisons, it is necessary to establish a formal definition and derive corresponding metrics. These could quantify the frequency and magnitude of constraint violations, deviations from physically plausible ranges, and cumulative errors over time, enabling a nuanced assessment of how well models respect physical laws while maintaining predictive accuracy. Finally, to counter the prevailing benefit-skewed view, evaluations should routinely quantify the trade-offs and costs of \ac{PIML} (e.g., computational load, convergence behavior, training stability, and engineering effort) alongside any performance gains. This would facilitate a balanced assessment of whether physics integration justifies its associated overhead in a given deployment context.

% 5.3 - generalizable approaches
Although deeper study of different asset types may mitigate the contextual limitations identified in Section~\ref{subsec:contextual_limitations}, prioritizing generalizable solutions over asset-specific studies represents a more productive allocation of research effort. By systematically analyzing and abstracting the mechanisms that have proven most effective, researchers can derive models applicable across entire classes of assets. This challenge hinges on shared fundamental principles governing diverse degradation phenomena, highlighting the need to study trade-offs between asset-specific and general prior knowledge to understand how performance is gained or sacrificed in pursuit of broader applicability. A key enabler in this regard is modular design patterns that facilitate adaptation to new assets without requiring substantial engineering overhead. Taken to its logical conclusion, this points to the development of physics-informed foundational degradation models that are trained in a multi-task fashion to simultaneously address all core \ac{PHM} tasks across multiple assets and operating conditions. Such models would essentially function as universal backbones, obviating laborious  problem-specific development.

% 5.4 - barriers to deployment
Translating \ac{PIML} from research to industry faces the deployment barriers discussed in Section~\ref{subsec:barriers_to_deployment}, necessitating progress along multiple interrelated axes. Complementing the efforts toward transferable solutions outlined above, automated or semi-automated methods for selecting, calibrating, and embedding prior physical knowledge would lower the entry barrier for practitioners who possess domain expertise but lack specialized \ac{ML} engineering skills. 

Moreover, uncertainty quantification must shift from an optional addition to an integral design objective. Future work should both systematically investigate how incorporated physics affects the quality of uncertainty estimates and develop diagnostic indicators that signal when a model's physics assumptions are being violated, providing actionable safeguards against overconfident predictions in safety-critical scenarios.

Beyond this, future reporting should adopt full transparency regarding computational costs, while research should concurrently investigate lightweight physics-informed architectures suitable for execution on resource-constrained edge hardware, including model compression and pruning techniques.

Moreover, while generalization from simulation data to test bench data is frequently studied, the subsequent transfer to field data remains unaddressed. Future studies must demonstrate generalization to customized industrial machines, especially in settings where idealized physical laws may no longer apply because environmental influences or multi-component interactions are superimposed on the embedded prior knowledge. By extension, models must be capable of updating as operating conditions drift or new failure modes emerge, pointing to research on online adaptation mechanisms. 

Finally, demonstrating closed-loop integration with industrial monitoring and maintenance infrastructure (e.g., programmable logic controllers, supervisory control and data acquisition systems, or cloud-edge data pipelines) would constitute a critical step toward elevating \ac{PIML}-based \ac{PHM} from laboratory validation to operational maturity.

% 5.5 streamlining terminology
In future research, reaching a consensus on core terminology would facilitate clearer communication, more consistent methodology, and more meaningful synthesis of findings across studies. Addressing this challenge (identified in Sec.~\ref{subsec:terminology}) entails both conceptual and practical efforts: conceptually, by establishing precise definitions---such as that of physical consistency---which can underpin the development of relevant metrics, and practically, by adhering to a consistent naming convention when introducing novel methods. With respect to the latter, the results suggest adhering to \enquote{physics-informed} as the descriptor, in line with \ac{PIML}. If deviation from this term is justified, either an explicit explanation should be provided or the alternative term should be unambiguous by default, as exemplified by the \enquote{Koopman-informed neural network}~\citep{wang2025KoopmanInformedNeuralNetwork}. In both cases, it is crucial that the chosen term be used consistently throughout the study.

% - - - - - - - - - - section - - - - - - - - - - %
\section{Conclusion}
\label{sec:conclusion}

% Echoing the introduction
\ac{PHM} is increasingly expected to provide reliable, trustworthy, and data-efficient decision support from sparse, noisy, and heterogeneous data. Given these demands, this review set out to examine how \ac{PIML} is currently leveraged in \ac{PHM} by systematically addressing several research questions that explore the prior knowledge employed, its incorporation, and corresponding implications for practice. By conducting the most comprehensive systematic literature review to date at the intersection of \ac{PIML} and \ac{PHM}, covering 212 studies, this work provides comprehensive answers to these questions.

% Answering the research questions posed in the introduction
% Research question 1 - Knowledge
% (a) What types of prior physical knowledge are being leveraged, and [...]
\begin{enumerate}
    \item \textbf{Knowledge} (a) What types of prior physical knowledge are being leveraged?
\end{enumerate}

In terms of prior physical knowledge, the field relies predominantly on mechanistic models and explicit degradation laws, whereas structural and causal relationships and qualitative degradation properties see less frequent application. 

% (b) what forms of representation are employed?
\begin{enumerate}
    \item \textbf{Knowledge} (b) What forms of representation are employed?
\end{enumerate}

Consequently, the corresponding forms of representation span from highly formalized expressions, through empirical and phenomenological formulations, down to implicit forms that capture principled assumptions. While the current focus capitalizes on well-established physical understanding, evidence from the reviewed literature indicates that prior knowledge across the full spectrum of types and representations can be productively leveraged, with each type contributing differently to the trade-off between physical fidelity and transferability.

% Research question 2 - Incorporation
% (a) How can prior physical knowledge be incorporated, and [...]
\begin{enumerate}[start=2]
    \item \textbf{Incorporation} (a) How can prior physical knowledge be incorporated?
\end{enumerate}

Methodologically, three overarching pathways constitute distinct yet complementary classes of approaches to incorporating prior physical knowledge: observational bias, inductive bias, and learning bias. Among these, learning-bias approaches are the most widely adopted, followed by inductive-bias and observational-bias approaches. This pattern likely reflects the advantages of learning-bias approaches in balancing flexibility regarding the types of prior physical knowledge that can be incorporated with practical implementation feasibility, while still imposing soft constraints that effectively regularize model behavior. In contrast, observational-bias approaches require a simulator of sufficient fidelity, whereas inductive-bias approaches face inherent difficulties in their tailored implementation---both presenting practical challenges that limit, to some extent, their adoption. 

In terms of methodological maturity, observational-bias approaches leave little room (by definition) for alternative ways of incorporating physics. The remaining two pathways encompass broader design spaces, which account for the observed heterogeneity in how the corresponding methods are implemented. Although distinct schemes for introducing inductive bias have emerged, they share only broad conceptual similarities. A similar pattern is observed regarding learning bias, where prior knowledge is almost exclusively embedded via additional loss terms, though the form and weighting of these constraints vary widely. In parallel to these three pathways, hybrid approaches form a conceptually distinct fourth class, where a physics-based model and an \ac{ML} model are either coupled in parallel or in series. Although \ac{PIML} is generally subsumed under \textit{hybrid} in the context of \ac{PHM}, distinguishing physics-informed from hybrid approaches elucidates the distinct ways in which prior knowledge is applied: the former integrate prior knowledge directly into the ML pipeline itself, whereas the latter retain an explicit stand-alone physics-based model that remains an integral part of the prediction pipeline during inference. Yet hybrid approaches constitute the least frequently adopted class.

% (b) how does the form of representation influence which approaches to incorporation are feasible?
\begin{enumerate}[start=2]
    \item \textbf{Incorporation} (b) How does the form of representation influence which approaches to incorporation are feasible?
\end{enumerate}

In terms of feasibility, the form in which prior physical knowledge is represented is not a minor implementation detail but the primary design lever that determines which pathways are viable. Mechanistic models are flexible enough to support all four classes of approaches, while being most readily incorporated as observational bias or in a hybrid setting. Structural and topological information naturally maps to inductive bias via graph-based approaches, whereas qualitative properties are predominantly expressed as learning bias and, in some cases, as simple architectural constraints. As indicated earlier, this creates a systematic trade-off: richer, more formal representations permit closer alignment with physics but require substantial modeling effort and are often asset-specific, whereas qualitative properties are inherently limited in their physical fidelity yet typically retain broad applicability. Accordingly, with development largely being shaped by the form of representation, the effort to transform prior knowledge into suitable representations becomes instrumental in opening previously inaccessible pathways---an aspect that has received little attention.

% Research question 3 - Practice
% (a) How does incorporating prior physical knowledge help overcoming limitations regarding purely data-driven methods, and [...] 
\begin{enumerate}[start=3]
    \item \textbf{Practice} (a) How does incorporating prior physical knowledge help overcome limitations of purely data-driven methods?
\end{enumerate}

Purely data-driven \ac{PHM} applications continue to face several challenges in real-world settings, including scarce and imbalanced degradation data, sensitivity to distributional shift across operating conditions and assets, poor extrapolation beyond the training regime, and predictions that can be physically implausible. Although \ac{PIML} is explicitly intended to mitigate these inherent shortcomings, empirical evidence only partially supports improvements in these areas. This does not imply that \ac{PIML} is incapable of delivering improvements, but that the number of systematic experiments specifically targeting these areas remains limited across the literature. Nevertheless, across all four classes of approaches, the reviewed studies consistently demonstrate improved predictive performance for each \ac{PHM} task and across a broad range of assets. 

There is also accumulating, though less robust, evidence that incorporating prior physical knowledge can enhance data efficiency, support more stable generalization, and enhance extrapolation. Furthermore, claims regarding improved interpretability, robustness, and physical consistency remain weakly substantiated: interpretability gains are typically inferred solely from the integration of prior knowledge but not quantified, robustness is defined inconsistently and is rarely evaluated across a sufficiently broad range of operating and environmental conditions, and physical consistency lacks a shared definition and corresponding metrics, thereby leaving reported improvements largely qualitative. 

Lastly, some approaches are by definition structurally ill-suited to deliver on certain promises. Observational-bias methods, for example, can plausibly tackle data scarcity and improve in-distribution generalization, but they leave the hypothesis space unchanged and are therefore poorly equipped to enforce physical consistency or principled extrapolation beyond the regimes spanned by the (simulated) training data. Similarly, hybrid approaches that employ an in-series coupling, in which a physics-based model feeds an \ac{ML} model, largely inherit the shortcomings attributed earlier to purely data-driven methods. 

Overall, the current evidence indicates that physics-informed approaches already provide tangible advantages, yet broader benefits frequently ascribed to \ac{PIML} in prior work are only partially substantiated and will require more rigorous, multi-dimensional evaluation before they can be regarded as confirmed.

% (b) what are the primary challenges in developing and applying physics-informed approaches?
\begin{enumerate}[start=3]
    \item \textbf{Practice} (b) What are the primary challenges in developing and applying physics-informed approaches?
\end{enumerate}

Despite the promising outlook suggested by the reviewed literature, efforts to develop and apply \ac{PIML} approaches within the field of \ac{PHM} remain subject to major challenges. With the current landscape being highly fragmented, the field largely lacks standardized design patterns. 

Furthermore, each method is characterized by a certain degree of entanglement between the embedded physics and the specific asset addressed---an entanglement that determines its applicability to other settings. Given that the literature has largely focused on a limited set of assets (where the same problem-specific limitations apply), the field is still regarded as nascent in terms of empirically grounded development of transferable methods. The evidence suggests that continued incremental work within existing silos is unlikely to produce cumulative progress. Instead, advancement requires moving from problem-specific solutions toward foundational degradation models that facilitate physics-informed modeling irrespective of the asset in question. 

Furthermore, uncertainty quantification, which is widely regarded as indispensable for risk-aware maintenance planning and safety-critical decision-making, is largely absent. Likewise, as noted earlier, the lack of evidence for interpretability impedes real-world adoption, since practitioners demand predictions that are not only accurate but also transparent and trustworthy. 

Lastly, online capability is rarely addressed explicitly, and closed-loop integration with existing monitoring and maintenance workflows is virtually never demonstrated. Consequently, from a deployment perspective, most methods remain at the level of offline prototypes rather than operational tools. Thus, the practical feasibility of large-scale industrial deployment has not yet been explored.

% - - - - - - - - - - declarations - - - - - - - - - - %
\section*{Declarations}
\textbf{Funding} The research leading to these results received funding from Deutsche Forschungsgemeinschaft (DFG, German Research Foundation) under grant number 514247199 as part of the research project \textit{TheoMation}.\\

\noindent\textbf{Conflict of Interest} The authors have no competing interests to declare that are relevant to the content of this article.




\setlength{\bibsep}{0pt}
\bibliography{references}
% \begin{appendices}
\onecolumn
\section*{Appendix}
\label{appendix}
\subsection*{Excluded Studies}
\label{appendix:excluded_records}

After full-text analysis of the 212 eligible studies, 83 were excluded, most commonly due to a small number of recurring reasons. Many contributions focused on domains or assets outside the \ac{PHM} scope adopted for this review (e.g., civil infrastructure or buildings), or did not perform any \ac{PHM} task (e.g., quality-control or design-phase fatigue life prediction only). A substantial share of studies did not satisfy the operational definition of \ac{PIML} used here: they were purely data‑driven, relied exclusively on simulated data without real measurements, or used domain knowledge only in the form of conventional feature engineering. Additional studies were excluded because they did not employ \ac{ML} at all, were abstract‑only or doctoral‑symposium contributions, or had been retracted. Table~\ref{tab:excluded_records} details the excluded records (totaling 83) and the rationale for their exclusion to ensure completeness and enhance the transparency of the review process.

\setlength{\tabcolsep}{4pt}
\setlength{\LTcapwidth}{\textwidth}
\begin{longtable}{
  >{\RaggedRight\arraybackslash}p{3cm}
  >{\RaggedRight\arraybackslash}p{11.2cm}
  >{\RaggedRight\arraybackslash}p{0.8cm}
}
    
    \caption{\raggedright This table lists all studies deemed outside the scope of this review following full-text analysis. For each study, the reference, the rationale for exclusion, and the reviewer are provided. Reviewers are identified as \enquote{A} and \enquote{B,} corresponding to the two equally contributing authors (in no particular order). The rationale is not a summary of the respective work but concisely explains the basis for exclusion, i.e., some rationales may require consultation of the original study for full context. }
    \label{tab:excluded_records}
    \\
    
    % FIRST HEAD — only appears on the first page
    \hline
    \textbf{Reference} & \textbf{Rationale} & \textbf{Rev.} \\
    \hline
    \endfirsthead
    
    % HEAD FOR FOLLOWING PAGES — repeats on subsequent pages
    \hline
    \textbf{Reference} & \textbf{Rationale} & \textbf{Rev.} \\
    \hline
    \endhead
    
    % Optional footer for all pages except the last
    \hline
    \multicolumn{3}{r}{\textit{Continued on next page}} \\
    \endfoot
    
    % Footer for the last page
    \hline
    \endlastfoot

    \cite{carter2025Imperfectphysicsguidedneural} & The use of abstract simulated systems (Tinkerbell attractor, R\"ossler attractor, and a continuous stirred tank reactor) leads to a scope mismatch, as the work is detached from industrial settings and lacks \ac{PHM} relevance (see Sec.~3; Fig.~5). & B \\

    \cite{che2025UnlockingInterpretablePrediction} & The \enquote{physics extractor} \ac{CNN} is trained to learn the electrochemical impedance spectroscopy from early $Q$-$V$ curves, after which \enquote{the learned physical features were augmented to the measured features} for capacity prediction, representing standard deep learning (see Fig.~1; Sec.~4; Eq.~1). & A \\

    \cite{han2025Physicsinformedsymbolicregression} & The selection of a recursive model due to the dynamic nature of tool wear as well as the selection of input features based on domain knowledge is insufficient to qualify as \ac{PIML} (see Sec.~4.5; Fig.~3). & B \\

    \cite{hong2025Physicsinformedmachinelearning} & A model is derived that can be used to quantify valve flow rate, but no \ac{PHM} task is addressed, i.e., the work rather serves as a \enquote{reference for research on the design and control methods of hydraulic control systems} (see Sec.~5). & B \\

    \cite{kadiwala2025Decodingdegradation:The} & The \ac{PDE} is identified via sparse regression on the same empirical dataset and the \enquote{solution of the \ac{PDE} is then integrated into the feature set} for a standard Gaussian process regression model, i.e., feature engineering (see Sec.~2.3/2.4; Fig.~4). & A \\

    \cite{li2025Ahybridphysics} & The proposed approach is designed to predict the fatigue life of alloy samples under multiaxial loading. Since the current state of the system is neglected by not taking live sensor data into account, this cannot be considered \ac{PHM} (see Sec.~2.2.5; Fig.~4).& B \\

    \cite{li2025Remainingusefullife} & Alongside end-to-end feature extraction from raw data, time-, frequency-, and time-frequency-domain features are computed, misleadingly framed as leveraging prior knowledge, i.e., standard feature engineering (see Fig.~3; Sec.~3.1.2; Tab.~1). & B \\

    \cite{liao2025Classifierguidedneuralblind} & Although labeled \enquote{physics-informed}, the method employs a conventional multi-term loss function that comprises kurtosis, $l_2$/$l_4$ norm and cross-entropy, without incorporating any explicit physical laws or constraints (see Tab.~1; Sec.~3; Eq.~29). & B \\

    \cite{lu2025Priorknowledgeembedding} & Although they are integrated into the latent space of an \ac{AE}, the prior information corresponds to common features (such as root mean square, standard deviation, kurtosis), which are therefore insufficient to qualify as \ac{PIML} (see Sec.~3.1.1; Tab.~1). & B \\

    \cite{luo2025Amethodfor} & Although the approach is physics-informed—using a loss term to regularize the monotonic relationship between membrane resistance and capacity loss—it is excluded by definition because the model is trained solely on simulated data (see Sec.~2.2; Fig.~4). & A \\

    \cite{stoyanov2025Modellingthefatigue} & By definition, using only so-called \enquote{physics-informed datasets} generated from high-fidelity thermo-mechanical finite-element simulations, without any real data, does not constitute \ac{PIML} (see Fig.~1; Sec.~3/4.1). & B \\

    \cite{sun2025Amethodfor} & This work is excluded as it was retracted at the request by the Editor-in-Chief due to plagiarism (see \href{https://www.sciencedirect.com/science/article/pii/S0378775324017191}{\textit{A method for estimating lithium-ion battery state of health based on physics-informed machine learning}}, accessed on December 23, 2025). & A \\

    \cite{wang2025DigitalTwinDrivenPhysically} & Since real-world data are used solely to validate the digital twin, and the proposed approach inherently relies on generated data to train the fault diagnosis model, this work is excluded by definition (see Fig.~9; Sec.~III-D/IV-C). & A \\
    
    \cite{wang2025Fewshotfaultdiagnosis} & The additional loss term designed to preserve distance information in the embedding space of continuous wavelet transformation snippets that are input to a vision Transformer cannot be considered as physically meaningful prior knowledge (see Sec.~3.2). & B \\

    \cite{yuwang2025MetaLearningandKnowledge} & The method assumes degradation follows an unknown \ac{PDE} in a latent state space, inferring the governing dynamics from data, thereby learning a \ac{PDE}-like relationship between the hidden state and \ac{RUL} without physical priors (see Sec.~3.2.3; Fig.~2). & A \\

    \cite{zhang2025Priorknowledgeinformedmultitask} & By learning ten \enquote{signal feature indicators} (such as max, min, standard deviation) through an auxiliary task in a shared \ac{CNN}, the method provides only feature self-supervision and does not incorporate prior physical knowledge (see Tab.~2; Sec.~2.4; Fig.~2). & B \\

    \cite{zheng2025Predictionmodeloptimization} & Although claiming to introduce prior knowledge, the method does not incorporate domain knowledge, relying instead on pre-trained weights and teacher outputs for transfer learning and distillation, i.e., a purely data-driven approach (see Sec.~3.1; Fig.~3). & A \\

    \cite{zhong2025MIPISincNet:Anexplainable} & A \enquote{physics-informed convolutional layer} with analytically designed, fixed kernels based on bearing fault orders and Sinc-based bandpass filters is used as the first network layer, which effectively amounts to feature engineering (see Sec.~3.2; Tab.~1). & B \\

    \hline % - - - 2024

    \cite{chen2024OpticalSpectralPhysicsInformed} & The \enquote{physics-informed} part fuses raw optical spectra (grayscale images) with a second input channel (selected emission lines, statistical/time-frequency features, and operating parameters), effectively performing feature engineering (see Sec.~III-C; Fig.~6). & B \\

    \cite{chen2024KnowledgeInformedWheelWear} & The authors refer to their method as \enquote{interpretable feature engineering,} relying on spectral denoising and extraction of the wheel perimeter-related dominant frequency as the interpretable feature, which does not qualify as \ac{PIML} (see Sec.~III-A/B; Fig.~8). & B \\

    \cite{chen2024KnowledgeEmbeddedAutoencoder} & Although integrated into the latent space of a convolutional \ac{AE}, the prior knowledge used solely consists of trivial statistical and frequency-domain features, which does not represent prior physical knowledge (see Sec.~III; Fig.~1). & B \\

    \cite{johannesexenberger2024GeneralizableTemperatureNowcasting} & Although the work is framed as relevant to predictive maintenance, it is limited to now-casting bearing temperature and does not perform any actual \ac{PHM} tasks, leaving the connection to \ac{PHM} superficial (see Sec.~1/2). & A \\

    \cite{fernandez2024Trainingofphysicsinformed} & Although technically representing a physics-informed approach, the work is excluded as it solely targets end-of-discharge prediction, where \enquote{further research should [be] undertaken about the aging effect in Li-ion batteries} (see Sec.~4).  & A \\

    \cite{fernandez2024Physicsguidedrecurrentneural} & While the proposed approach is indeed physics-informed, it specifically addresses accelerations in concrete buildings under seismic events, thereby illustrating a scope mismatch (see Sec.~4.2). & B \\

    \cite{ge2024Domainadaptationfor} & While physics-informed, the work falls outside the scope of this review, addressing structural health monitoring of civil infrastructure, i.e., steel beams and truss bridges  (see Sec.~3/4). & A \\

    \cite{han2024AnInterpretableCNN} & A \ac{CNN} with a built-in wavelet feature-extraction layer and reinforcement learning-guided selection is built, where \enquote{the validation set accuracy [...] is taken as a priori knowledge,} obtained during pretraining, making the approach purely data-driven (see Sec.~II-A; Fig.~1; Eq.~9). & A \\

    \cite{jia2024KneePointConsciousBatteryAging} & Extracting 17 statistical features from discharge, incremental capacity, and differential voltage curves, and augmenting them with degradation indicators from open-circuit voltage reconstruction and hybrid pulse power characterization testing, constitutes mere feature engineering (see Sec.~3; Fig.~4/5). & A \\

    \cite{feilongjiang2024SpatiotemporalAttentionbasedHidden} & While the approach may appear physics-informed, it assumes (without theoretical or domain-specific justification) a generic \ac{PDE} on a learned latent state with empirically chosen derivative order, resulting in an arbitrary modeling choice (see Sec.~2.3/3.3). & A \\

    \cite{kayedpour2024WindTurbineHybrid} & The proposed \enquote{hybrid physics-based deep learning framework} for wind turbine diagnosis is solely trained on data generated from a multiphysics simulation and is therefore excluded from this review by definition (see Sec.~I). & B \\

    \cite{kim2024Singledomaingeneralizable} & Using \enquote{prior knowledge that the bearing fault signals are impulse excitation signals} for signal processing, followed by a vanilla \ac{CNN} and an explainable artificial intelligence technique, the method combines conventional feature engineering with a post-hoc explanation (see Fig.~2; Sec.~3.2). & A \\

    \cite{kumari2024Efficientstochasticparametric} & By definition, training solely on simulated data, with no real-world data incorporated---as exemplified here by a stochastic battery degradation model---does not qualify as \ac{PIML} (see Fig.~1). & B \\
    
    \cite{lai2024PhysicsInformeddeepAutoencoder} & Due to the fact that the method for fault detection in electro-hydraulic servo actuators used in turbofan engine fuel systems involves solely simulated data for the development of the reconstruction model, the work is excluded by definition (see Sec.~3.1). & A \\
    \cite{li2024Particlefilterbasedfatiguedamageprognosisbyfusingmultipledegradationmodels} & Since the data-driven components are limited to fixed low-order polynomial regressions identified from experimental and simulated data, with no explicit ML model being employed, the study ultimately lies outside the \ac{PIML} scope (see Tab.~3). & A \\

    \cite{shang2024Anoveldata} & By aligning run-to-failure sequences with dynamic time warping and averaging them via weighted barycenter to create additional time series, the approach is an interpolative, correlation-aware yet standard data augmentation method (see Fig.~2/4; Sec.~3.1). & A \\

    \cite{su2024Knowledgeinformeddeepnetworks} & Although labeled \enquote{knowledge-informed}, the so-called knowledge-based features are merely standard time- and frequency-domain statistics, effectively amounting to conventional feature engineering (see Sec.~2.1; Fig.~2). & B \\

    \cite{shengyutao2024NondestructiveDegradationPattern} & The proposed \enquote{ultra-early prototype verification method} focuses on post-production quality assessment of lithium-ion batteries and is therefore excluded, as it does not implement operational \ac{PHM} (see Fig.~1; Discussions). & A \\

    \cite{xie2024Aknowledgedistillation} & The teacher Transformer is pre-trained to learn degradation patterns and provide \enquote{prior knowledge of degradation laws} to the student \ac{CNN}---knowledge that is merely learned from data (see Fig.~1/2; Methodology). & A \\

    \cite{yan2024DiscriminationandSparsityDrivenWeightOriented} & The method weights spectral lines using a generalized Rayleigh quotient eigenproblem, fuses each spectrum into a single \enquote{degradation feature,} and classifies them by Euclidean distance---reflecting classical linear algebra rather than \ac{ML} (see Sec.~II-C/D). & A \\

    \cite{yan2024NovelAnchorDiscrimination} & With the method being \enquote{formulated as a generalized Rayleigh quotient, which can be conveniently and easily solved by a maximum eigenvalue problem,} it exemplifies classical linear algebra rather than \ac{ML} (see Sec.~II-C; Eq.~11). & B \\

    \cite{yan2024PhysicsEnhancedNMFToward} & This study is excluded because it investigates high-speed train carriages, which are considered transportation systems and are therefore outside the scope of this review (see Fig.~1; Sec.~VII). & A \\

    \cite{yang2024Detectionofwind} & The proposed method, embedding a vanilla \textit{Neural~ODE} (a purely data-driven model that learns dynamics from data) into an \ac{AE} for anomaly detection, does not incorporate prior knowledge and is therefore not physics-informed (see Sec.~2.2/3.1.2). & A \\

    \cite{ye2024Amethodfor} & Ambiguous \enquote{secondary training} protocols (testing/validation), altering the physics constraint mid-work, and misreporting physical loss for the baseline neural network undermine the study's scientific rigor and comparability (see Sec.~2.3/3; Eq.~8/13; Fig.~8). & A \\

    \cite{zhou2024PriorKnowledgeAugmentedMetaLearning} & While a \enquote{prior knowledge-augmented} meta-learning framework is proposed, it merely uses generic time-domain indicators as pseudolabels and heatmaps (gradient-weighted class activation mapping) fused at test time, not constituting a physics-informed approach (see Fig.~1/2; Tab.~1; Sec.~III-A). & A \\

    \hline % - - - 2023

    \cite{abadi2023PhysicsInformedDeepLearningBased} & While appearing in the proceedings of the annual \ac{PHM} Society conference, it constitutes a PhD research outline for the doctoral symposium track rather than a full technical paper, and is therefore excluded (see Sec.~3). & A \\

    \cite{cvijic2023NeedforAI} & Although the approach is basically suitable for diagnosis, prognosis, and \ac{RUL} prediction of transformers, no quantitative results incorporating ground truth values are reported for the full framework (see Sec.~4). & B \\

    \cite{divyanshidwivedi2023DynamoPMU:APhysics} & Despite being labeled \enquote{physics-inspired,} the method for detecting anomalous events is entirely data-driven, relying solely on \enquote{$\mu$PMU measurement data without any information about the network model or prior labeling of the events} (see Abstract; Sec.~I‑B). & B \\
    
    \cite{fricke2023MissionSpecificPrognosisof} & Even though it appears in the proceedings of the annual \ac{PHM} Society conference, it represents a PhD research outline submitted to the doctoral symposium track rather than a full technical paper, and is therefore excluded (see Sec.~3). & A \\

    \cite{furlong2023APhysicsinformedTransfer} & Although published in the proceedings of the annual \ac{PHM} Society conference, it constitutes a PhD research outline submitted to the doctoral symposium track rather than a technical paper, and is therefore excluded (see Sec.~3). & A \\

    \cite{garpelli2023Physicsguidedneuralnetworks} & Although the approach embeds physics via a residual loss, it is excluded by definition because \enquote{simulated data is used during the learning process of the neural network, while experimental data is employed for the testing phase} (see Conclusion). & B \\

    \cite{alfonsogijon2023Predictionofwind} & The approach targets wind turbine power modeling, explicitly framed as a \enquote{previous step to developing optimal controllers and applying failure detection methods,} resulting in a scope mismatch, as no actual \ac{PHM} tasks are performed (see Sec.~1). & A \\

    \cite{he2023Asystematicmethod} & While presented as a \enquote{novel physics-informed loss function,} the additional term simply penalizes overestimation of \ac{RUL} more than underestimation for safety or economic reasons, which does not constitute prior physical knowledge (see Sec.~2.2). & B \\

    \cite{he2023Multiaxialfatiguelife} & This approach is designed for offline fatigue life prediction, mapping from load parameters (stress/strain) to the cycle-based fatigue life. The current system health state is not considered. Therefore, this method doesn't qualify as \ac{PHM} (see Sec.~1; Fig.~3). & B \\

    \cite{koutsoupakis2023Machinelearningbased} & By definition, the method does not qualify as \ac{PIML}, since the \ac{CNN} \enquote{is trained on numerical data alone,} generated via simulation for the purpose of identifying damage across different health states (see Sec.~5.1). & B \\

    \cite{kumar2023EstimatingRemainingUseful} & Contrary to the claim of proposing a novel approach, "which integrates machine learning techniques with electrochemical modeling," the implemented method is a conventional neural network, reflecting standard deep learning (see Sec.~2/4). & A \\

    \cite{lee2023Developmentofthe} & While motivated by \ac{PHM}, the work does not directly address \ac{PHM}, as the fault detection and diagnosis methodology \enquote{that uses the developed estimation model will be proposed in the future,} resulting in a scope mismatch (see Sec.~5). & A \\

    \cite{lei2023Priorknowledgeembeddedmetatransfer} & \enquote{Embedding prior knowledge} is limited to computed order tracking-based data augmentation---resampling vibration signals via pseudo‑speed ratios with amplitude scaling and Gaussian noise---followed by a standard metric‑based meta‑learner (see Sec.~3.1). & A \\

    \cite{liao2023Remainingusefullife} & With the underlying physics “completely unknown, a deepHPM can be used to approximate it,” the method effectively infers a \ac{PDE}-like relationship between the hidden state and \ac{RUL} from data, rather than integrating prior physical knowledge (see Sec.~2.2; Fig.~2). & B \\

    \cite{liu2023KnowledgeEmbeddedLightweight} & A method  is proposed, \enquote{where the prior knowledge of machining parameters is fused with features extracted from multiple sensor information} to construct \enquote{hybrid texture data} fed into a vision Transformer, i.e., feature engineering (see Sec.~2.2; Fig.~1/2). & A \\

    \cite{ma2023PhysicsInformedMachineLearning} & While both \enquote{physics‑informed} feature extraction and parameter tuning are claimed, the former reduces to selecting leakage‑related inputs and adopting rise time as the health indicator, and the latter is merely a vanilla hyperparameter search (see Sec.~2.1/2.2/3.2). & A \\

    \cite{mochammad2023EnhancingRealisticRemaining} & The work lacks key methodological details (e.g., low-fidelity model identification, datasets, training procedure, used loss function, and questionable baseline comparisons), preventing a rigorous assessment of its contribution (see Sec.~2). & B \\

    \cite{tu2023Integratingphysicsbasedmodeling} & Although a series of hybrid models are proposed, with the sole focus of enabling highly accurate voltage prediction for lithium-batteries, the work does not directly address \ac{PHM} and is therefore excluded (see Sec.~2.1; Fig.~2). & A \\

    \cite{wang2023InherentlyInterpretablePhysicsInformed} & The approach is indeed \ac{PIML}. As it is used for end-of-discharge prediction of lithium-ion batteries, hence not taking aging effects into account, it cannot be seen as \ac{PHM} (see~Sec. IV-F). & B \\

    \cite{wang2023Interpretableconvolutionalneural} & The method is a standard supervised \ac{CNN} trained with cross-entropy, merely embedding discrete wavelet transform and attention as signal-processing layers to improve feature learning and noise robustness (see Fig.~2; Sec.~3.5). & A \\

    \cite{weddle2023Batterystateofhealthdiagnostics} & By definition, training (in this case) VGG-16 solely on simulated battery degradation does not qualify as \ac{PIML} due to the absence of real-world measurements (see Sec.~3.4; Fig.~5). & A \\

    \cite{wu2023Physicsinformedgatedrecurrent} & Given that the \enquote{Secure Water Treatment testbed (SWaT) and Water Distribution testbed (WADI)} datasets are designed for cybersecurity research, the anomaly detection task does not target asset health, placing the work outside the scope of \ac{PHM} (see Sec.~4.1). & B \\

    \cite{zhang2023DuAK:ReinforcementLearningBased} & While industrial, the work focuses on quality inspection of manufactured steel products rather than the health of industrial assets, which is the core concern of \ac{PHM}, resulting in an inherent scope mismatch (see Sec.~I/II-A). & A \\

    \hline % - - - 2022

    \cite{ariaschao2022Fusingphysicsbasedand} & While theoretically physics-informed, the use of \enquote{a discrete-time counterpart of the physics-based model F in the form of a deep neural network} effectively yields a purely data-driven approach (see Sec.~3.1; Eq.~5; Fig.~3). & A \\

    \cite{chen2022PhysicsInformedLSTMhyperparameters} & The optimal parameters of the gearbox fault detection model considered in this study are determined by superimposing crack fault signatures onto validation data for hyperparameter optimization. This is a data augmentation technique (see Sec.~3.1; Fig.~3). & B \\

    \cite{hajiha2022Aphysicsregularizeddatadriven} & The proposed  \enquote{physics-regularized data-driven approach} relies, upon closer examination, exclusively on classical probabilistic modeling rather than traditional \ac{ML}, revealing a scope mismatch (see Sec.~2). & A \\

    \cite{russell2022Physicsinformeddeeplearning} & While stating that \enquote{including a loss term during AE training that is sensitive to frequency content introduces a physically informed objective,} additional autocorrelation and Fast Fourier transform-based losses are effectively equivalent to standard multi-task learning (see Sec.~2.4). & A \\

    \cite{zgraggen2022PhysicsInformedDeep} & Focusing on tracker faults that usually occur \enquote{when the tracker gets stuck at a certain orientation instead of tracking the sun,} this work addresses operational anomaly detection rather than asset health, and is therefore excluded (see Sec.~1). & A \\

    \hline % - - - 2021

    \cite{guo2021Parameteridentificationof} & In this approach, solely a \ac{NN} is used to estimate battery health indicators. A fractional-order model reconstructs measured quantities based on the health indicators for validation purposes, yet not directly relevant to health assessment (see Sec.~4.2/4.3). & B \\

    \cite{keizers2021UnscentedKalmanFiltering} & Although the method aims to predict \ac{RUL}, its reliance on an Unscented Kalman Filter for online parameter and state estimation rather than on \ac{ML} leads to a fundamental mismatch in scope (see Sec.~3.3/3.4/4.3). & A \\  
    
    \cite{li2021Particlefilterbasedhybriddamageprognosisconsideringmeasurementbias} & The data-driven part is limited to an offline fitted, low-order polynomial measurement equation whose parameters remain fixed during prognosis, resulting in a scope mismatch due to the lack of \ac{ML} (see Sec.~3.2; Eq.~18). & A \\

    \cite{lyathakula2021Aprobabilisticfatigue} & The work targets probabilistic fatigue life prediction based on offline test data, intended to replace/reduce expensive fatigue testing in the design phase of adhesively bonded joints, essentially not performing any \ac{PHM} task using in-situ monitoring data (see Sec.~2). & A \\

    \cite{wang2021Digitaltwinenhanced} & By definition, an approach that relies entirely on simulated data from a digital twin to train a \ac{CNN} for autoclave fault prediction, without using actual operational data, does not qualify as \ac{PIML} (see Sec.~6). & B \\

    \hline % - - - 2020

    \cite{cofremartel2020Aphysicsinformeddeep} & This work is an abstract-only contribution and is therefore excluded for providing insufficient methodological and empirical detail (see \href{https://www.rpsonline.com.sg/proceedings/esrel2020/html/4973.xml}{\textit{A physics-informed deep learning approach for fatigue crack propagation}}, accessed on December 23, 2025). & B \\

    \cite{kobrich2020Physicsbaseddeep} & Targeting crack growth prediction, the work proposes training a deep learning model solely on data simulated using extended finite element method, which by definition precludes it from being a \ac{PIML} approach due to the absence of real-world measurements (see Sec.~4.2). & A \\

    \hline % - - - 2019

    \cite{akkad2019Aphysicsbased} & Although it appears in the proceedings of the annual \ac{PHM} Society conference, it reflects a PhD research outline submitted to the doctoral symposium track rather than a full technical paper, and is therefore excluded (see Sec.~3). & A \\

    \cite{sadoughi2019Adeeplearning} & Using \enquote{physics-based feature extraction, which is based on conventional signal processing techniques in time and frequency domain} to obtain features such as root mean square, peak-to-peak, and kurtosis does not constitute a \ac{PIML} approach (see Sec.~I/II-B). & A \\

    \hline %  - - - 2018

    \cite{neerukatti2018Ahybridprognosis} & The proposed \enquote{hybrid prognosis model} for predicting crack propagation is solely trained on data generated from finite element simulations, and is therefore excluded from this review by definition (see Prognosis model). & A \\

    \hline %  - - - 2015

    \cite{liu2015Faultdiagnosisfor} & The work focuses on a solar‑assisted heat pump system and proposes a fault diagnosis method which is solely trained on incomplete simulation data, and is therefore excluded by definition (see Abstract; Sec.~5). & A \\

    \hline % - - - 2013

    \cite{kulkarni2013Physicsbaseddegradation} & The development of physics-based degradation models is proposed to predict electrolytic capacitor aging under thermal overstress, using experimental data for calibration without employing any \ac{ML} techniques (see Sec.~3.2; Fig.~2).  & A \\
 
    \hline
\end{longtable}
\twocolumn
% \end{appendices}

\end{document}